\documentclass[10pt,journal,compsoc]{IEEEtran}

\ifCLASSOPTIONcompsoc
  \usepackage[nocompress]{cite}
\else
  \usepackage{cite}
\fi
\usepackage{iftex}
\ifPDFTeX
  \usepackage[utf8]{inputenc}
  \usepackage[T1]{fontenc}
  \usepackage{CJKutf8}
\else
  \usepackage{fontspec}
  \usepackage{xeCJK}
  \defaultfontfeatures{Ligatures=TeX}
  \IfFontExistsTF{Noto Serif CJK SC}{%
    \setCJKmainfont{Noto Serif CJK SC}%
  }{%
    \IfFontExistsTF{Source Han Serif SC}{%
      \setCJKmainfont{Source Han Serif SC}%
    }{%
      \IfFontExistsTF{Noto Sans CJK SC}{%
        \setCJKmainfont{Noto Sans CJK SC}%
      }{%
        \IfFontExistsTF{AR PL UMing CN}{%
          \setCJKmainfont{AR PL UMing CN}%
        }{}%
      }%
    }%
  }%
  \XeTeXlinebreaklocale "zh"
\fi

\usepackage{hyperref}
\usepackage{xurl}
\usepackage{amsfonts}
\usepackage{amsmath}
\usepackage{amssymb}
\usepackage{dsfont}

\usepackage{graphicx}
\usepackage{subfigure}
\usepackage{booktabs}
\usepackage{multirow}
\usepackage{adjustbox}
\usepackage{dblfloatfix}
\usepackage{caption}
\usepackage{xltabular}
\usepackage{longtable}
\usepackage{tabularx}
\usepackage{array}

\usepackage[dvipsnames]{xcolor}
\usepackage{tikz}
\usetikzlibrary{calc,arrows.meta,intersections,patterns,positioning,shapes.misc,fadings,through,decorations.pathreplacing,shadows,chains,mindmap}
\usepackage{forest}

\usepackage[normalem]{ulem}

\usepackage{dramatist}
\usepackage{xspace}
\usepackage{pifont}
\usepackage{tcolorbox}
\usepackage[bottom]{footmisc}
\usepackage{setspace}
\usepackage{lipsum}
\usepackage{multicol}
\usepackage{lineno}

\usepackage{microtype}

\makeatletter
\def\@BTrule[#1]{%
  \ifx\longtable\undefined
    \let\@BTswitch\@BTnormal
  \else\ifx\hline\LT@hline
    \nobreak
    \let\@BTswitch\@BLTrule
  \else
     \let\@BTswitch\@BTnormal
  \fi\fi
  \global\@thisrulewidth=#1\relax
  \ifnum\@thisruleclass=\tw@\vskip\@aboverulesep\else
  \ifnum\@lastruleclass=\z@\vskip\@aboverulesep\else
  \ifnum\@lastruleclass=\@ne\vskip\doublerulesep\fi\fi\fi
  \@BTswitch}
\makeatother

 {\begin{list}{}%
         {\setlength{\leftmargin}{#1}}%
         \item[]%
 }
 {\end{list}}
\title{Vision-and-Language Navigation for UAVs: Progress, Challenges, and a Research Roadmap}

\author{Hanxuan~Chen,
        Jie~Zheng,
        Siqi~Yang,
        Tianle~Zeng,
        Siwei~Feng,
        Songsheng~Cheng,
        Ruilong~Ren,
        Hanzhong~Guo,
        Shuai~Yuan,
        Xiangyue~Wang,
        Kangli~Wang,
        and~Ji~Pei%
\thanks{J. Zheng and S. Yang contributed equally and are co-second authors.}%
\thanks{H. Chen, S. Feng, S. Cheng, R. Ren, X. Wang, K. Wang, and J. Pei are with Autel Robotics, Shenzhen, China.}%
\thanks{J. Zheng is with the School of Intelligent Software and Engineering, Nanjing University, Nanjing, China.}%
\thanks{S. Yang is with the School of Computer, Data \& Information Sciences, University of Wisconsin-Madison, Madison, WI, USA.}%
\thanks{T. Zeng is with Southern University of Science and Technology, Shenzhen, China.}%
\thanks{H. Guo is with The University of Hong Kong, Hong Kong SAR, China.}%
\thanks{S. Yuan is with the School of Software and Microelectronics, Peking University, Beijing, China.}%
\thanks{Corresponding author: J. Pei (peiji@autelrobotics.com).}%
}
\begin{document}

\IEEEtitleabstractindextext{%
\begin{abstract}

Vision-and-Language Navigation for Unmanned Aerial Vehicles (UAV-VLN) represents a pivotal challenge in embodied artificial intelligence, focused on enabling UAVs to interpret high-level human commands and execute long-horizon tasks in complex 3D environments. This paper provides a comprehensive and structured survey of the field, from its formal task definition to the current state of the art. We establish a methodological taxonomy that charts the technological evolution from early modular and deep learning approaches to contemporary agentic systems driven by large foundation models, including Vision-Language Models (VLMs), Vision-Language-Action (VLA) models, and the emerging integration of generative world models with VLA architectures for physically-grounded reasoning. The survey systematically reviews the ecosystem of essential resources simulators, datasets, and evaluation metrics that facilitates standardized research. Furthermore, we conduct a critical analysis of the primary challenges impeding real-world deployment: the simulation-to-reality gap, robust perception in dynamic outdoor settings, reasoning with linguistic ambiguity, and the efficient deployment of large models on resource-constrained hardware. By synthesizing current benchmarks and limitations, this survey concludes by proposing a forward-looking research roadmap to guide future inquiry into key frontiers such as multi-agent swarm coordination and air-ground collaborative robotics.

\end{abstract}

}

\maketitle
\IEEEdisplaynontitleabstractindextext

\section{Introduction} \label{sec:Introduction}

\begin{figure*}[!htb]
    \centering
    \includegraphics[width=0.9\textwidth,height=0.5\textheight,keepaspectratio]{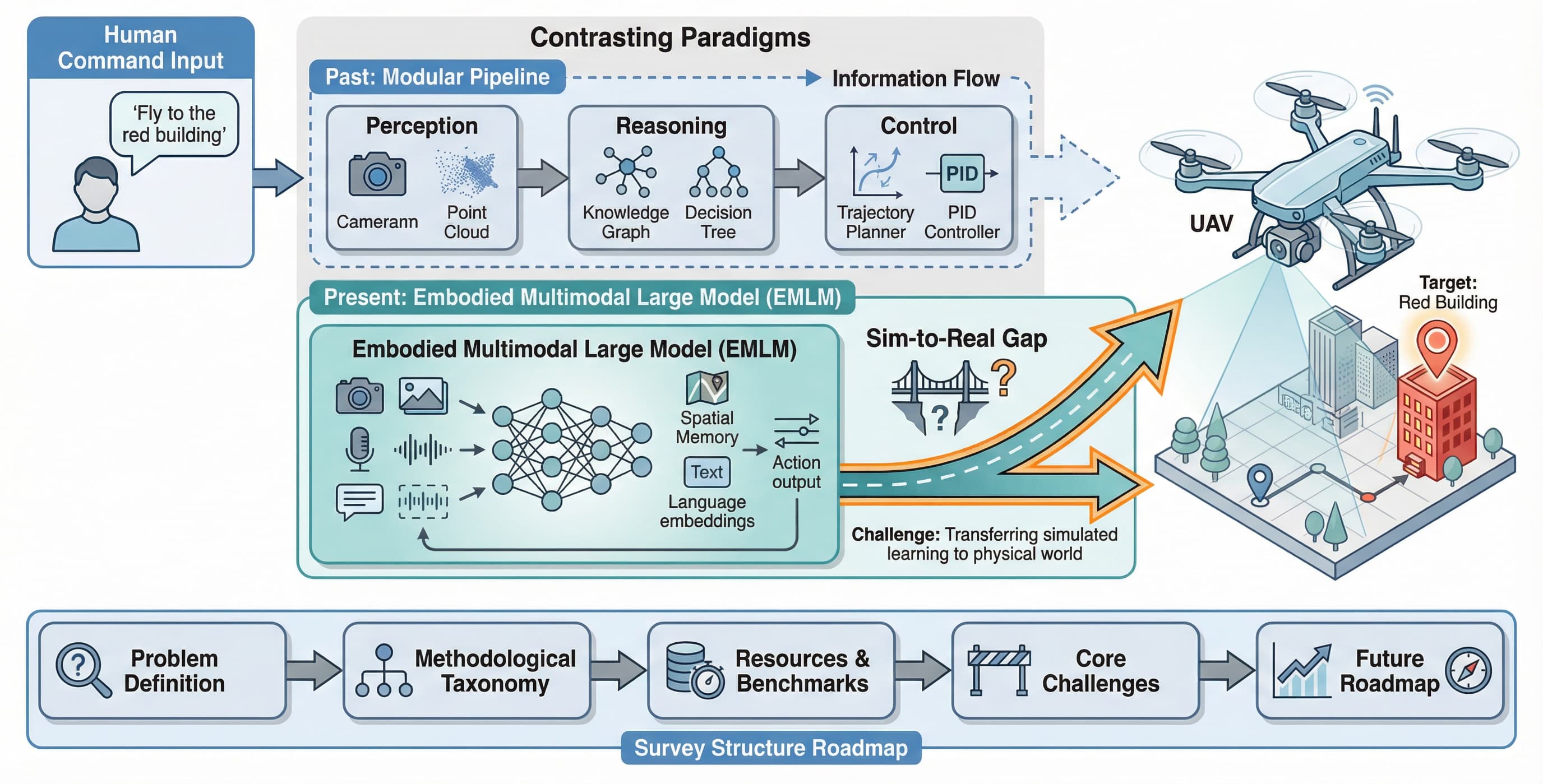}
    \caption{An overview of the UAV-based Vision-and-Language Navigation (UAV-VLN) research landscape, illustrating the core components and methodological evolution. The process begins with a natural language command that directs an autonomous agent. This figure contrasts the traditional modular pipeline, which separates perception, reasoning, and control, with the modern integrated approach centered on Embodied Multimodal Large Models (EMLMs). The UAV executes the resulting action in a complex 3D world, navigating key challenges such as the simulation-to-reality gap. The bottom panel outlines the structure of this survey, mapping the progression from foundational definitions to a forward-looking research roadmap.}
    \label{fig:main_figure}
\end{figure*}
Enabling an Unmanned Aerial Vehicle (UAV) to navigate a complex, three-dimensional world from a simple human command such as ''fly past the collapsed bridge and find people waving from the rooftops'' represents a pivotal challenge at the intersection of robotics, computer vision, and natural language understanding. This capability, known as UAV-based Vision-and-Language Navigation (UAV-VLN), is a crucial subfield within the broader pursuit of embodied artificial intelligence, which aims to develop autonomous agents that interpret linguistic instructions and execute long-horizon tasks in the physical world \cite{embodied_ai2024T, large_model_embodied_ai2025, embodied_ai2024Y2024}. The transition towards intuitive, language-based control moves beyond traditional interfaces, enhancing human-drone collaboration for complex operations and making sophisticated aerial platforms more accessible \cite{large_language_models2025K2025, human_drone2025, companion_uavs2020}. The real-world applications are profound, spanning time-critical search and rescue operations in GPS-denied environments \cite{air_ground_robots2025, flightgpt2024, navigation_gps2026}, wildfire monitoring \cite{wildfire_detection2025}, automated inspection of large-scale infrastructure \cite{waypoint_planning2026, lightweight_drone2024}, and dynamic logistics within smart cities \cite{logisticsvln2025, semantic_segmentation2024}. As UAVs become integral to these domains, the demand for intelligent, language-driven autonomy has motivated a surge of research \cite{uav_agentic2025}, necessitating a structured synthesis to guide future progress in this domain of applied spatial intelligence \cite{spatial_assisted2024, spatial_intelligence2025}.

The field of UAV-VLN is currently experiencing a paradigm shift, propelled by the confluence of mature aerial platforms and the transformative capabilities of large foundation models \cite{vision_language_navigation2024, foundation_models2023}. This evolution marks a significant departure from earlier modular pipelines toward integrated Embodied Multimodal Large Models (EMLMs) that unify perception, reasoning, and control into a cohesive framework \cite{embodied_multimodal2025, pure_vla2025, vision_language_action_models, embodied_navigation_foundation_model2024}. Most notably, the latest frontier involves the deep integration of generative world models with Vision-Language-Action (VLA) policies, as seen in models like $\pi_0$~\cite{pi0_2024}, GR00T N1~\cite{groot_n1_2025}, and Cosmos-Reason1~\cite{cosmos_reason1_2025}, which equip agents with physical common sense and predictive capabilities for robust, long-horizon reasoning. While substantial progress has been made in Vision-and-Language Navigation (VLN) for ground-based robots since the seminal work on navigating from photorealistic images \cite{vision_language_navigation2018, vision_language_navigation2021}, the aerial domain introduces a distinct and more complex set of challenges, a gap first systematically addressed by benchmarks like AerialVLN \cite{aerialvln2023}. These challenges, which have historically limited research in outdoor aerial settings \cite{openfly2024}, include navigating continuous 3D action spaces without predefined graphs, performing long-horizon reasoning with complex spatial relationships in unstructured environments \cite{aeroduo2025, citynavagent2025, citynav2025}, and overcoming a severe simulation-to-reality gap that complicates real-world deployment \cite{robotic_navigation2025}. Recent benchmarks for both outdoor \cite{openfly2025} and indoor \cite{indooruav2025} scenarios are beginning to address the data scarcity that has created a fragmented research landscape, which this survey aims to unify.

This paper provides a comprehensive and structured survey of UAV-based Vision-and-Language Navigation, charting the field from its conceptual foundations to the current state of the art. Our scope encompasses the entire research pipeline, commencing with the formal mathematical definition of the task and tracing the methodological evolution from early learning systems to contemporary agents driven by foundation models \cite{uavs_llms2025, large_language_models2024}. We survey the critical ecosystem of resources including high-fidelity simulators, benchmark datasets for diverse domains such as agriculture and urban reconnaissance \cite{agrivln2024, urban_monitoring2025}, and standardized evaluation metrics that underpins reproducible research. A core focus is the critical analysis of fundamental challenges, namely the sim-to-real gap, perception robustness, reasoning with language ambiguity, and the efficient deployment of large models on resource-constrained hardware. We explicitly exclude topics outside this focus, such as low-level flight controller design, detailed UAV hardware avionics \cite{uavs_avionics2024}, and broad surveys on ground-based VLN \cite{vision_language_navigation2021}. Furthermore, we distinguish the egocentric, action-oriented perspective of UAV-VLN from the exocentric, analytical tasks addressed by Remote Sensing Foundation Models \cite{foundation_models_survey, vision_language_modeling2025}, referencing them only to contextualize the unique challenges of aerial navigation.

The primary contributions of this survey are the establishment of a clear methodological taxonomy that traces the field's trajectory, a synthesized analysis of the core technical challenges impeding real-world deployment, and a forward-looking research roadmap. To deliver these contributions, the remainder of this paper is structured to guide the reader from foundational concepts to future frontiers. Section 2 establishes the theoretical groundwork by formally defining the UAV-VLN problem as a Partially Observable Markov Decision Process. Section 3 presents our methodological taxonomy, charting the evolution of agent architectures from modular and early learning approaches to modern foundation model-driven systems. Section 4 provides a comprehensive overview of the essential simulators, datasets, and evaluation protocols that enable standardized benchmarking. Sections 5 and 6 are dedicated to deep dives into the primary challenges, focusing on the sim-to-real gap and the intertwined issues of robustness, safety, and efficiency. Section 7 outlines a research roadmap toward future frontiers, including multi-agent swarm coordination \cite{uav_swarm2024, intelligent_optimization2025, multi_uav_systems} and air-ground collaborative robotics \cite{air_ground_robots2025}. Finally, Section 8 concludes with a summary of our key findings and reiterates the most promising directions for future inquiry.

\section{Formalizing the UAV-VLN Problem} \label{sec:Formalizing the UAV-VLN Problem}

To establish a rigorous mathematical foundation, the UAV-VLN task is formally modeled as a Partially Observable Markov Decision Process (POMDP), the standard framework for sequential decision-making under uncertainty in robotics and embodied AI \cite{safe_uav2025, on_device_learning2024}. This formalism is particularly apt because the agent's true state is never fully known; it must infer its condition from a stream of incomplete and often noisy sensory data, a defining characteristic of navigation in complex or GPS-denied environments \cite{vision_based2024, autonomous_navigation2025}. The UAV-VLN problem is thus defined by the tuple $\mathcal{M} = (\mathcal{S}, \mathcal{A}, \mathcal{T}, \mathcal{R}, \Omega, \mathcal{O})$, where $\mathcal{S}$ is the set of unobservable world states, $\mathcal{A}$ is the agent's action space, $\mathcal{T}$ represents the probabilistic state transition dynamics, $\mathcal{R}$ is the reward function, $\Omega$ is the set of possible observations, and $\mathcal{O}$ is the observation function \cite{llm_enhanced_path2025}. For cooperative multi-agent scenarios, this formulation extends to a Decentralized POMDP (DEC-POMDP) to model joint actions and observations \cite{scalable_cooperative2023, multi_agent_drl_uav}. The objective is to find an optimal policy, $\pi$, which maps a history of observations to the next action, thereby realizing the perception-reasoning-action loop fundamental to creating active, decision-making agents \cite{pure_vla2025, uav_agentic2025}.

\begin{figure*}[!htb]
    \centering
    \includegraphics[width=0.9\textwidth,height=0.5\textheight,keepaspectratio]{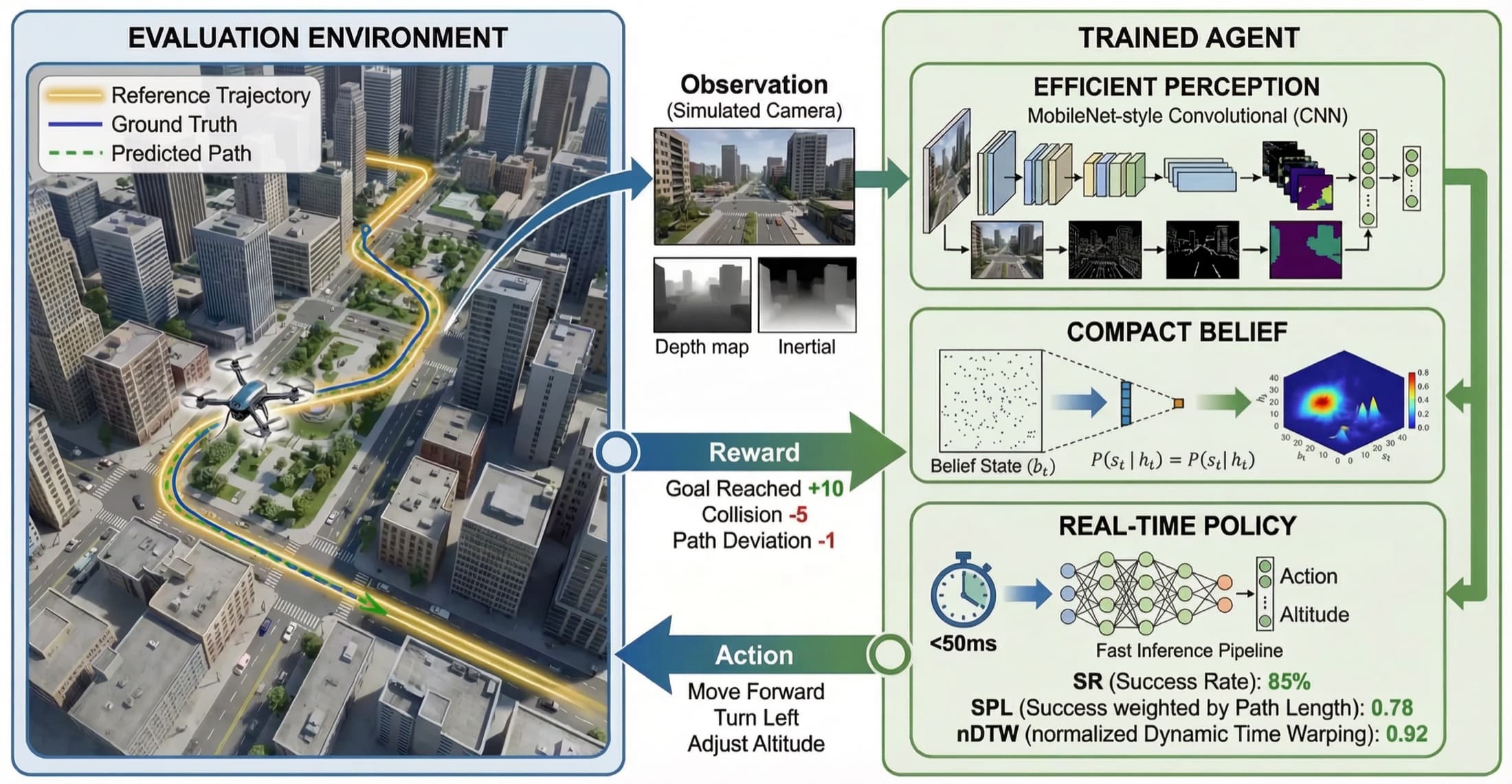}
    \caption{The UAV-VLN task modeled as a Partially Observable Markov Decision Process (POMDP). The agent receives an observation and reward from the environment, updates its internal belief state, and selects an action based on its policy to influence the environment's next state.}
    \label{fig:formalizing_the_uav-vln_problem_auto_1}
\end{figure*}

The state and observation spaces formalize the agent's perception, encapsulating the difficulty of grounding linguistic concepts onto vast, unstructured 3D environments from a limited sensory viewpoint. The unobservable state $s_t \in \mathcal{S}$ at timestep $t$ encompasses the true, complete state of the world, including the UAV's full 6-DoF pose and the environment's geometry. In contrast, the observation $o_t \in \Omega$ represents the partial information available to the agent, typically a composite of visual input from its egocentric camera(s) \cite{memory_based_deep2018}, proprioceptive data from an Inertial Measurement Unit (IMU), and the natural language instruction $\mathcal{L}$ \cite{aerial_vision_dialog2024}. A primary challenge is the inherent ambiguity in these instructions \cite{vision_language2024}, requiring the agent to perform robust cross-modal reasoning to ground abstract descriptions, such as occluded or implicitly referenced objects, onto the visual scene \cite{reasongrounder2025}. This has motivated the development of explicit world representations, such as spatial maps that fuse visual-language features \cite{visual_language_maps2023} or semantic octrees \cite{semantic_octree_mapping}, to augment the agent's belief over the true state $s_t$.

The action space $\mathcal{A}$ for aerial navigation introduces a fundamental trade-off between control fidelity and planning complexity, a significant departure from the constrained action sets of ground-based agents.

\begin{figure*}[!htb]
    \centering
    \includegraphics[width=0.9\textwidth,height=0.5\textheight,keepaspectratio]{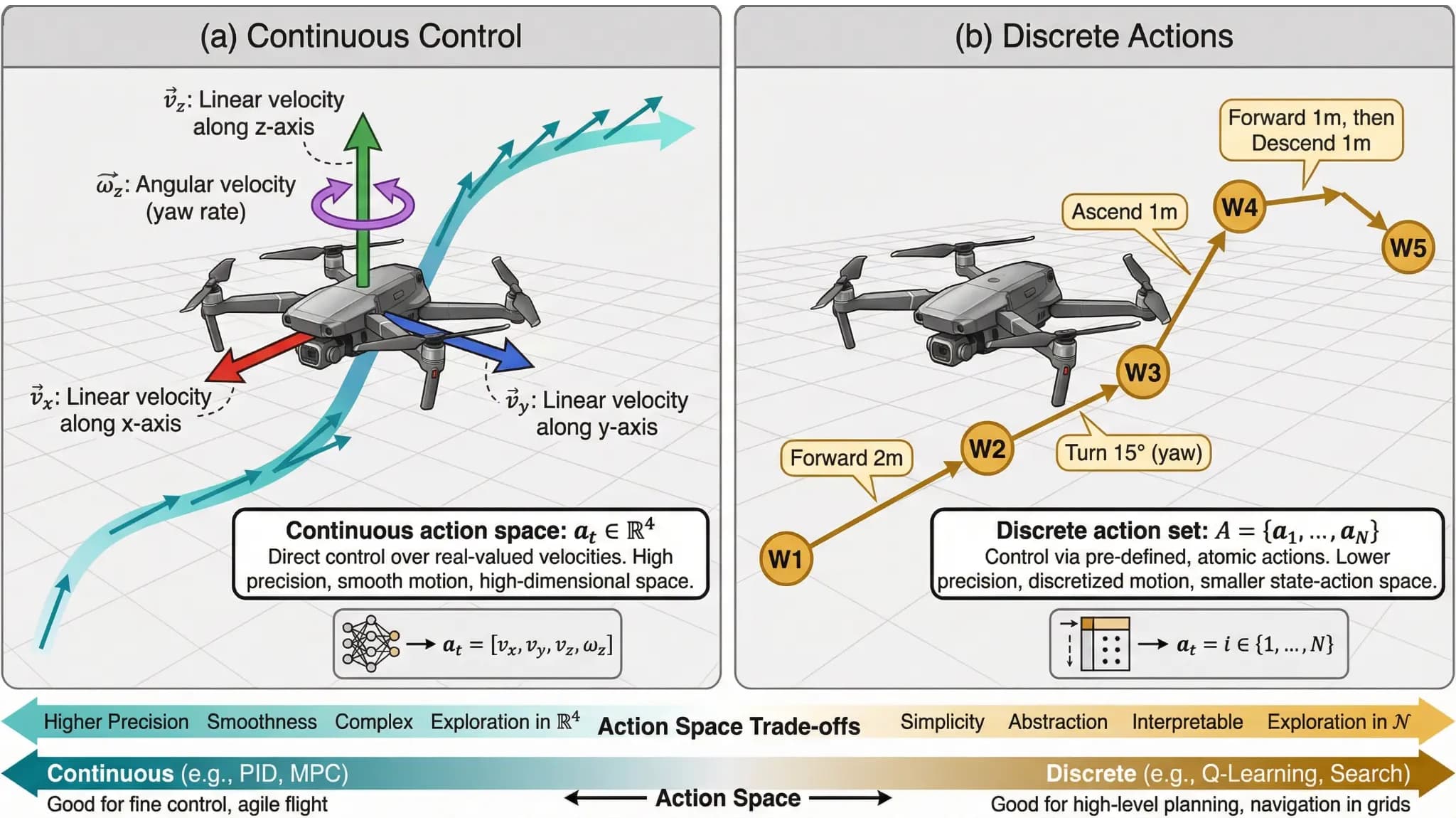}
    \caption{A comparison of action space formulations for UAV navigation. Low-level continuous control offers fine-grained maneuverability at the cost of planning complexity, whereas high-level discretized actions simplify decision-making but can result in less efficient paths.}
    \label{fig:formalizing_the_uav-vln_problem_auto_2}
\end{figure*}
 UAVs operate in a continuous 3D space, framing the task as Vision-and-Language Navigation in Continuous Environments (VLN-CE) \cite{dagger_diffusion2021}. This space is typically represented in two ways: low-level continuous control, where an action $a_t \in \mathcal{A} \subseteq \mathbb{R}^k$ is a vector of linear and angular velocity commands (e.g., $a_t = [v_x, v_y, v_z, \omega_z]$), which offers fine-grained maneuverability but creates a high-dimensional search space that complicates long-horizon planning \cite{model_free_uav2024, uav_flow2025, long_horizon_vln2025}. Alternatively, a high-level discretized action space maps continuous commands to a finite set of maneuvers, such as waypoint navigation or predefined flight primitives (`fly forward 2 meters`, `rotate 15 degrees`). This approach simplifies the decision-making process but can lead to suboptimal paths and may not fully exploit the UAV's dynamic capabilities \cite{uav_vln2025, waypoint_models2021, uav_vla2025}.

The agent's objective is operationalized through a reward function $\mathcal{R}$ and a set of standardized success metrics that evaluate both task completion and navigation efficiency. The primary goal is to reach a target location described by the instruction $\mathcal{L}$, with success often defined as stopping within a predefined radius of the goal \cite{flightgpt2024}. The reward function $\mathcal{R}: \mathcal{S} \times \mathcal{A} \to \mathbb{R}$ is typically sparse, providing a large positive reward for task success and small negative penalties for actions that consume time or result in collisions \cite{reprohrl2023, path_planning_algorithm}. A policy $\pi$ is optimized to maximize the expected cumulative discounted reward, often incorporating an entropy term to encourage exploration in reinforcement learning settings \cite{multi_uav_cooperative_path2024}:
\begin{equation}
J(\pi) = \mathbb{E}_{\tau \sim \pi}\left[\sum_{t=0}^{H} \gamma^t \mathcal{R}(s_t, a_t) + \alpha \mathcal{H}(\pi(\cdot|s_t))\right]
\end{equation}
where $\tau$ is a trajectory, $H$ is the horizon, $\gamma \in [0, 1]$ is a discount factor, and $\alpha$ balances the reward with the policy entropy $\mathcal{H}$. Performance is evaluated using metrics adapted from ground-based VLN, notably Success Rate (SR) and Success weighted by Path Length (SPL) \cite{aerialvln2023}. More sophisticated, dense reward functions are also being explored, incorporating terms for curiosity, battery efficiency, and even feedback from large language models \cite{prompt_informed_reinforcement2025}, alongside objectives that ensure safety \cite{asma2026} or sociability in human-centric contexts \cite{human_centric_uav2024, companion_uavs2020}.

\section{A Methodological Taxonomy of UAV-VLN Agents} \label{sec:A Methodological Taxonomy of UAV-VLN Agents}

\begin{table*}[ht]
\centering
\resizebox{\textwidth}{!}{%
\begin{tabular}{lccc}
\toprule
\textbf{{Feature}} & \textbf{Classical Decomposed Robotics Pipelines} & \textbf{Modular Deep Learning Architectures} & \textbf{Emergence of End-to-End Learning Policies} \\
\midrule
\textbf{Architectural Approach} & Decomposed Algorithmic Stages & Modular Neural Components & Direct Sensor-to-action Mapping \\
\textbf{Core Component} & Classical Planner/controller & Cross-modal Fusion Module & Unified Visuomotor Policy \\
\textbf{Generated Output} & Explicit Geometric Trajectory & Fused Feature Representation & Direct Flight Commands \\
\bottomrule
\end{tabular}}
\caption{A comparative analysis of foundational architectural paradigms in early UAV-VLN systems. The table illustrates the evolution from classical decomposed robotics pipelines to modular deep learning architectures and end-to-end policies, contrasting their core architectural approach, primary components, and generated outputs.}
\label{tab:comparison_table}
\end{table*}
This section presents a methodological taxonomy charting the architectural evolution of UAV-VLN agents, organized according to their core reasoning mechanisms and degree of system integration. The classification traces the field's progression from functionally decomposed pipelines to integrated, end-to-end policies and, ultimately, to agentic systems driven by large foundation models. As illustrated in Table~\ref{tab:taxonomy_tree}, this taxonomy delineates three main paradigms: (1) \textit{Modular and Early Learning Approaches}, which established foundational techniques through hierarchical task decomposition; (2) \textit{Architectures for Long-Horizon Spatiotemporal Understanding}, which introduced advanced memory and world modeling capabilities; and (3) \textit{Foundation Model-Driven Agentic Systems}, which leverage large models for integrated reasoning and control. To provide a consolidated reference for the methods discussed throughout this section, Table~\ref{tab:summary_table} offers a comprehensive summary organized according to this methodological taxonomy.
\begin{table*}[!htb]
\centering
\resizebox{\textwidth}{!}{%
\begin{tabular}{llllll}
\toprule
\textbf{Paradigm} & \textbf{Method} & \textbf{Year} & \textbf{Model Type} & \textbf{Core Technique} & \textbf{Key Contribution} \\
\midrule
\multirow{6}{*}{\parbox{3.2cm}{\textbf{Modular \& Early\\Learning}}}
 & CMA / LAG baselines \cite{aerialvln2023} & 2023 & CNN-RNN & Cross-modal attention & First large-scale UAV-VLN benchmark baselines \\
 & RRT + NLU pipeline \cite{uav_path2023} & 2023 & Hybrid & RNN parser + RRT planner & NLU-guided classical planning for UAVs \\
 & SLAM + Visual Servoing \cite{gps_denied_ibvs2025} & 2025 & Classical & SLAM + IBVS control & GPS-denied navigation with decoupled perception-control \\
 & DRL Obstacle Avoidance \cite{drone_obstacle_avoidance} & 2021 & DRL & DQN / A2C policies & End-to-end visuomotor learning for collision avoidance \\
 & Memory-based DRL \cite{memory_based_deep2018} & 2018 & DRQN & CNN-RNN + RL & Recurrent memory for sequential navigation at 60\,Hz \\
 & DAGGer + Diffusion \cite{dagger_diffusion2021} & 2021 & IL + Diffusion & Imitation + diffusion policy & Multimodal action distributions for VLN-CE \\
\midrule
\multirow{6}{*}{\parbox{3.2cm}{\textbf{Long-Horizon\\Spatiotemporal}}}
 & HAMT \cite{history_aware2023} & 2023 & Transformer & History-aware self-attention & Full trajectory encoding via transformer sequence model \\
 & VLMaps \cite{visual_language_maps2023} & 2023 & VLM + Map & VLM embeddings on 3D map & Open-vocabulary spatial querying via language-grounded maps \\
 & GeoNav \cite{geonav2025} & 2025 & Hierarchical & Cognitive map + scene graph & Coarse-to-fine reasoning with dual spatial memory \\
 & TGDN \cite{target_grounded2023} & 2023 & Graph-Attn & Graph-attention on dialog & Dialog-grounded landmark navigation via structured history \\
 & TriVLA \cite{trivla2025} & 2025 & VLM + Diffusion & Video diffusion world model & Episodic prediction of future visual experiences \\
 & ReasonGrounder \cite{reasongrounder2025} & 2025 & 3DGS + VLM & Hierarchical 3D Gaussians & Adaptive multi-scale instruction grounding in 3D scenes \\
\midrule
\multirow{16}{*}{\parbox{3.2cm}{\textbf{Foundation Model-\\Driven Agentic}}}
 & FlightGPT \cite{flightgpt2024} & 2024 & VLM & CoT + GRPO & Interpretable CoT reasoning for city-scale target prediction \\
 & SkyVLN \cite{skyvln2025} & 2025 & LLM & Hierarchical semantic planning & LLM-driven sub-goal decomposition for aerial navigation \\
 & CityNavAgent \cite{citynavagent2025} & 2025 & VLM & Multi-scale scene reasoning & VLM cognitive core with topological map construction \\
 & LM-Nav \cite{lm_nav2022} & 2022 & VLM + LLM & Zero-shot composition & Training-free navigation combining VLM and LLM \\
 & OpenVLA \cite{openvla2024} & 2024 & VLA (7B) & Action tokenization + LoRA & Open-source generalizable VLA with efficient fine-tuning \\
 & $\pi_0$ \cite{pi0_2024} & 2024 & VLA Flow & Flow matching on VLM & Vision-Language-Action flow model for general robot control \\
 & GR00T N1 \cite{groot_n1_2025} & 2025 & VLA & Dual-system architecture & Open foundation model for generalist humanoid robots \\
 & NWM \cite{nwm2025} & 2025 & World Model & Controllable video generation & Navigation World Models predicting future visual observations \\
 & Cosmos-Reason1 \cite{cosmos_reason1_2025} & 2025 & WM + Reasoning & Physical common sense & Combining world models with embodied reasoning \\
 & GRaD-Nav++ \cite{grad_nav2025} & 2025 & VLA & MoE + Diff.\ RL in 3DGS & Lightweight onboard VLA at 25\,Hz real-time control \\
 & RaceVLA \cite{racevla2025} & 2025 & VLA & End-to-end FPV control & First VLA for high-speed autonomous drone racing \\
 & TrackVLA \cite{trackvla2025} & 2025 & VLA & Unified LLM backbone & Integrated target recognition and trajectory planning \\
 & NaVILA \cite{navila2025} & 2025 & Hierarchical & VLM planner + RL executor & Mid-level language actions bridging reasoning and control \\
 & Swarm-GPT \cite{swarm_gpt2023} & 2023 & LLM + Safety & LLM planner + safety filter & Language-based multi-UAV swarm coordination \\
 & TACOS \cite{tacos2025} & 2025 & LLM + MARL & LLM decomposition + RL exec & Hierarchical planning with safety-constrained execution \\
 & SwarmVLM \cite{swarmvlm2025} & 2025 & VLM-RAG & Real-time parameter tuning & VLM-RAG orchestration of heterogeneous UAV-AGV swarms \\
\bottomrule
\end{tabular}}
\caption{A comprehensive summary of representative UAV-VLN methods organized by methodological paradigm. For each method, we detail the core technical approach, the model type, and the key contribution to the field. VLM = Vision-Language Model; VLA = Vision-Language-Action Model; LLM = Large Language Model; RL = Reinforcement Learning; IL = Imitation Learning.}
\label{tab:summary_table}
\end{table*}

This review commences by examining the foundational paradigms in \textit{Modular and Early Learning Approaches}, where the integration of foundation models like Large Language Models (LLMs) and Vision-Language Models (VLMs) established powerful methods for hierarchical task decomposition and demonstrated the viability of language-guided aerial navigation through high-level semantic reasoning \cite{hierarchical_language_models2025,foundation_model2025}. To clarify the core distinctions among these foundational approaches, Table \ref{tab:comparison_table} provides a comparative summary of their key architectural and functional characteristics. The subsequent sections will systematically analyze each paradigm, highlighting their architectural innovations and methodological contributions to the field of UAV-VLN.
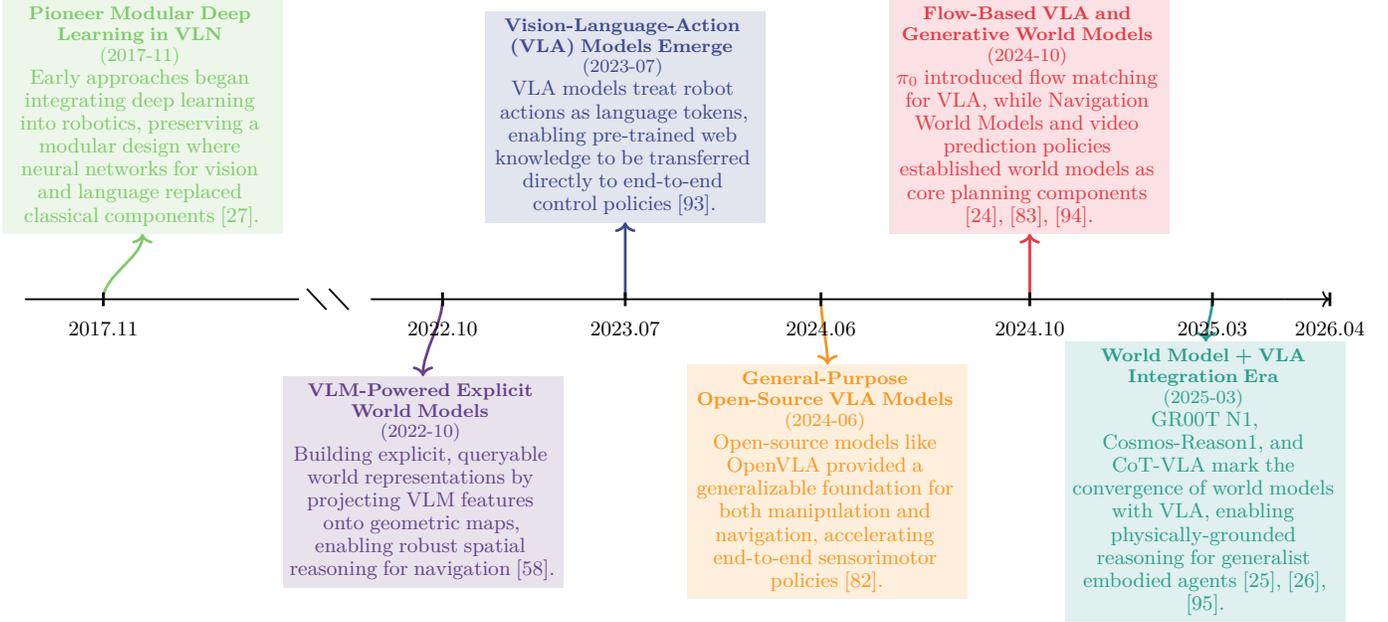
\begin{figure*}[h]
    \centering
    \begin{adjustbox}{max width=\textwidth, center}
    \begin{tikzpicture}[very thick, black]
        \small
        
        \definecolor{c0}{HTML}{7fc96b}
\definecolor{c1}{HTML}{66418a}
\definecolor{c2}{HTML}{3b498e}
\definecolor{c3}{HTML}{f8961e}
\definecolor{c4}{HTML}{e63946}
\definecolor{c5}{HTML}{2a9d8f}

        \tikzstyle{descript} = [text = black,align=center, minimum height=1.8cm, align=center, outer sep=0pt,font = \footnotesize]
        
        \coordinate (start_axis) at (0,0);
        \coordinate (break_left) at (4.20,0);
        \coordinate (break_right) at (5.30,0);
        \coordinate (end_axis) at (20.0,0);
        
        \coordinate (P0) at (1.20,0);
\coordinate (P1) at (6.40,0);
\coordinate (P2) at (9.20,0);
\coordinate (P3) at (12.20,0);
\coordinate (P4) at (15.40,0);
\coordinate (P5) at (18.20,0);

        \node[descript, fill=c0!15, text=c0](D0) at (1.80, 2.8) {
    \begin{minipage}{0.22\textwidth}
        \centering \textbf{Pioneer Modular Deep Learning in VLN} \\ (2017-11) \\ \small Early approaches began integrating deep learning into robotics, preserving a modular design where neural networks for vision and language replaced classical components \cite{vision_language_navigation2018}.
    \end{minipage} };
\node[descript, fill=c1!15, text=c1](D1) at (6.10, -2.8) {
    \begin{minipage}{0.22\textwidth}
        \centering \textbf{VLM-Powered Explicit World Models} \\ (2022-10) \\ \small Building explicit, queryable world representations by projecting VLM features onto geometric maps, enabling robust spatial reasoning for navigation \cite{visual_language_maps2023}.
    \end{minipage} };
\node[descript, fill=c2!15, text=c2](D2) at (9.20, 2.8) {
    \begin{minipage}{0.22\textwidth}
        \centering \textbf{Vision-Language-Action (VLA) Models Emerge} \\ (2023-07) \\ \small VLA models treat robot actions as language tokens, enabling pre-trained web knowledge to be transferred directly to end-to-end control policies \cite{rt2_2023}.
    \end{minipage} };
\node[descript, fill=c3!15, text=c3](D3) at (12.30, -2.8) {
    \begin{minipage}{0.22\textwidth}
        \centering \textbf{General-Purpose Open-Source VLA Models} \\ (2024-06) \\ \small Open-source models like OpenVLA provided a generalizable foundation for both manipulation and navigation, accelerating end-to-end sensorimotor policies \cite{openvla2024}.
    \end{minipage} };
\node[descript, fill=c4!15, text=c4](D4) at (15.40, 2.8) {
    \begin{minipage}{0.22\textwidth}
        \centering \textbf{Flow-Based VLA and Generative World Models} \\ (2024-10) \\ \small $\pi_0$ introduced flow matching for VLA, while Navigation World Models and video prediction policies established world models as core planning components \cite{pi0_2024, nwm2025, video_prediction_policy2025}.
    \end{minipage} };
\node[descript, fill=c5!15, text=c5](D5) at (18.10, -2.8) {
    \begin{minipage}{0.22\textwidth}
        \centering \textbf{World Model + VLA Integration Era} \\ (2025-03) \\ \small GR00T N1, Cosmos-Reason1, and CoT-VLA mark the convergence of world models with VLA, enabling physically-grounded reasoning for generalist embodied agents \cite{groot_n1_2025, cosmos_reason1_2025, cot_vla2025}.
    \end{minipage} };

        \path[->, color=c0] (P0) edge [out=90, in=-90, looseness=0.8] (D0);
\path[->, color=c1] (P1) edge [out=-90, in=90, looseness=0.8] (D1);
\path[->, color=c2] (P2) edge [out=90, in=-90, looseness=0.8] (D2);
\path[->, color=c3] (P3) edge [out=-90, in=90, looseness=0.8] (D3);
\path[->, color=c4] (P4) edge [out=90, in=-90, looseness=0.8] (D4);
\path[->, color=c5] (P5) edge [out=-90, in=90, looseness=0.8] (D5);

        \draw[thick] (start_axis) -- (break_left);
        \draw[thick] (break_right) -- (19.3,0);
        \draw[->, thick] (19.3,0) -- (end_axis);
        \draw[thick] ($(break_left)+(0.12,0.16)$) -- ($(break_left)+(0.42,-0.16)$);
        \draw[thick] ($(break_left)+(0.46,0.16)$) -- ($(break_left)+(0.76,-0.16)$);
        
        \foreach \i/\j in {1.20/2017.11,6.40/2022.10,9.20/2023.07,12.20/2024.06,15.40/2024.10,18.20/2025.03,20.00/2026.04} {\draw (\i cm,3pt) -- (\i cm,-3pt) node[below=3pt] {\j};}

    \end{tikzpicture}
    \end{adjustbox}
    \caption{A timeline of the methodological evolution in UAV Vision-and-Language Navigation (UAV-VLN). The figure charts the progression from early modular deep learning systems to VLM-powered world models, the emergence of VLA models, the development of open-source VLA foundations, the rise of flow-based VLA and generative world models, and culminating in the current era of world model and VLA integration for physically-grounded embodied reasoning.}
    \label{fig:timeline_fig_2}
\end{figure*}

\begin{table*}[!htb]
\centering
\resizebox{\textwidth}{!}{%
\begin{tabular}{p{3.8cm}p{4.5cm}p{6cm}p{5cm}}
\toprule
\textbf{Paradigm} & \textbf{Sub-Category} & \textbf{Core Approach} & \textbf{Characteristics \& Limitations} \\
\midrule
\multirow{3}{*}{\parbox{3.8cm}{\textbf{Modular \& Early\\Learning Approaches}}}
 & Classical Decomposed Robotics Pipelines & Decomposes navigation into discrete stages: perception (SLAM), planning (A*, RRT), and control (sense-think-act). & High interpretability; suffers from brittleness and cascading error propagation. \\
\cmidrule(l){2-4}
 & Modular Deep Learning Architectures & Replaces classical modules with CNNs (vision) and RNNs (language); uses cross-modal attention for fusion. & More robust perception; creates information bottlenecks and struggles with complex reasoning. \\
\cmidrule(l){2-4}
 & Emergence of End-to-End Learning Policies & Learns direct sensor-to-action mapping via DRL or Imitation Learning, bypassing explicit intermediate representations. & Conceptually simpler; challenged by sample inefficiency and long-term credit assignment. \\
\midrule
\multirow{3}{*}{\parbox{3.8cm}{\textbf{Architectures for\\Long-Horizon\\Spatiotemporal\\Understanding}}}
 & Temporal Transformers & Treats navigation as a sequence modeling problem; uses transformer self-attention to encode full trajectory history. & Enables direct access to past events and improves long-term memory over recurrent models. \\
\cmidrule(l){2-4}
 & Visual Language Maps & Reframes navigation as spatial reasoning; fuses VLM features with geometric reconstructions for queryable maps. & Grounds language in the physical environment; high computational cost and sensitivity to odometry errors. \\
\cmidrule(l){2-4}
 & Cognitive Maps \& Scene Graphs & Employs multi-scale spatial memories with global cognitive maps and local scene graphs for coarse-to-fine reasoning. & Mimics human navigation strategies; computationally complex in large-scale environments. \\
\midrule
\multirow{5}{*}{\parbox{3.8cm}{\textbf{Foundation Model-\\Driven Agentic\\Systems}}}
 & VLMs as Cognitive Core & Leverages VLMs as deliberative planners using Chain-of-Thought (CoT) to decompose commands into executable sub-goals. & Interpretable planning; latency overhead and susceptibility to hallucination. \\
\cmidrule(l){2-4}
 & VLA Models as End-to-End Policies & Maps multimodal perception directly to low-level motor commands; treats actions as tokens in a VLM vocabulary. & Generalizable end-to-end control; computationally expensive and brittle for long-horizon tasks. \\
\cmidrule(l){2-4}
 & World Model + VLA Integration & Combines generative world models (e.g., video prediction) with VLA policies for physically-grounded reasoning. & Enables robust planning and open-world generalization; requires massive training data and compute. \\
\cmidrule(l){2-4}
 & Hierarchical \& Hybrid Agentic Systems & Combines a deliberative VLM planner (``cerebrum'') with a reactive low-level controller (``cerebellum''). & Balances reasoning and real-time demands; improves safety and transparency. \\
\cmidrule(l){2-4}
 & Multi-UAV Swarm Coordination & Uses MLLMs as a centralized ``swarm brain'' to interpret missions and coordinate collaborative multi-agent behaviors. & Integrates strategic MLLM planning with MARL-based tactical execution. \\
\bottomrule
\end{tabular}}
\caption{A methodological taxonomy of UAV Vision-and-Language Navigation (UAV-VLN) agents, organized into three main paradigms: Modular and Early Learning Approaches, Architectures for Long-Horizon Spatiotemporal Understanding, and Foundation Model-Driven Agentic Systems. Each paradigm is further decomposed into its constituent sub-categories with their core approaches and key characteristics.}
\label{tab:taxonomy_tree}
\end{table*}

\subsection{Modular and Early Learning Approaches} \label{subsec:Modular and Early Learning Approaches}

Early approaches to UAV-VLN established a foundational problem decomposition into perception, cognition, and control, evolving from explicit classical pipelines to modular deep learning systems. This evolution was guided by the sense-think-act paradigm, where functionally distinct modules handle different stages of the navigation task. While these initial methods proved the feasibility of language-guided aerial navigation, the inherent information bottlenecks and limited adaptability of their decoupled architectures motivated the subsequent shift toward more integrated, end-to-end models~\cite{uav_agentic2025, vision_language_navigation2024}. This section traces this progression, visually summarized in \autoref{fig:tiny_tree_figure_0}, which illustrates the evolution of early learning approaches categorized into three primary paradigms. It begins with Classical Decomposed Robotics Pipelines that separate tasks into discrete stages, progresses to Modular Deep Learning Architectures where classical components are replaced by neural networks, and culminates in the Emergence of End-to-End Learning Policies that learn direct sensor-to-action mappings.

\subsubsection{Classical Decomposed Robotics Pipelines}
The first generation of language-guided UAV systems adapted classical robotics pipelines, decomposing the navigation task into discrete, algorithmically-solved stages of environmental mapping, path planning, and flight control, which offered high interpretability at the cost of adaptability in dynamic environments. This approach embodies the sense-think-act loop, where the agent's behavior is the product of a sequence of independent modules. As conceptualized in early frameworks~\cite{uav_agentic2025}, this can be represented as a layered architecture:
\begin{equation}
o_t = \phi(s_t), \quad a_t = \pi(g, o_t), \quad u_t = \psi(a_t, x_t)
\end{equation}
where an observation $o_t$ is extracted from the state $s_t$ by a perception module $\phi$; a high-level action $a_t$ is determined by a planner $\pi$ based on the goal $g$ and observation; and a low-level controller $\psi$ generates control commands $u_t$ to execute the action given the current UAV state $x_t$. The perception stage often relied on building an explicit world model, using techniques like Simultaneous Localization and Mapping (SLAM) to create geometric maps~\cite{uav_navigation2019, onboard_slam2023}, with some systems tightly coupling SLAM-estimated states with low-level controllers~\cite{simultaneous_control_uavs2024}. Others focused on constructing semantic and topological maps to ground linguistic concepts~\cite{goal_oriented_semantic_exploration2020, visual_features2024}. These explicit world models can be categorized into spatial grid maps, topological maps, and dense geometric maps, each offering different trade-offs in representing semantic information~\cite{semantic_mapping2025}. Based on this world model, a planner would compute an optimal trajectory using a wide array of algorithmic methods, including graph-based search (e.g., A*, D*)~\cite{optimizing_uav2024}, sampling-based algorithms (e.g., Rapidly-exploring Random Trees, RRT)~\cite{unmanned_aerial_vehicle_path_planning, uav_path2023}, potential field methods~\cite{uav_trajectory2025}, and other optimization techniques for multi-agent systems~\cite{hierarchical_multi_uav2024, multi_uav_formation2024}. The primary advantage of this modularity was its transparency, as each component, such as a visual servoing module for tracking combined with a separate obstacle avoidance system~\cite{gps_denied_ibvs2025}, could be debugged independently. However, these systems were inherently brittle; high latency and computational complexity coupled with errors in the initial perception stages could propagate catastrophically, leading to failures in planning and control~\cite{path_planning2022, visual_features2024}.

\subsubsection{Modular Deep Learning Architectures}
The initial integration of deep learning preserved the modular system design, replacing classical components with specialized neural networks for vision and language encoding, where cross-modal fusion mechanisms served as the primary bridge for translating multimodal understanding into navigational decisions. This paradigm shift addressed the brittleness of hand-crafted feature extractors by leveraging deep learning for more robust perception and semantic grounding~\cite{uav_recognition2024, deep_multimodal_fusion2024}. In a typical architecture, a visual encoder, such as a Convolutional Neural Network (CNN), processes camera imagery, while a language encoder, like a Recurrent Neural Network (RNN), processes the textual instruction. The outputs of these encoders are then fused to inform a decision-making module, a structure pioneered in ground-based VLN using sequence-to-sequence models with attention~\cite{vision_language_navigation2018}. Other models explicitly decomposed the task into neural sub-modules for goal prediction and action generation~\cite{mapping_instructions2024}. A common fusion strategy is cross-modal attention, which allows the model to dynamically weigh the importance of different visual regions based on the linguistic query~\cite{vlca2026, real_time_multi_modal2022}, part of a broader family of fusion techniques including early, late, and hybrid approaches~\cite{common_practices2026}. Some systems demonstrated a hybrid approach, using an RNN-based Natural Language Understanding (NLU) module to parse commands that would then parameterize a traditional planner like RRT~\cite{uav_path2023}. The baseline models evaluated in the AerialVLN benchmark, such as Cross-Modal Attention (CMA) and Look-ahead Guidance (LAG), exemplify this modular deep learning approach~\cite{aerialvln2023}. However, their low success rates (e.g., 5.1\% SR for LAG) highlighted a critical flaw: simply replacing classical modules with neural networks created significant information bottlenecks, struggled with integrating fine-grained vision-language features~\cite{vl_nav2025}, and was insufficient for the complex, long-horizon reasoning required in aerial navigation.

\subsubsection{Emergence of End-to-End Learning Policies}
Early end-to-end models, trained primarily via reinforcement and imitation learning, established the viability of learning a direct mapping from raw perceptual inputs to flight commands, thereby bypassing explicit intermediate representations like geometric maps and symbolic planners. This approach posits that intermediate representations are suboptimal and that a policy can learn to implicitly perform localization, planning, and control by mapping sensor streams directly to actions~\cite{model_free_uav2024}. These methods train a single neural network, often a CNN-RNN architecture that can process sequences of observations and maintain a memory of past states~\cite{memory_based_deep2018}, using either Deep Reinforcement Learning (DRL) or Imitation Learning (IL). In DRL-based approaches, the agent learns a visuomotor policy by maximizing a reward signal through trial and error, with algorithms like DQN, SARSA, and A2C being benchmarked for tasks such as obstacle avoidance~\cite{drone_obstacle_avoidance, explainable_deep_reinforcement2021, reinforcement_learning_uav2023}. Some of these systems retained a degree of modularity, with the DRL agent outputting high-level waypoints to a separate, native flight controller~\cite{path_planning_algorithm}. In contrast, IL methods train the policy to mimic expert trajectories, learning a compact representation for robust navigation from demonstrations~\cite{vision_based_2d_navigation}. While conceptually simpler and potentially more optimal by avoiding human-designed bottlenecks, these early end-to-end models faced significant challenges in sample efficiency and long-term credit assignment. Their initial successes and inherent limitations, alongside the evolution of more powerful backbones like Vision Transformers (ViTs) for control~\cite{vision_transformers2025}, highlighted a clear need for architectures with stronger reasoning capabilities, paving the way for the foundation models discussed in subsequent sections.

\definecolor{color_root}{HTML}{bebada}
\definecolor{color_node}{HTML}{b3de69}
\definecolor{color_list}{HTML}{fccde5}

\begin{figure*}[h]
   \centering
   \begin{adjustbox}{max width=\textwidth, max height=0.5\textheight, center}
      \begin{tikzpicture}[
         start chain = going below,
         node distance=1mm,
         sibling distance=40mm,
         edge from parent/.style={->,draw},
         >=latex,
         basic/.style  = {
            draw, text width=1cm, drop shadow, rectangle
         },
         root/.style   = {
            basic, rounded corners=2pt, 
            thin, align=center,text width=15em, 
            font=\small\bfseries,
            fill=blue!20
         },
         nodeB/.style = {
            basic, rounded corners=5pt, 
            thin,align=center, fill=green!20,
            font=\scriptsize,
            text width=8em
         },
         nodeL/.style = {
            basic, thin, 
            align=center, 
            font=\scriptsize\itshape, fill=pink!20, text width=9em
         }
      ]
   
\node[root, fill=color_root!60] {Evolution of Modular and Early Learning Approaches} child {node[nodeB, fill=color_node!40] (leaf_3) {Classical Decomposed Robotics Pipelines}}child {node[nodeB, fill=color_node!40] (leaf_4) {Modular Deep Learning Architectures}}child {node[nodeB, fill=color_node!40] (leaf_5) {Emergence of End-to-End Learning Policies}} ;
\begin{scope}[start chain=3 going below]
\chainin (leaf_3) [on chain, join];
\node[nodeL, on chain, fill=color_list!15, xshift=6mm, yshift=-5mm] (l30) {Sense-think-act paradigm\cite{uav_agentic2025}};
\node[nodeL, on chain, fill=color_list!15] (l31) {Explicit world models};
\node[nodeL, on chain, fill=color_list!15] (l32) {SLAM for mapping\cite{uav_navigation2019, simultaneous_control_uavs2024}};
\node[nodeL, on chain, fill=color_list!15] (l33) {Semantic maps\cite{goal_oriented_semantic_exploration2020, semantic_mapping2025}};
\node[nodeL, on chain, fill=color_list!15] (l34) {Algorithmic path planning};
\node[nodeL, on chain, fill=color_list!15] (l35) {Graph-based search\cite{optimizing_uav2024}};
\node[nodeL, on chain, fill=color_list!15] (l36) {Sampling-based RRT\cite{uav_path2023}};
\node[nodeL, on chain, fill=color_list!15] (l37) {High interpretability};
\node[nodeL, on chain, fill=color_list!15] (l38) {Error propagation issues\cite{path_planning2022, visual_features2024}};

\end{scope}
\draw[->] ($(leaf_3.south west) + (2mm, 0)$) |- (l30.west);
\draw[->] ($(leaf_3.south west) + (2mm, 0)$) |- (l31.west);
\draw[->] ($(leaf_3.south west) + (2mm, 0)$) |- (l32.west);
\draw[->] ($(leaf_3.south west) + (2mm, 0)$) |- (l33.west);
\draw[->] ($(leaf_3.south west) + (2mm, 0)$) |- (l34.west);
\draw[->] ($(leaf_3.south west) + (2mm, 0)$) |- (l35.west);
\draw[->] ($(leaf_3.south west) + (2mm, 0)$) |- (l36.west);
\draw[->] ($(leaf_3.south west) + (2mm, 0)$) |- (l37.west);
\draw[->] ($(leaf_3.south west) + (2mm, 0)$) |- (l38.west);

\begin{scope}[start chain=4 going below]
\chainin (leaf_4) [on chain, join];
\node[nodeL, on chain, fill=color_list!15, xshift=6mm, yshift=-5mm] (l40) {Neural network modules};
\node[nodeL, on chain, fill=color_list!15] (l41) {Specialized encoders\cite{uav_recognition2024}};
\node[nodeL, on chain, fill=color_list!15] (l42) {Cross-modal fusion\cite{deep_multimodal_fusion2024}};
\node[nodeL, on chain, fill=color_list!15] (l43) {Attention mechanisms\cite{vision_language_navigation2018, real_time_multi_modal2022}};
\node[nodeL, on chain, fill=color_list!15] (l44) {Hybrid fusion types\cite{common_practices2026}};
\node[nodeL, on chain, fill=color_list!15] (l45) {Goal prediction modules\cite{mapping_instructions2024}};
\node[nodeL, on chain, fill=color_list!15] (l46) {AerialVLN baselines\cite{aerialvln2023}};
\node[nodeL, on chain, fill=color_list!15] (l47) {Information bottlenecks\cite{vl_nav2025}};

\end{scope}
\draw[->] ($(leaf_4.south west) + (2mm, 0)$) |- (l40.west);
\draw[->] ($(leaf_4.south west) + (2mm, 0)$) |- (l41.west);
\draw[->] ($(leaf_4.south west) + (2mm, 0)$) |- (l42.west);
\draw[->] ($(leaf_4.south west) + (2mm, 0)$) |- (l43.west);
\draw[->] ($(leaf_4.south west) + (2mm, 0)$) |- (l44.west);
\draw[->] ($(leaf_4.south west) + (2mm, 0)$) |- (l45.west);
\draw[->] ($(leaf_4.south west) + (2mm, 0)$) |- (l46.west);
\draw[->] ($(leaf_4.south west) + (2mm, 0)$) |- (l47.west);

\begin{scope}[start chain=5 going below]
\chainin (leaf_5) [on chain, join];
\node[nodeL, on chain, fill=color_list!15, xshift=6mm, yshift=-5mm] (l50) {Direct sensor-to-action mapping\cite{model_free_uav2024}};
\node[nodeL, on chain, fill=color_list!15] (l51) {Bypass intermediate representations};
\node[nodeL, on chain, fill=color_list!15] (l52) {Deep Reinforcement Learning (DRL)\cite{drone_obstacle_avoidance}};
\node[nodeL, on chain, fill=color_list!15] (l53) {Imitation Learning (IL)\cite{vision_based_2d_navigation}};
\node[nodeL, on chain, fill=color_list!15] (l54) {CNN-RNN architectures\cite{memory_based_deep2018}};
\node[nodeL, on chain, fill=color_list!15] (l55) {Sample efficiency challenges};
\node[nodeL, on chain, fill=color_list!15] (l56) {Paved way for ViTs\cite{vision_transformers2025}};
\node[nodeL, on chain, fill=color_list!15] (l57) {Waypoint generation\cite{path_planning_algorithm}};

\end{scope}
\draw[->] ($(leaf_5.south west) + (2mm, 0)$) |- (l50.west);
\draw[->] ($(leaf_5.south west) + (2mm, 0)$) |- (l51.west);
\draw[->] ($(leaf_5.south west) + (2mm, 0)$) |- (l52.west);
\draw[->] ($(leaf_5.south west) + (2mm, 0)$) |- (l53.west);
\draw[->] ($(leaf_5.south west) + (2mm, 0)$) |- (l54.west);
\draw[->] ($(leaf_5.south west) + (2mm, 0)$) |- (l55.west);
\draw[->] ($(leaf_5.south west) + (2mm, 0)$) |- (l56.west);
\draw[->] ($(leaf_5.south west) + (2mm, 0)$) |- (l57.west);

\end{tikzpicture}
\end{adjustbox}

\caption{This figure illustrates the evolution of early learning approaches in UAV-VLN, categorized into three primary paradigms. It begins with Classical Decomposed Robotics Pipelines that separate tasks into discrete stages, progresses to Modular Deep Learning Architectures where classical components are replaced by neural networks, and culminates in the Emergence of End-to-End Learning Policies that learn direct sensor-to-action mappings.}
\label{fig:tiny_tree_figure_0}

\end{figure*}
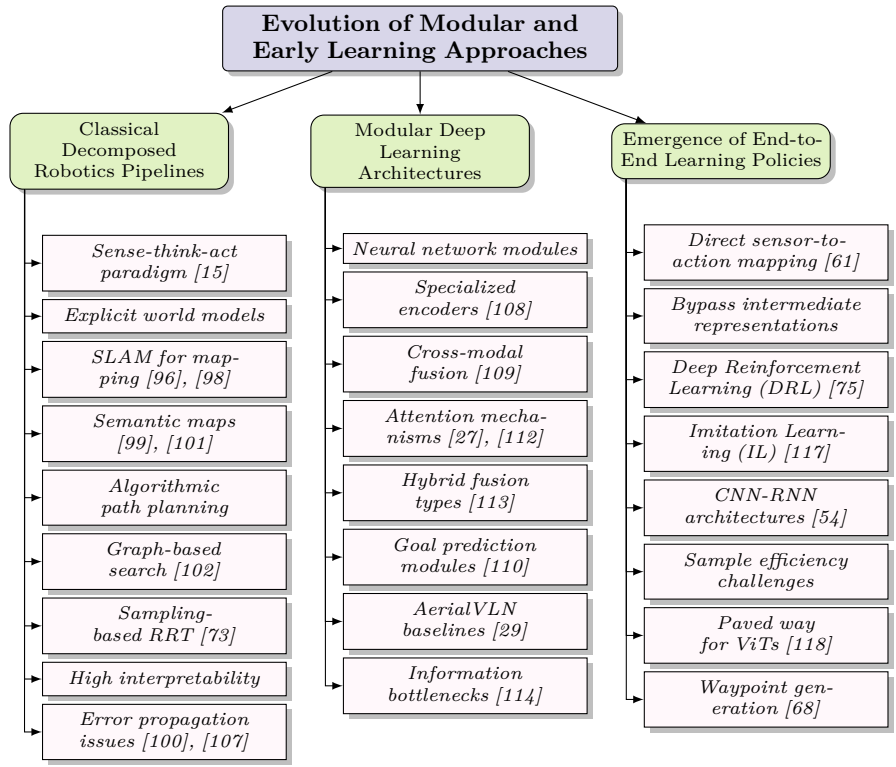
\subsection{Architectures for Long-Horizon Spatiotemporal Understanding} \label{subsec:Architectures for Long-Horizon Spatiotemporal Understanding}

Effective long-horizon spatiotemporal understanding in UAV-VLN requires architectures that move beyond simple recurrent memory, which is often insufficient for retaining critical information over extended trajectories. The field has evolved toward two primary paradigms: sophisticated temporal history encoding, which treats navigation as a sequence modeling problem, and the construction of explicit, semantically-rich world models, which reframes it as a spatial reasoning task.
\begin{figure*}[!htb]
    \centering
    \includegraphics[width=0.9\textwidth,height=0.5\textheight,keepaspectratio]{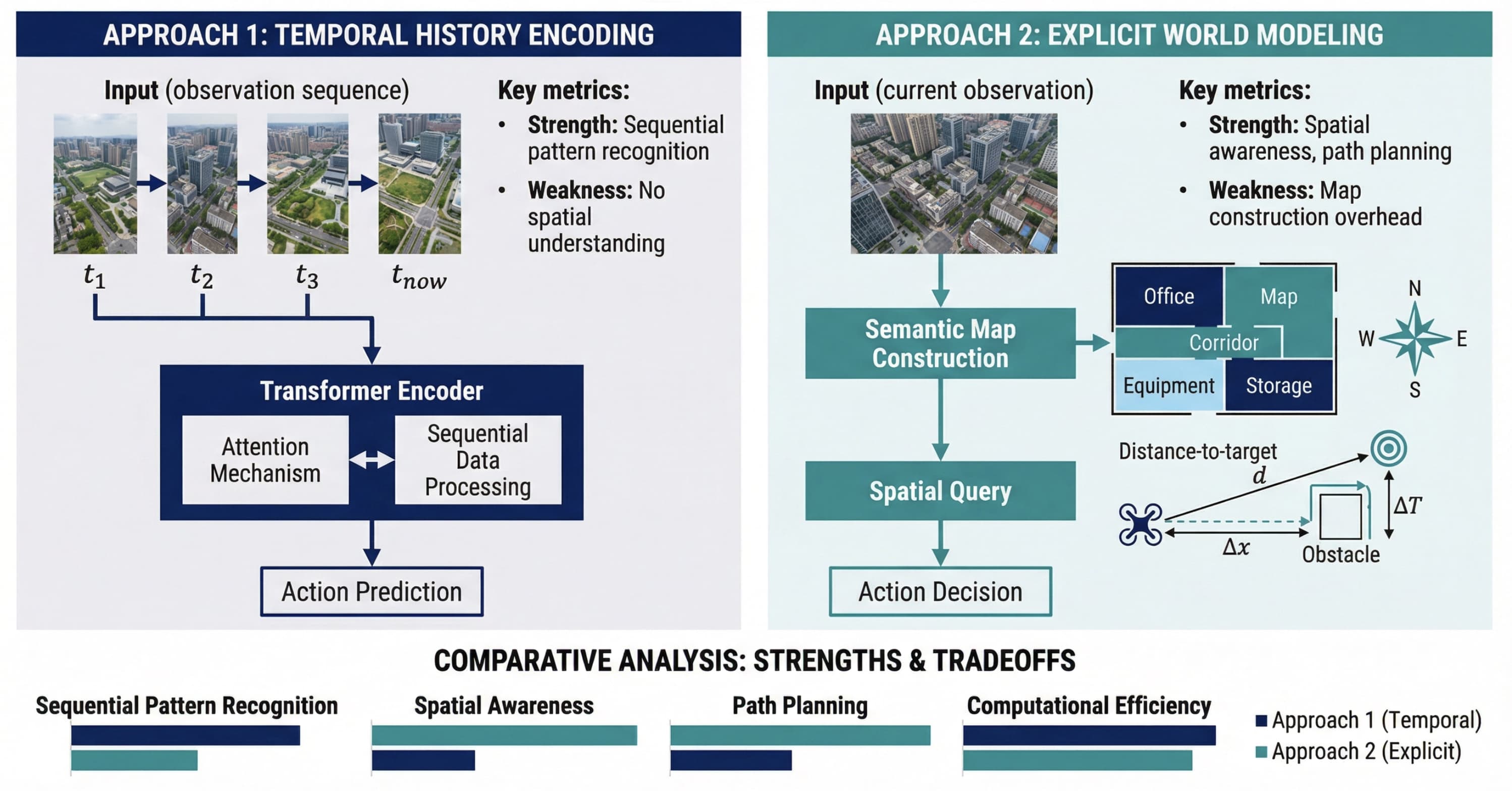}
    \caption{Two primary architectural paradigms for long-horizon navigation. Temporal history encoding (left) treats the task as a sequence modeling problem, using transformers to reason over past observations and actions, while explicit world modeling (right) reframes it as a spatial reasoning task by constructing a queryable semantic map of the environment.}
    \label{fig:a_methodological_taxonomy_of_uav-vln_agents_auto_1}
\end{figure*}
 These architectural philosophies address the core challenge of managing and reasoning over accumulated spatiotemporal information by building multi-scale spatial representations, such as global cognitive maps and local scene graphs, which support coarse-to-fine reasoning and enable agents, particularly UAVs, to handle complex instructions in urban environments that reference distant landmarks or require backtracking \cite{geonav2025,citynavagent2025,reasoning_unseen2024,foundation_models2025,navagent_urban2024}

\subsubsection{Encoding Long-Term Dependencies with Temporal Transformers}
To overcome the information decay inherent in recurrent models over long trajectories, a primary architectural approach is to explicitly encode an agent's observation and action history using transformer-based sequence models. The core problem addressed by this line of work is the information bottleneck of RNNs, such as LSTMs with temporal attention, which struggle to retain salient details from early in a long trajectory~\cite{memory_based_deep2018}. Early learning models sought to manage complexity by decomposing the task, for instance by using one module for goal prediction and another RNN for action generation~\cite{mapping_instructions2024}, or by using cross-modal attention to generate waypoints from panoramic observations~\cite{waypoint_models2021}. The key architectural shift, exemplified by the History Aware Multimodal Transformer (HAMT), is to treat the entire trajectory history of panoramic observations and actions as a unified token sequence~\cite{history_aware2023}. This allows the model to use self-attention mechanisms to directly access and reason about past events, irrespective of their temporal distance.

This paradigm has been extended with more sophisticated memory and modeling structures. For instance, some frameworks use a dual-branch architecture to dynamically capture sequential scene features while robustly extracting navigational cues from language~\cite{dual_branch2025}. Others integrate short-term perceptual blurring with long-term retrieval to maintain coherence over extended tasks~\cite{long_horizon_vln2025}. A forward-looking approach involves building episodic world models using video diffusion models, enabling the agent to not only recall but also predict future sequential experiences~\cite{trivla2025}. The policy is typically trained to predict the next action $a_t$ given the instruction $W$, panoramic observation history $H_t$, and current visual inputs $O_t, O_t^c$ by minimizing the negative log-likelihood:
\begin{equation}
\mathcal{L}(\theta) = -\log p(a_t|W, H_t, O_t, O_t^c; \theta)
\end{equation}
where $\theta$ represents the model parameters. This approach improves performance on tasks requiring long-term memory, such as referencing distant landmarks. Supporting works also leverage transformers to process historical context, for instance by using graph-attention mechanisms to associate dialog with structured historical observations for improved landmark grounding~\cite{target_grounded2023} or by processing dialog history to predict waypoints and human attention~\cite{aerial_vision_dialog2024}.

\subsubsection{Building Explicit World Representations with Visual Language Maps}
To ground language commands in the physical environment for precise localization, agents are increasingly designed to construct explicit, semantically-rich spatial maps by fusing features from pre-trained Vision-Language Models (VLMs) with geometric reconstructions. This approach argues that long-horizon navigation is not just a memory problem but a spatial reasoning problem that benefits from an externalized, persistent world model. The core technical route, demonstrated by VLMaps~\cite{visual_language_maps2023}, involves projecting pixel-level embeddings from a VLM onto a 3D point cloud or top-down grid map built during exploration. This creates a queryable representation where locations can be indexed by open-vocabulary natural language. A range of world models are used, each presenting different trade-offs between computational cost and informational richness~\cite{air_ground_robots2025}. These include lightweight semantic octrees~\cite{semantic_octree_mapping}, allocentric semantic maps built via cross-modality supervision~\cite{real_time_multi_modal2022}, and geographic-semantic representations for large-scale urban navigation~\cite{citynav2025}.

The quality of these explicit representations is enhanced by integrating fine-grained, pixel-wise vision-language features~\cite{vl_nav2025} and fusing multimodal data sources, including depth information and SLAM map data, to improve positional accuracy~\cite{multimodal_pretrained_knowledge2023}. These semantically rich maps are not just passive data stores but active components in the decision-making process. For example, the VLM-HLP framework leverages VLMs to infer structural patterns and affordances from incomplete maps, enabling more efficient and intelligent high-level planning~\cite{vision_language_models2025}. Similarly, some methods construct a semantic traversable map from historical observations and use 3D feature fields to predict novel views, enhancing perception from a limited field of view~\cite{sim_to_real2024}. This VLM-based spatial understanding can also guide classical planners, as seen in hybrid frameworks that use VLMs to dynamically steer the sampling process of algorithms like RRT, improving both efficiency and path quality.

\subsubsection{Hierarchical Reasoning with Cognitive Maps and Scene Graphs}
For efficient navigation in large-scale outdoor environments, state-of-the-art architectures build hierarchical spatial memories, such as cognitive maps and scene graphs, to support multi-stage, coarse-to-fine reasoning that mimics human navigation strategies. A single, flat semantic map can become computationally intractable for navigating expansive environments. To overcome this, agents like GeoNav construct two representations: a global cognitive map for coarse landmark navigation and a local hierarchical scene graph for precise target search~\cite{geonav2025}. This dual structure facilitates a three-phase reasoning process landmark navigation, target search, and precise localization that efficiently prunes the search space. This principle is extended in methods that use 3D scene graphs with metric and semantic edges to resolve complex spatial queries~\cite{open_vocabulary_object_grounding} or build 3D Hierarchical Scene Graphs (HSGs) integrated with an LLM for complex indoor spatial reasoning~\cite{open_vocabulary_indoor2024}.

The frontier of this approach lies in creating more detailed, interactive, and physically-aware representations. Recent work leverages 3D Gaussian Splatting to build hierarchical feature fields, enabling adaptive grouping of environmental features based on physical scale for precise instruction grounding~\cite{reasongrounder2025}. This technique also supports the creation of executable 3D models that decouple object appearance from physics, allowing for photorealistic rendering alongside accurate collision detection and interaction simulation~\cite{executable_3d_gaussian2024}. This structured approach aligns with the trend toward agentic architectures that separate high-level strategic planning from low-level action execution~\cite{uav_agentic2025}. In such systems, agents perform multi-stage missions by inferring object roles from monocular images~\cite{morphonavi2025} or dynamically construct virtual graphs to reason about paths in obstructed environments~\cite{navigating_beyond2024}. This indicates a clear shift from monolithic decision-makers to more modular and cognitively plausible systems.

\subsection{Foundation Model-Driven Agentic Systems} \label{subsec:Foundation Model-Driven Agentic Systems}

The advent of large foundation models has catalyzed a paradigm shift in UAV-VLN, moving the field toward integrated agentic systems that unify perception, reasoning, and action within a single cognitive architecture~\cite{uav_agentic2025, multimodal_spatial_reasoning2025}. This evolution is characterized by the development of agents capable of goal-directed reasoning, planning, and adaptation, driven by pre-trained models that leverage vast datasets to achieve unprecedented generalization~\cite{vision_language_action_models, vision_language_navigation2024, large_model_embodied_ai2025, embodied_ai2024, embodied_multimodal2025}. The integration of these models follows two primary technical routes, which are often complementary.

\begin{figure*}[!htb]
    \centering
    \includegraphics[width=0.9\textwidth,height=0.5\textheight,keepaspectratio]{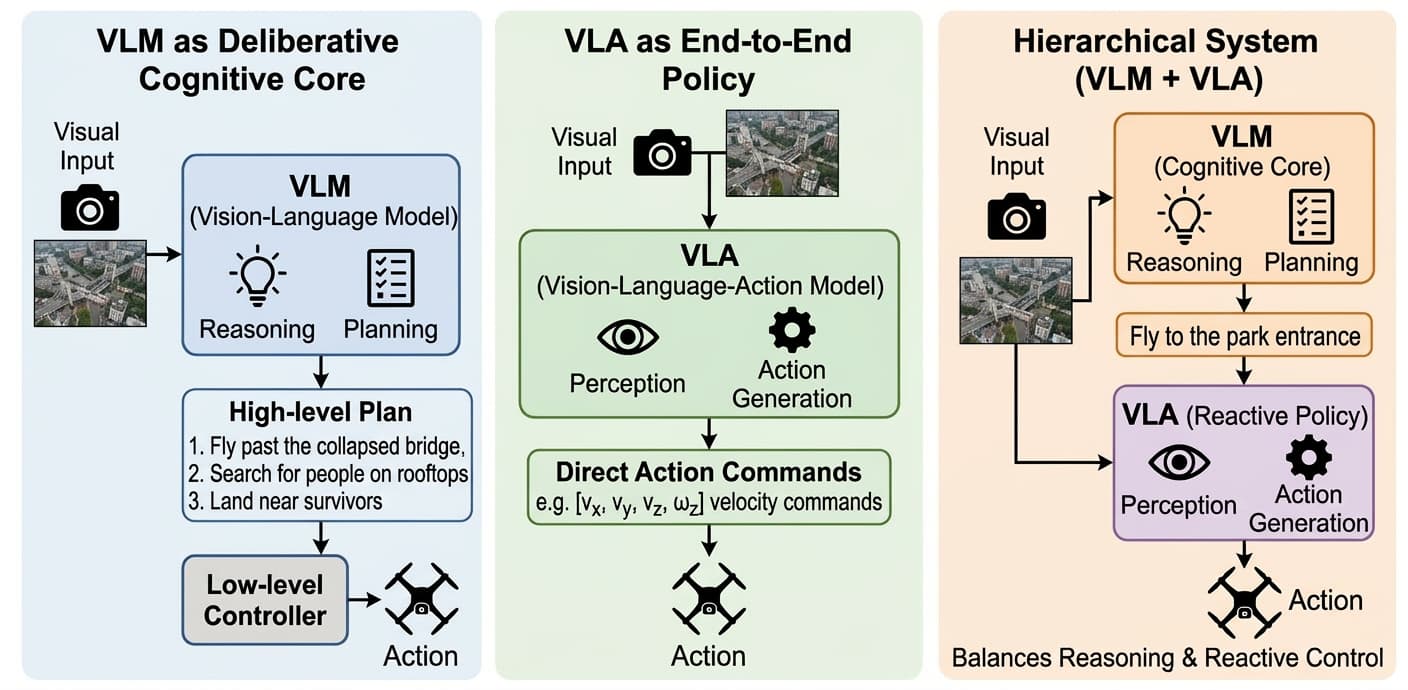}
    \caption{Architectural paradigms for foundation model-driven agents. VLMs can act as a deliberative cognitive core to generate high-level plans (left), VLAs can serve as end-to-end policies mapping perception directly to action (center), or they can be combined in a hierarchical system that balances reasoning and reactive control (right).}
    \label{fig:a_methodological_taxonomy_of_uav-vln_agents_auto_2}
\end{figure*}
 The first leverages Vision-Language Models (VLMs) as a high-level cognitive core for reasoning and planning~\cite{geospatial_reasoning2023}, while the second employs Vision-Language-Action (VLA) models as end-to-end sensorimotor policies that directly generate control commands~\cite{embodied_ai2024T}. This pursuit ultimately aims to create more autonomous and sociable "companion" UAVs that can engage in complex, long-horizon tasks through natural human interaction~\cite{companion_uavs2020, human_drone2025, spatial_assisted2024}. A critical aspect of this paradigm is the development of efficient adaptation methods, such as prompt engineering and parameter-efficient fine-tuning (PEFT) techniques like LoRA, which enable large pre-trained models to be specialized for UAV-specific tasks without costly full retraining~\cite{satellite_image2025, nuplanqa2025, large_language_models2024}. This section details these distinct approaches, examines their architectural integration, and explores how they scale to orchestrate multi-UAV swarms.

\subsubsection{VLMs as a Cognitive Core for High-Level Reasoning and Planning}
The primary role of foundation models in many agentic UAVs is to serve as a cognitive core, using techniques like Chain-of-Thought (CoT) to translate ambiguous natural language instructions into explicit, interpretable navigation plans~\cite{embodied_navigation2024}. This approach treats the VLM not as a direct controller but as a deliberative planner that reasons about the task and environment to produce a series of high-level goals~\cite{llm_robotic_autonomy2025}. For instance, FlightGPT exemplifies this by employing a two-stage training pipeline and a CoT prompting mechanism to decompose a complex command into a sequence of logical steps, culminating in a precise target prediction for a separate navigation module~\cite{flightgpt2024}. The powerful reasoning capabilities of these models are increasingly being benchmarked for complex spatial understanding tasks~\cite{open3d_vqa2024, rs_gpt4v2024}, and their effectiveness is built upon powerful alignment techniques, primarily contrastive learning, which maps vision and language into a shared embedding space~\cite{geospatial_foundation2025, vision_language_models2024K2024, vision_language_models2023}. This is often optimized using an InfoNCE loss function:
\begin{equation}
\mathcal{L} = -\log\frac{\exp(s_{i,i}/\tau)}{\sum_{j=1}^N \exp(s_{i,j}/\tau)}
\end{equation}
where $s_{i,j}$ is the cosine similarity between the embeddings of image $i$ and text $j$, and $\tau$ is a temperature parameter~\cite{vision_language_models2024K2024}. This paradigm extends robustly to specialized domains like remote sensing, where a growing ecosystem of Remote Sensing Foundation Models (RSFMs) has emerged, categorized into visual, vision-language, and other generative models~\cite{foundation_models_survey}. These models handle high-resolution geospatial imagery via task-driven resolution adjustment~\cite{geomag2025}, dynamic token pruning~\cite{vision_language_model2025}, and unified architectures that process image, region, and pixel-level tasks simultaneously~\cite{falcon2025, omnigeo2025, remote_sensing_vision_language2025, urban_monitoring2025}, with some employing semantic-augmented alignment and expert modeling to enrich visual features~\cite{remote_sensing_lvlm2025}. Novel training strategies have also emerged to adapt internet-scale VLMs to these domains without task-specific annotations, for example, by using ground-level images as a semantic bridge to satellite imagery~\cite{remote_sensing2023} or through multi-stage, prompt-based fine-tuning~\cite{vision_language_models2024S2024}.

This paradigm of using VLMs for high-level cognition has been widely adopted, with a series of methods leveraging LLMs for hierarchical semantic planning and decomposing long-horizon navigation tasks into manageable sub-goals~\cite{citynavagent2025, skyvln2025, reasoning_unseen2024}. Some frameworks construct multi-scale environmental representations, using VLMs for deductive scene reasoning and to synthesize global topological maps with local landmark recognition for robust decision-making~\cite{navagent_urban2024, reasongrounder2025, open_vocabulary_object_grounding}. A notable trend is the rise of training-free frameworks that leverage the zero-shot capabilities of pre-trained VLMs; systems such as LM-Nav, EzReal combine VLMs and LLMs to enable navigation without any fine-tuning~\cite{lm_nav2022, uav_vla2025, semantic_scene2024, zeng2025ezreal}. These cognitive cores are also integrated with classical planners or control policies in innovative ways. For example, some frameworks use an LLM to dynamically adjust navigation algorithms in real time based on new instructions~\cite{contextualized_drone2025}, while others like SoraNav blend zero-shot VLM reasoning with geometric priors to constrain the action space and enhance decision quality~\cite{soranav2025}. LLMs are also used to generate offline safety rules to inform a risk-aware policy~\cite{neuro_symbolic2026, safe_uav2025} or to dynamically shape reward functions in reinforcement learning~\cite{prompt_informed_reinforcement2025}.

\subsubsection{Vision-Language-Action Models as End-to-End Control Policies}
Pre-trained Vision-Language-Action (VLA) models provide a powerful, generalizable foundation for UAV control by directly mapping multimodal perception to low-level motor commands, effectively creating end-to-end sensorimotor policies. A key insight driving this approach is that robot actions can be treated as text tokens within a VLM's vocabulary, allowing pre-trained knowledge to be transferred directly to control tasks through fine-tuning, a concept pioneered by models like Robotics Transformer 2 (RT-2)~\cite{rt2_2023}. This approach bypasses explicit intermediate representations, training a single model to learn the entire sense-act loop from raw inputs~\cite{vision_based_learning2024, embodied_ai2024, language_driven_representation2023}. VLA methods can be broadly categorized based on their underlying generative mechanism, including autoregression-based, diffusion-based, reinforcement-based, flow-based, and hybrid paradigms~\cite{pure_vla2025}. Recent advances have introduced flow matching architectures ($\pi_0$)~\cite{pi0_2024}, efficient action tokenization ($\pi_0$-FAST)~\cite{pi0_fast2025}, optimized fine-tuning recipes (OpenVLA-OFT)~\cite{openvla_oft2025}, diffusion-based bimanual control (RDT-1B)~\cite{rdt1b2025}, data-efficient approaches (TinyVLA)~\cite{tinyvla2025}, and hybrid diffusion-autoregressive models (HybridVLA)~\cite{hybridvla2025}, collectively demonstrating the rapid diversification of VLA architectures. The landmark OpenVLA model demonstrates this by fine-tuning a VLM on robotics data to predict discretized action tokens, creating a generalizable policy for both manipulation and navigation~\cite{openvla2024}. This has been followed by more advanced architectures like WALL-OSS, which employs a tightly coupled Mixture-of-Experts (MoE) design and a multi-stage training process to bridge internet-scale VLM knowledge with embodied intelligence~\cite{wall_oss2025}. In the aerial domain, models like RaceVLA are fine-tuned for specialized tasks such as autonomous drone racing~\cite{racevla2025}, while frameworks like TrackVLA integrate target recognition and trajectory planning within a unified LLM backbone~\cite{trackvla2025}. These VLAs predominantly use transformer-based backbones paired with policies that can model complex, multi-modal action distributions, with some systems incorporating predictive world models, such as video diffusion models, to enhance foresight~\cite{vision_language_action2024, trivla2025}. This methodology embodies a shift towards policies learned from sensorimotor experience, where the objective is to learn a policy $\pi_{\theta}$ that maps an observation history $o_{\leq t}$ and a language instruction $\mathcal{L}$ to an action using a next-token prediction framework~\cite{aerial_vision2025}:
\begin{equation}
a_t \sim \pi_{\theta}(\cdot | o_{\leq t}, \mathcal{L})
\end{equation}
where $\theta$ represents the model's parameters, and the action $a_t$ is sampled from the policy's output distribution. The versatility of the VLA architecture is demonstrated by its extension to other modalities, such as integrating haptic feedback by fine-tuning an OpenVLA backbone~\cite{vlh2025}.

The development of this paradigm is rapidly advancing toward practical deployment on resource-constrained hardware. Lightweight VLA models such as GRaD-Nav++ are designed to run fully onboard UAVs, enabling real-time, language-guided flight by combining pretrained VLMs with efficient policy networks like Mixture-of-Experts (MoE) trained via reinforcement learning~\cite{grad_nav2025}. To further optimize performance, some methods offload computationally intensive tasks like object detection to edge servers while retaining reactive planning onboard the drone~\cite{autonomous_navigation2025, lightweight_drone2024}. To bridge the sim-to-real gap in policy training, some pipelines blend high-fidelity scene reconstruction with generalizable VLM features to train lightweight visuomotor policies from scratch~\cite{singer2025}. To make these large pre-trained models adaptable to specific UAV tasks, advanced techniques such as few-shot meta-offline reinforcement learning~\cite{resilient_uav2025} and inverse reinforcement learning from a self-built world model~\cite{irl_vla2025} are being developed. However, challenges such as computational overhead, brittle long-horizon planning, and a lack of real-world robustness remain significant hurdles~\cite{embodied_ai2024, vl_nav2025}, motivating the hybrid architectures discussed next.

\subsubsection{World Model and VLA Integration for Physically-Grounded Reasoning}
The most recent paradigm shift in embodied AI involves the deep integration of generative world models with VLA policies, addressing the limitations of purely reactive end-to-end models by equipping agents with foresight and physical common sense~\cite{embodied_llm_wm2025, wm_vla_survey2026}. While early VLAs like OpenVLA~\cite{openvla2024} demonstrated the power of action tokenization, they often struggled with long-horizon planning and open-world generalization due to a lack of explicit environmental understanding. To overcome this, researchers have developed models like $\pi_0$, which utilizes a novel flow matching architecture built on a pre-trained VLM to inherit internet-scale semantic knowledge for general robot control~\cite{pi0_2024}, and its efficient variant $\pi_0$-FAST for optimized action tokenization~\cite{pi0_fast2025}. Concurrently, the development of Navigation World Models (NWM)~\cite{nwm2025} and Video Prediction Policies~\cite{video_prediction_policy2025} has shown that controllable video generation can serve as a powerful predictive engine for planning.

This convergence is exemplified by recent foundation models such as NVIDIA's Cosmos~\cite{cosmos2025} and GR00T N1~\cite{groot_n1_2025}, which employ dual-system architectures to balance fast, reactive control with slow, deliberative reasoning grounded in physical reality. Models like Cosmos-Reason1 explicitly combine world models with embodied reasoning to achieve physical common sense~\cite{cosmos_reason1_2025}. Furthermore, frameworks like 3D-VLA~\cite{3dvla2024}, FLIP~\cite{flip2025}, UP-VLA~\cite{upvla2025}, and GEVRM~\cite{gevrm2025} construct generative world models that allow agents to simulate future states and evaluate potential actions before execution, while unified architectures like RoboBrain~\cite{robobrain2025} bridge abstract reasoning with concrete manipulation through multi-level world modeling. This integration is further enhanced by techniques such as visual chain-of-thought reasoning (CoT-VLA)~\cite{cot_vla2025} and the incorporation of web-scale knowledge (GR-2)~\cite{gr2_2024}. By combining the predictive power of world models with the actionable outputs of VLAs, these integrated systems represent a significant step toward truly generalist embodied agents capable of robust, long-horizon navigation in complex, dynamic environments.

\subsubsection{Architectural Integration: Hierarchical and Hybrid Agentic Systems}
Effective agentic UAVs increasingly adopt hierarchical or hybrid architectures that decouple high-level, deliberative reasoning from low-level, reactive control, balancing the semantic understanding of large models with the real-time demands of flight~\cite{real-time_cooperative2024}. A purely end-to-end VLA may struggle with long-horizon planning and safety verification, while a pure VLM planner can lack the responsiveness needed for dynamic environments. Consequently, structured frameworks have emerged that organize the agentic system into distinct layers for perception, cognition, and control~\cite{uav_agentic2025, embodied_ai2024Y2024, agentic_uavs2025}. This separation of concerns, which reflects a broader trend in robotics of integrating foundation models into the full autonomy stack~\cite{foundation_models2023, foundation_model2025}, allows for a synergistic combination of different models and algorithms, each optimized for a specific function~\cite{large_model_embodied_ai2025}.

This decoupling principle is realized through various architectural patterns that often resemble a "dual-brain" system, with a large model acting as a "cerebrum" for slow deliberation and a smaller module as a "cerebellum" for fast reaction~\cite{agent_controller2023}. Some agents, for instance, run large LLMs for task planning directly on edge hardware, while offloading reactive control to dedicated modules~\cite{aerial_agents2025}. Frameworks like AERMANI-VLM adapt pre-trained VLMs for aerial manipulation without fine-tuning by using structured prompting to select from a predefined library of flight-safe skills~\cite{aermani_vlm2025}. Similarly, the OK-Robot system integrates pre-trained VLMs for perception with navigation primitives to perform tasks in unseen environments without any training~\cite{ok_robot2024}, and Agentic Robot uses a brain-inspired planner-executor-verifier structure for long-horizon manipulation~\cite{agentic_robot2025}. The NaVILA framework uses a VLM to generate mid-level actions in natural language, which are then executed by a separate low-level locomotion policy~\cite{navila2025}. This modularity is also applied in complex logistics tasks, where lightweight LLMs and VLMs are integrated into a pipeline to handle distinct subtasks~\cite{logisticsvln2025}. This layered design also improves transparency, a critical factor for trust in autonomous systems, which can be enhanced with explainable AI (XAI) frameworks that provide visual and textual justifications for an agent's decisions~\cite{explainable_ai2026, uav_navigation_xai2024}.

\subsubsection{Scaling Agentic Intelligence to Multi-UAV Swarms}
Multimodal Large Language Models (MLLMs) are emerging as central planners for UAV swarms, capable of interpreting complex mission goals, dynamically allocating tasks, and coordinating collaborative behaviors among multiple agents in real time. The agentic paradigm extends naturally from a single vehicle to multi-agent systems, with the foundation model functioning as a "swarm brain" or centralized coordinator~\cite{uav_swarms2024}. For example, in a simulated wildfire scenario, an MLLM can process real-time visual data and dynamically assign firefighting tasks to different UAVs, adapting the plan as the situation evolves~\cite{multimodal_uav2025}. This fits into a broader vision of autonomous swarm robotics for applications like self-firefighting and automated resource management~\cite{swarms_uavs2022}. The use of MLLMs is part of a larger ecosystem of deep learning techniques for swarm intelligence, which also includes federated learning and multi-agent reinforcement learning (MARL)~\cite{distributed_machine_learning2023}.

This capability marks a significant shift toward language-based fleet management, where frameworks such as Swarm-GPT~\cite{swarm_gpt2023}, TACOS~\cite{tacos2025}, and LLM-MARS~\cite{llm_mars2023} translate a single human command into a synchronized plan, often using model-based safety filters to ensure collision-free execution. While MARL offers a powerful decentralized paradigm for learning cooperative behaviors~\cite{uav_control2025, multi_uav_cooperative_path2024}, the latest research integrates these approaches. For instance, some frameworks inject LLM-generated expert policies directly into the MARL process~\cite{llm_meets_sky2021, scalable_uav2021}, use LLMs for regional task decomposition in complex environments~\cite{task_offloading2024}, or leverage VLM-RAG systems to dynamically tune control parameters for the swarm in real-time~\cite{swarmvlm2025}. In addition to LLM-centric approaches, other advanced coordination methods like hierarchical reinforcement learning~\cite{scaleable_communication2024} and co-evolutionary algorithms~\cite{co_evolutionary_algorithm2023} also contribute to intelligent swarm management. These hybrid methods leverage the semantic reasoning of MLLMs for strategic planning and the efficiency of MARL or other distributed algorithms for real-time tactical execution, paving the way for more flexible and intelligent multi-UAV systems.

\section{Essential Resources: Simulators, Datasets, and Evaluation} \label{sec:Essential Resources: Simulators, Datasets, and Evaluation}

Progress in UAV-based Vision-and-Language Navigation is critically dependent on the shared infrastructure that facilitates reproducible research and standardized benchmarking. This section provides a structured overview of the three pillars of this infrastructure, as illustrated in Table~\ref{tab:resources_tree}: the simulation platforms that serve as virtual testbeds, the benchmark datasets that define canonical tasks, and the evaluation protocols that quantify agent performance. The table presents a hierarchical overview of these essential resources, categorizing them into three distinct pillars---High-Fidelity Simulation Environments, Benchmark Datasets, and Evaluation Protocols---while also detailing their evolution from foundational tools to advanced, multi-faceted systems designed to improve model generalization and bridge the sim-to-real gap. Our analysis focuses on resources that have been pivotal in shaping the field, tracing their evolution from adapted ground-based tools to specialized aerial systems designed to address challenges like the sim-to-real gap. We begin by examining the cornerstone of this ecosystem, the development of \textit{High-Fidelity Simulation Environments}, which are instrumental in training and evaluating navigation policies by overcoming the high cost and time complexity of physical experiments, enabling the large-scale parallel training necessary for modern learning algorithms and facilitating sim-to-real transfer despite the persistent challenge of the sim-to-real gap \cite{robotic_navigation2025,benchmarking_deep_reinforcement2024,sim_to_real2020,task_oriented_flying2025,navrl2025}.
	\subsection{High-Fidelity Simulation Environments} \label{subsec:High-Fidelity Simulation Environments}
	
	High-fidelity simulation is not merely a tool for data collection but a core enabling technology for UAV-VLN. To provide a consolidated overview of the simulation ecosystem, Table~\ref{tab:simulators} compares the major platforms discussed in this section along key dimensions including rendering fidelity, physics support, and primary research contributions. The evolution from foundational game-engine-based simulators to diverse, photorealistic, and differentiable platforms directly mirrors the field's progress from high-level planning to mastering end-to-end control for effective sim-to-real transfer. This progression is driven by the need to close the persistent sim-to-real gap, which has both a perceptual component (visual fidelity) and a dynamics component (action execution)~\cite{visual_semantic_navigation2025, sim_to_real2020}. Addressing this gap necessitates simulators that can provide not only visually realistic scenes but also accurate physical interactions, paving the way for policies trained in simulation to be deployed successfully on real-world hardware~\cite{vision_language_action2024, task_oriented_flying2025, vision_transformers2025, rt2_2023}.

\begin{table*}[!t]
\centering
\resizebox{\textwidth}{!}{%
\begin{tabular}{llccccl}
\toprule
\textbf{Platform} & \textbf{Rendering Engine} & \textbf{Physics} & \textbf{ROS} & \textbf{Multi-Engine} & \textbf{3DGS} & \textbf{Primary Use in UAV-VLN} \\
\midrule
AirSim \cite{autonomous_uav_navigation2022} & Unreal Engine 4 & Fast Physics & \checkmark & \ding{55} & \ding{55} & High-fidelity visual training; AerialVLN testbed \\
Gazebo + ROS \cite{spar2025} & OGRE / SDF & ODE / Bullet & \checkmark & \ding{55} & \ding{55} & Digital twins; physics-accurate robotics integration \\
CARLA \cite{logisticsvln2025} & Unreal Engine 4 & PhysX & \checkmark & \ding{55} & \ding{55} & Urban driving scenes; logistics VLN \\
Unity ML-Agents \cite{morphonavi2025} & Unity HDRP & PhysX & \checkmark & \ding{55} & \ding{55} & Morphing vehicle simulation; indoor scenes \\
AerialVLN Sim \cite{aerialvln2023} & Unreal Engine 4 & AirSim & \ding{55} & \ding{55} & \ding{55} & First dedicated city-scale UAV-VLN benchmark \\
OpenFly \cite{openfly2025} & UE + GTA V + Google Earth & Mixed & \ding{55} & \checkmark & \checkmark & Large-scale diverse data generation (100K trajectories) \\
NVIDIA Isaac Sim \cite{raven2025} & RTX (Omniverse) & PhysX 5 & \checkmark & \ding{55} & \ding{55} & Photorealistic rendering; parallelized RL training \\
3DGS Environments \cite{grad_nav2025} & 3D Gaussian Splatting & Differentiable & \ding{55} & \ding{55} & \checkmark & End-to-end differentiable sim-to-real transfer \\
CARLA-Air \cite{zeng2026carla} & Unreal Engine 4 & PhysX + AirSim & \checkmark & \checkmark & \ding{55} & Unified air-ground embodied sim; dual native API; 18 synced sensor streams \\
\bottomrule
\end{tabular}}
\caption{Comparison of high-fidelity simulation platforms used in UAV-VLN research. Platforms are categorized by their rendering engine, physics fidelity, and primary contribution to the field. 3DGS = 3D Gaussian Splatting; UE = Unreal Engine; ROS = Robot Operating System. CARLA-Air uniquely integrates CARLA and AirSim within a single UE4 process, providing both ground and aerial native APIs with shared physics tick and rendering pipeline.}
\label{tab:simulators}
\end{table*}

	\subsubsection{Foundational Simulators: Establishing the AerialVLN Task}
	The initial development of UAV-VLN necessitated a shift from ground-based simulators to dedicated aerial environments, which, while foundational, primarily served to establish task-specific benchmarks and expose the significant performance gap between simulated agents and real-world execution. While ground-based VLN research matured in platforms built from real-world imagery like the Matterport3D Simulator~\cite{vision_language_navigation2018} or in simulated indoor environments like Habitat and AI2-THOR~\cite{vision_language_navigation2021, goal_oriented_semantic_exploration2020, l3mvn2023, indooruav2025, visual_language_maps2023, octonav2025}, the unique challenges of 3D flight required new testbeds. The community widely adopted a diverse range of platforms, a landscape surveyed in works like~\cite{drone_simulators2019}. High-fidelity game engines, primarily Unreal Engine (UE) with the AirSim plugin~\cite{autonomous_uav_navigation2022, uavs_llms2025, skyvln2025, vision_based2024, reinforcement_learning_uav2023} and Unity~\cite{morphonavi2025}, offered photorealistic rendering. Concurrently, the Robot Operating System (ROS) integrated with Gazebo provided robust physics simulation and robotics integration, forming the basis for many high-fidelity digital twins~\cite{spar2025, hierarchical_multi_uav2024, memory_based_deep2018, fm_planner2025, soranav2025, uav_path2023, agentic_uavs2025, model_free_uav2024, path_planning_algorithm}. Other specialized simulators like CARLA~\cite{logisticsvln2025, pure_vla2025}, Webots~\cite{ground_aerial_transportation2025}, and even MATLAB~\cite{unmanned_aerial_vehicle_path_planning} were also adapted for various research needs. The first large-scale benchmark specifically for UAV-VLN was introduced with the AerialVLN simulator~\cite{aerialvln2023}, which quickly became a key testbed~\cite{aerial_vision2025, urbanvideo_bench}. Built on UE4 and AirSim, it provided city-level environments with dynamic conditions, formalizing the task and quantifying the poor performance of initial models, thereby underscoring the severity of the sim-to-real challenge. Recent work has sought to unify these complementary capabilities; for example, CARLA-Air~\cite{zeng2026carla} integrates CARLA and AirSim within a single Unreal Engine process, providing synchronized aerial-ground simulation with both native APIs preserved, which directly supports air-ground cooperative VLN research.

\subsubsection{Scaling Diversity with Multi-Engine Simulation Platforms}
To overcome the generalization limits of single-source simulators, the field progressed towards comprehensive platforms that integrate multiple rendering engines and automated data-generation pipelines, enabling the creation of large-scale, diverse datasets essential for training robust, generalizable navigation models. Relying on a single game engine, even a high-fidelity one, creates a data diversity bottleneck, as agents can overfit to its specific rendering style and environmental assets. To address this, a comprehensive solution emerged with platforms like OpenFly~\cite{openfly2025}, which tackles the diversity challenge by integrating multiple rendering engines and data sources including Unreal Engine, Grand Theft Auto V (GTA V), Google Earth, and 3D Gaussian Splatting through an automated data generation pipeline~\cite{openfly2024, multimodal_spatial_reasoning2025, geospatial_reasoning2023}.

\begin{figure*}[!htb]
    \centering
    \includegraphics[width=0.9\textwidth,height=0.5\textheight,keepaspectratio]{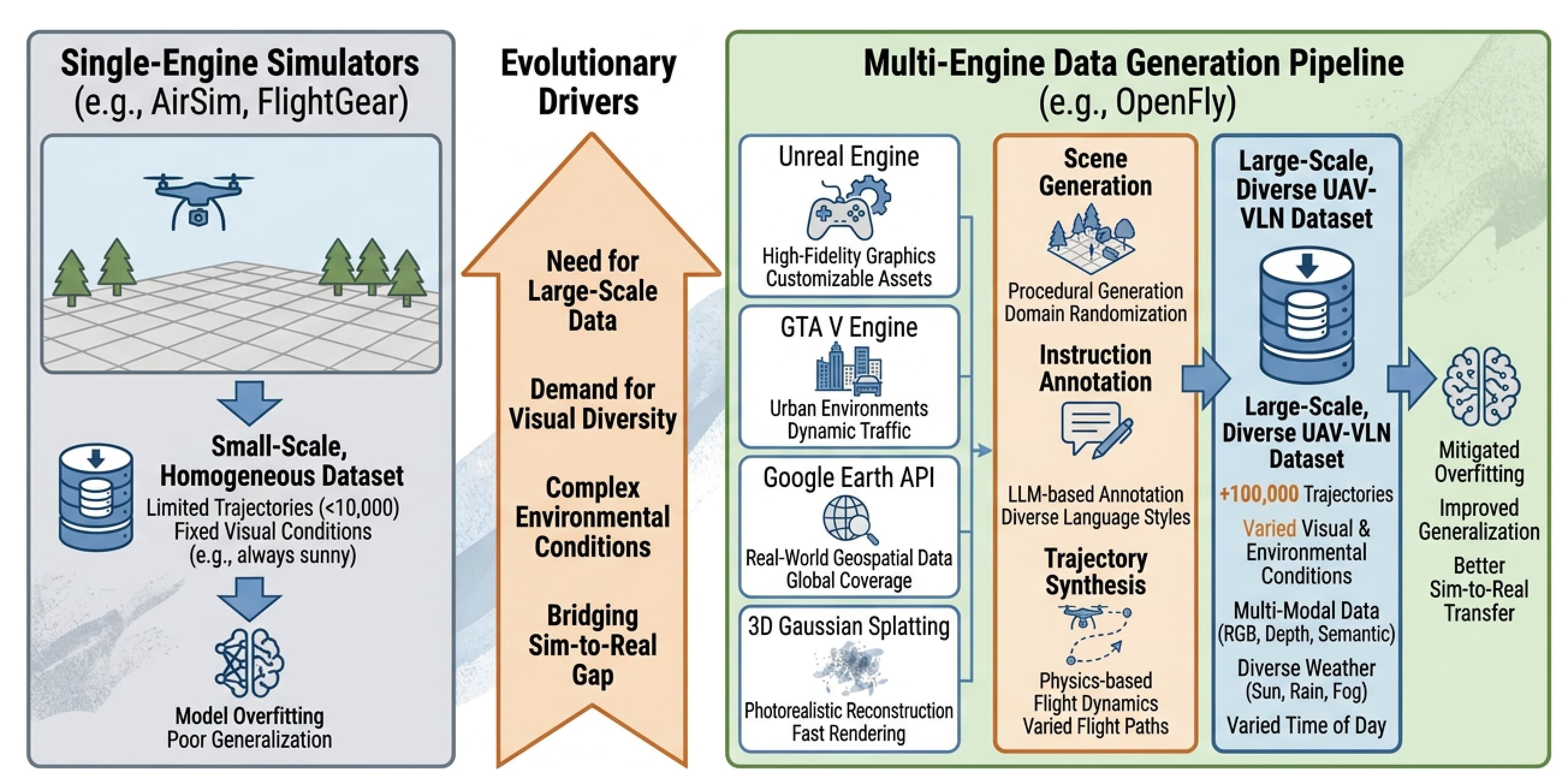}
    \caption{The evolution from single-engine simulators to multi-engine data generation pipelines. Platforms like OpenFly integrate diverse sources like game engines and geospatial data to create large-scale, varied datasets, mitigating model overfitting and improving generalization.}
    \label{fig:essential_resources__simulators,_datasets,_and_evaluation_auto_1}
\end{figure*}
 This methodological advance of programmatically combining disparate data sources is exemplified by techniques that use tools like CityEngine to generate realistic urban scenes with aligned multimodal data~\cite{synthetic_urban2026} or create pseudo-synthetic aerial datasets from large-scale 3D city meshes~\cite{aerialmegadepth2025}. This approach provides a scalable way to source training data with varied visual and environmental conditions, which is essential for training models that can generalize across unseen landscapes~\cite{uav_vlpa2025}.

\subsubsection{Photorealistic and Differentiable Worlds for End-to-End Transfer}
The current frontier in simulation leverages photorealistic, differentiable rendering techniques like 3D Gaussian Splatting (3DGS) to create environments so faithful to reality that they permit direct, sample-efficient training of end-to-end, low-level control policies that successfully transfer to physical hardware. This represents a leap from high-fidelity visuals to functionally high-fidelity, interactive environments. Platforms such as NVIDIA Isaac Sim are increasingly used for their photorealistic rendering and sophisticated physics engines, enabling complex evaluations that more closely mirror real-world dynamics~\cite{raven2025, robotic_navigation2025, dialogue_px42025, navila2025}. The key innovation is the use of rendering techniques that are not only realistic but also fully differentiable. 3DGS, for instance, allows for the creation of high-fidelity scene representations from real-world captures~\cite{embodied_gap2025}, enabling both photorealistic real-time rendering~\cite{reasongrounder2025} and the generation of vast synthetic datasets~\cite{singer2025}. Crucially, this technology is evolving beyond static rendering to create executable environments with object-level semantics and physics, allowing for realistic agent interaction~\cite{executable_3d_gaussian2024}. When this differentiable rendering is combined with a differentiable physics model, it becomes possible to train a policy end-to-end, allowing gradients from a high-level task objective to flow back through the entire perception-action loop to update the control policy parameters $\theta$:
{\small
\begin{equation}
\nabla_{\theta} J(\pi) = \mathbb{E}_{\tau \sim \pi} \left[ \sum_{t=0}^{H} \nabla_{\theta} \log \pi(a_t \mid o_t)\, G_t \right]
\end{equation}
	}
	where $G_t = \sum_{t'=t}^{H} \gamma^{t'-t} \mathcal{R}(s_{t'}, a_{t'})$, and $o_t$ is differentiable with respect to the scene representation. As shown by Grad-Nav~\cite{grad_nav2025} and SINGER~\cite{singer2025}, this enables direct learning of continuous low-level control from pixels and text, narrowing both the perceptual and dynamics sim-to-real gaps.
	
	\begin{table*}[!htb]
\centering
\resizebox{\textwidth}{!}{%
\begin{tabular}{p{3.8cm}p{4.2cm}p{10.5cm}}
\toprule
\textbf{Pillar} & \textbf{Evolution Stage} & \textbf{Key Developments} \\
\midrule
\multirow{3}{*}{\parbox{3.8cm}{\textbf{High-Fidelity\\Simulation\\Environments}}}
 & Foundational Simulators & Shifted from ground-based platforms (Matterport3D) to dedicated aerial environments. Utilized UE4 + AirSim and Gazebo + ROS for photorealistic rendering and physics. The AerialVLN simulator established the first large-scale benchmark and quantified the sim-to-real challenge. \\
\cmidrule(l){2-3}
 & Multi-Engine Platforms & Developed to overcome data diversity bottlenecks of single-engine simulators. OpenFly integrates UE, GTA V, Google Earth, and 3DGS through automated pipelines, enabling creation of 100K+ diverse training trajectories. \\
\cmidrule(l){2-3}
 & Photorealistic \& Differentiable Worlds & Leverages 3D Gaussian Splatting (3DGS) and NVIDIA Isaac Sim for photorealistic, differentiable environments. Enables direct end-to-end training of low-level control policies and closes both perceptual and dynamics portions of the sim-to-real gap. \\
\midrule
\multirow{4}{*}{\parbox{3.8cm}{\textbf{Benchmark Datasets\\and Task Variations}}}
 & City-Scale Navigation & Pioneering benchmarks (AerialVLN: 8,446 trajectories, 25K instructions) with long-horizon trajectories and continuous 4-DoF action spaces. Revealed a 5.1\% (model) vs.\ 80.8\% (human) success rate gap. \\
\cmidrule(l){2-3}
 & Interactive \& Fine-Grained Tasks & AVDN introduced multi-turn human-human dialog for ambiguity resolution. Expanded to delivery logistics, social navigation, and fine-grained control evaluation (UAV-Flow: 30K+ real-world trajectories). \\
\cmidrule(l){2-3}
 & Large-Scale for Generalization & CityNav (32,637 real-world trajectories), OpenFly (100K+), and SAGE-Bench (2M pairs in 3DGS worlds). Specialized domains: agriculture (A2A), disaster response (DisasterM3), indoor (IndoorUAV), and remote sensing. \\
\cmidrule(l){2-3}
 & Decomposed Skill Evaluation & BEDI benchmark decomposes missions into perception, planning, and control sub-skills. NAV-NUANCES tests atomic instructions. Richer annotations for visual grounding (AerialVG), landmarks (NavAgent-Landmark2K), and 3D QA (Open3D-VQA). \\
\midrule
\multirow{3}{*}{\parbox{3.8cm}{\textbf{Evaluation Protocols\\and Embodied Skill\\Assessment}}}
 & Holistic Task-Completion Metrics & SR, NE, OSR for goal achievement; SPL for path efficiency; nDTW, SDTW, CLS for trajectory fidelity. Effective for benchmarking end-to-end performance but obscure specific failure reasons. \\
\cmidrule(l){2-3}
 & Granular Decomposed Skills & BEDI: Dynamic Chain-of-Embodied-Task evaluating six sub-skills (semantic/spatial perception, motion control, tool use, planning, action generation). Skill-specific metrics: mIoU, mAP, BLEU/ROUGE, and GPT-4 automated assessment. \\
\cmidrule(l){2-3}
 & Sim-to-Real Context & Performance degrades significantly from simulation to reality (e.g., SR drops from 46.8\% with map to 22.5\% without; 33.2\% seen to 10.7\% unseen in OpenFly). Motivates diverse benchmarks and testing across broad scenario distributions. \\
\bottomrule
\end{tabular}}
\caption{A hierarchical overview of the essential resources for UAV-based Vision-and-Language Navigation research, organized into three pillars: High-Fidelity Simulation Environments, Benchmark Datasets and Task Variations, and Evaluation Protocols and Embodied Skill Assessment. Each pillar traces the evolution from foundational tools to advanced systems designed to improve model generalization and bridge the sim-to-real gap.}
\label{tab:resources_tree}
\end{table*}

	\subsection{Benchmark Datasets and Task Variations} \label{subsec:Benchmark Datasets and Task Variations}

\begin{table*}[!htb]
\centering
\resizebox{\textwidth}{!}{%
\begin{tabular}{llrrlllll}
\toprule
\textbf{Dataset} & \textbf{Year} & \textbf{\#Traj.} & \textbf{\#Instr.} & \textbf{Environment} & \textbf{Interaction} & \textbf{Action} & \textbf{Data Source} & \textbf{Distinguishing Feature} \\
\midrule
AerialVLN \cite{aerialvln2023} & 2023 & 8,446 & 25,024 & Outdoor urban/rural & Single instruction & C, 4-DoF & UE4 + AirSim & First city-scale UAV-VLN benchmark \\
AVDN \cite{aerial_vision_dialog2024} & 2023 & 3,064 & Multi-turn & Outdoor urban & Human-human dialog & C & xView imagery & Multi-turn dialog for ambiguity resolution \\
CityNav \cite{citynav2025} & 2025 & 32,637 & 32,637 & Real-world cities & Single instruction & C, 4-DoF & Google Earth & Largest real-world aerial navigation dataset \\
OpenFly \cite{openfly2025} & 2025 & 100,000+ & 100,000+ & Diverse outdoor & Single instruction & C & UE + GTA V + 3DGS & Multi-engine data generation; 18 scenes \\
UAV-Flow \cite{uav_flow2025} & 2025 & 30,000+ & --- & Real-world flight & Trajectory replay & C, 6-DoF & Real flight logs & Fine-grained continuous control evaluation \\
IndoorUAV \cite{indooruav2025} & 2025 & --- & --- & Indoor environments & Single instruction & C & Simulation & Complex indoor aerial navigation \\
LogisticsVLN \cite{logisticsvln2025} & 2025 & --- & --- & Terminal logistics & Task-specific & C & CARLA & Application-oriented delivery tasks \\
AgriVLN (A2A) \cite{agrivln2024} & 2024 & --- & --- & Agricultural fields & Single instruction & C & Custom sim & Domain-specific agricultural navigation \\
BEDI \cite{bedi_benchmark} & 2025 & --- & --- & Mixed (sim + real) & Chain-of-task & C & Hybrid & Decomposed sub-skill evaluation \\
SAGE-Bench \cite{executable_3d_gaussian2024} & 2025 & 2,000,000 & 2,000,000 & 3DGS environments & Single instruction & C & 3DGS & Physically valid executable 3D worlds \\
\bottomrule
\end{tabular}}
\caption{Comparison of major benchmark datasets for UAV-based Vision-and-Language Navigation. Datasets are organized chronologically and characterized by their scale, environment type, interaction modality, and action space. Traj.\ = Trajectories; Instr.\ = Instructions; DoF = Degrees of Freedom; C = Continuous; D = Discrete.}
\label{tab:datasets}
\end{table*}

The evolution of UAV-VLN benchmarks reflects a clear progression from establishing foundational navigation challenges in large-scale aerial environments to embracing greater task complexity, data scale, and standardized, multi-faceted evaluation paradigms to drive progress towards real-world generalization and capability. Table~\ref{tab:datasets} provides a structured comparison of the major benchmark datasets discussed in this section, highlighting the rapid growth in data scale and task diversity. Unlike ground-based VLN, which matured on datasets captured from constrained indoor environments, such as Matterport3D often utilized within simulators like Habitat~\cite{visual_language_maps2023, vision_language_navigation2018, vision_language_navigation2021}, or from large-scale manipulation-focused collections like the Open X-Embodiment dataset~\cite{pure_vla2025}, aerial navigation required new benchmarks to address the unique challenges of long-horizon flight in both expansive outdoor spaces and complex indoor environments~\cite{indooruav2025}. This section analyzes the pivotal datasets and task structures that have defined and propelled the field, addressing the data scarcity and diversity challenges critical for training robust models~\cite{uavs_llms2025, remote_sensing_foundation_models2024}.

\subsubsection{Foundational Benchmarks for City-Scale Navigation}
Pioneering benchmarks like AerialVLN established the fundamental difficulty of city-scale UAV navigation by introducing long-horizon trajectories, large 3D action spaces, and dynamic environments, revealing a significant performance gap between initial computational models and human pilots. The seminal AerialVLN dataset~\cite{aerialvln2023} formalized the task by providing over 8,400 trajectories with 25,000 instructions across diverse urban and rural scenes like downtowns, factories, and parks~\cite{aerial_vision2025G2025}. Its key contributions were quantifying the challenge with long flight paths averaging over one kilometer and a continuous 4-DoF action space, a significant departure from the discrete actions common in early ground-based navigation~\cite{vision_language_navigation2018}, although subsequent ground-based benchmarks also evolved to incorporate continuous environments~\cite{dagger_diffusion2021}. The dataset's inclusion of distinct unseen environments for validation and testing established a rigorous protocol for evaluating generalization. The stark results from the initial benchmark were telling: the best-performing baseline model achieved a mere 5.1\% success rate, compared to 80.8\% for human pilots, thereby defining the primary research objective for subsequent work as closing this vast performance gap. The influence of this foundational work is evident in subsequent benchmarks like AirVLN-S and AirVLN-E, which provide further city-level scenes with fine-grained instructions to evaluate more sophisticated agents~\cite{citynavagent2025, spatial_representation2025}.

\subsubsection{Extending to Interactive and Fine-Grained Tasks}
To address the limitations of one-way instruction following, the field evolved toward more complex and realistic interaction models, exemplified by the AVDN benchmark, which introduces multi-turn dialog to resolve ambiguity and enable collaborative navigation. Real-world instructions are often underspecified, requiring agents to seek clarification. The Aerial Vision-and-Dialog Navigation (AVDN) dataset was created to address this, shifting the paradigm from static instruction-following to a dynamic, conversational setting~\cite{aerial_vision_dialog2024, target_grounded2023}. Comprising over 3,000 trajectories with human-human dialogs, AVDN allows the agent to ask clarifying questions, enabling it to ground instructions more effectively and disambiguate references, a critical capability for robust navigation~\cite{enhancing_visual2025}. Concurrently, other datasets have expanded task complexity by introducing new, application-oriented scenarios, such as terminal delivery logistics~\cite{logisticsvln2025} and socially-aware navigation that requires understanding human-centric contexts~\cite{ha_vln2025}. This specialization is also reflected in benchmarks like the target-oriented UAV-Need-Help, which focuses evaluation on scenarios where the agent must identify and navigate to specific objects of interest~\cite{uav_vision_language2024}, and the Object-Centric VLN (OC-VLN) benchmark, introduced to isolate the challenge of grounding instructions to specific objects~\cite{zero_shot_object_centric2025}. In parallel, other benchmarks shifted focus from high-level dialog to low-level control fidelity. The UAV-Flow dataset, for example, provides over 30,000 real-world flight trajectories with synchronized flight state data, enabling the development and evaluation of policies for fine-grained, continuous action generation~\cite{uav_flow2025}.

\subsubsection{The Push for Scale and Diversity to Foster Generalization}
A major trend aimed at overcoming the poor generalization of early models has been the creation of large-scale, diverse datasets that leverage multiple data sources and real-world city data. The limited diversity of early datasets, along with a noted lack of geographic and cultural variety in broader urban datasets like Cityscapes and BDD100K~\cite{urban_monitoring2025}, was a critical bottleneck hindering the development of globally robust models~\cite{vision_language_action2025}. To address this, benchmarks like CityNav provide over 32,600 human demonstration trajectories across real-world cities, enabling the training of high-performance planning models~\cite{citynav2025, geonav2025, flightgpt2024, sa_gcs2024}. The push for diversity is epitomized by platforms like OpenFly, which integrate multiple rendering engines including Unreal Engine, GTA V, and Google Earth into an automated data generation pipeline to create over 100,000 varied and robust training trajectories~\cite{openfly2025, openfly2024}. This trend towards large-scale synthetic data is further exemplified by benchmarks like SAGE-Bench, which offers two million instruction-trajectory pairs in physically valid 3DGS environments~\cite{executable_3d_gaussian2024}, and UEMM-Air, which provides 120,000 image pairs with rich multi-modal annotations including depth and segmentation~\cite{uemm_air2021}. This expansion extends to specialized domains with datasets for agriculture (A2A)~\cite{agrivln2024}, disaster response (DisasterM3)~\cite{disasterm3_2025}, varied outdoor environmental settings (UAV-ON)~\cite{uav_on2025}, complex indoor navigation (IndoorUAV)~\cite{indooruav2025}, and human-centric interactions (UAV-Human)~\cite{uav_human2021}. This expansion also includes a growing ecosystem of large-scale datasets for remote sensing, encompassing massive instruction-following collections for training foundation models~\cite{falcon2025, geochat2023}, benchmarks with fine-grained annotations for visual grounding and relation detection on datasets like DIOR-RSVG~\cite{rsvg2022, skysensegpt2024}, and collections for question-answering across various benchmarks including RSVQA-LR/HR and RSGPT~\cite{rsgpt2023, vision_language_models2024S2024, remote_sensing_lvlm2025}. This rich ecosystem of annotated datasets is crucial for systematically benchmarking the capabilities of different Remote Sensing Foundation Models (RSFMs) across varied tasks and sensor modalities~\cite{foundation_models_survey}. This ecosystem is further specialized with datasets for specific modalities like SAR~\cite{sarlang-1m_benchmark}, ultra-high-resolution imagery~\cite{xlrs_bench2025}, multi-temporal urban analysis~\cite{dynamicvl2025}, disaster assessment~\cite{postdisaster_evaluation2026}, and referring image segmentation~\cite{aeroreformer2025}. Collectively, these resources provide the scale and variety necessary to train the next generation of generalizable foundation models.

\subsubsection{Task Decomposition and Standardized Skill Evaluation}
A concurrent evolution in benchmarking moves beyond holistic success metrics towards standardized, multi-faceted frameworks like BEDI, which decompose complex missions into core sub-skills for more granular and objective capability assessment. While metrics like Success Rate (SR) and Success weighted by Path Length (SPL) are useful for gauging overall performance, they fail to diagnose specific model weaknesses in perception, planning, or control. The BEDI benchmark addresses this by proposing a "Dynamic Chain-of-Embodied-Task" paradigm that separately evaluates these core competencies~\cite{bedi_benchmark}. Its dataset, comprising static imagery, dynamic videos, and a high-fidelity virtual environment, covers diverse terrains and scenarios to facilitate targeted analysis. This shift towards decomposed evaluation is further crystallized by benchmarks like NAV-NUANCES, which provides a framework for testing atomic instructions across fundamental categories like direction changes and landmark recognition~\cite{vision_language_navigation}. This trend also includes highly specialized benchmarks, such as UAV-VLPA-nano-30, designed to isolate and evaluate an agent's ability to generate coherent navigation plans from linguistic commands against human-generated references~\cite{uav_vla2025}. It is also supported by the creation of datasets with richer, more granular annotations. For example, AerialVG provides extensive object-level descriptions for visual grounding~\cite{Liu_2025_ICCV}, NavAgent-Landmark2K offers fine-grained landmark annotations in real urban scenes~\cite{navagent_urban2024}, and GeoText-1652 includes detailed region-level descriptions with spatial relationships~\cite{geotext1652_2024}. Similarly, benchmarks like VRSBench, UrbanVideo-Bench, and Open3D-VQA are designed to evaluate specific skills such as open-ended question answering over dynamic aerial videos and complex 3D spatial reasoning~\cite{vrsbench2024, urbanvideo_bench, open3d_vqa2024}. This movement toward more analytical benchmarking, often involving evaluation across a suite of different datasets and metrics, is critical for understanding the multifaceted capabilities of large foundation models~\cite{embodied_navigation_foundation_model2024} and guiding future research beyond simply optimizing a single, holistic metric.

\subsection{Evaluation Protocols and Embodied Skill Assessment} \label{subsec:Evaluation Protocols and Embodied Skill Assessment}

\begin{table*}[!htb]
\centering
\resizebox{\textwidth}{!}{%
\begin{tabular}{lllll}
\toprule
\textbf{Category} & \textbf{Metric} & \textbf{Abbreviation} & \textbf{Measures} & \textbf{Representative Benchmarks} \\
\midrule
\multirow{7}{*}{\parbox{2.8cm}{\textbf{Holistic\\Task-Completion}}}
 & Success Rate & SR & Binary goal arrival within threshold & AerialVLN, CityNav, OpenFly \\
 & Oracle Success Rate & OSR & Any waypoint within goal radius & AerialVLN, AeroDuo \\
 & Navigation Error & NE & Euclidean distance to goal at stop & AerialVLN, CityNav \\
 & Success weighted by Path Length & SPL & Goal success penalized by path efficiency & AerialVLN, CityNav, OpenFly \\
 & Normalized Dynamic Time Warping & nDTW & Path alignment with reference trajectory & AerialVLN, UAV-Flow \\
 & Success weighted by nDTW & SDTW & Path fidelity combined with task success & AerialVLN \\
 & Coverage weighted by Length Score & CLS & Path coverage and length efficiency & HAMT, AerialVLN \\
\midrule
\multirow{6}{*}{\parbox{2.8cm}{\textbf{Decomposed\\Sub-Skill}}}
 & Semantic Perception Accuracy & --- & Object recognition and scene understanding & BEDI, VRSBench \\
 & Spatial Perception Accuracy & --- & Distance estimation, spatial reasoning & BEDI, NAV-NUANCES \\
 & Motion Control Completeness & --- & Precise execution of movement commands & BEDI \\
 & Mean Intersection over Union & mIoU & Semantic segmentation quality & DisasterM3, UEMM-Air \\
 & Mean Average Precision & mAP & Object detection accuracy & UAV-Human, UEMM-Air \\
 & BLEU / METEOR / CIDEr & --- & Language generation quality & VRSBench, RSGPT \\
\midrule
\multirow{3}{*}{\parbox{2.8cm}{\textbf{Operational\\Robustness}}}
 & Collision Rate & CR & Frequency of obstacle collisions & HA-VLN, ASMA \\
 & Sim-to-Real SR Drop & $\Delta$SR & Performance degradation from sim to real & Sim-to-Real studies \\
 & Inference Frequency & Hz & Real-time control capability & GRaD-Nav++ (25\,Hz), RaceVLA \\
\bottomrule
\end{tabular}}
\caption{Taxonomy of evaluation metrics for UAV-VLN, categorized into holistic task-completion metrics and granular decomposed-skill assessments. Holistic metrics evaluate overall navigation performance, while decomposed metrics diagnose specific sub-skill competencies to guide targeted model improvement.}
\label{tab:metrics}
\end{table*}

Effective evaluation in UAV-VLN is evolving from monolithic, goal-oriented metrics towards diagnostic, multi-skill frameworks that decompose complex navigation tasks into fundamental embodied capabilities, such as direction change, landmark and region recognition, vertical movement, and numerical comprehension. Table~\ref{tab:metrics} provides a structured taxonomy of the evaluation metrics discussed in this section, organized across three categories: holistic task-completion, decomposed sub-skill assessment, and operational robustness. This fine-grained approach, which addresses the performance gaps highlighted in recent benchmarks, enables a more precise analysis of agent failures by identifying specific weaknesses like poor numerical understanding or directional biases, thereby guiding targeted model improvement \cite{uav_vision_language2024,uav_vln2025,grounded_vision2025,vision_language_navigation}. This section details the metrics used to quantify agent performance, from holistic measures of task completion to granular assessments of core competencies. It addresses not only the mathematical formulation of these metrics but also the critical context of the sim-to-real gap, which can significantly impact their validity.

\subsubsection{Holistic Metrics for Navigation Task Completion}
Standard evaluation in UAV-VLN is anchored by holistic metrics that quantify ultimate task success and path efficiency, with variants that also assess the fidelity of the trajectory against a reference path. The most fundamental metric is the Success Rate (SR), which measures the binary outcome of whether an agent stops within a predefined radius of the target destination, a standard established in early ground-based VLN~\cite{vision_language_navigation2018} and widely adopted in aerial settings~\cite{aerialvln2023, sim_to_real2020}. This is complemented by the Navigation Error (NE), the final Euclidean distance to the goal, and the Oracle Success Rate (OSR), which considers an episode a success if any point along the agent's path falls within the success radius~\cite{aerialvln2023, aeroduo2025, citynav2025, octonav2025}. To account for path efficiency, the Success weighted by Path Length (SPL) metric is widely adopted across benchmarks~\cite{history_aware2023, skyvln2025, uav_on2025, l3mvn2023, visual_language_maps2023}. SPL penalizes unnecessarily long or circuitous routes by normalizing the success outcome with the ratio of the optimal path length to the agent's executed path length. It is formally defined as:
\begin{equation}
\text{SPL} = \frac{1}{N} \sum_{i=1}^{N} S_i \frac{L_i^*}{\max(L_i, L_i^*)}
\end{equation}
where for episode $i$ out of a total of $N$ episodes, $S_i$ is a binary indicator of success, $L_i^*$ is the length of the shortest path, and $L_i$ is the length of the agent's actual path~\cite{robotic_navigation2025}. Together, these metrics form the cornerstone of evaluation in numerous studies~\cite{uav_vln2025, dagger_diffusion2021, waypoint_models2021, fine_grained_alignment2024, embodied_gap2025}, often supplemented by simpler measures like total Trajectory Length (TL)~\cite{fine_grained_alignment2024} and Goal Progress (GP)~\cite{target_grounded2023}. For reinforcement learning paradigms, evaluation may also focus on learning efficiency, using metrics like cumulative episode reward and the number of steps to convergence~\cite{reinforcement_learning_uav2023}.

To provide a more nuanced assessment of trajectory quality beyond simple goal achievement, a second set of metrics evaluates path fidelity against a reference or optimal trajectory. The normalized Dynamic Time Warping (nDTW) measures the alignment between the agent's path $P$ and the reference path $R$, providing a score that is robust to variations in speed and execution timing~\cite{uav_flow2025, history_aware2023, vision_language_navigation}. Its formulation is:
\begin{equation}
\text{nDTW}(P, R) = \frac{\text{DTW}(P, R)}{\max(|P|, |R|)}
\end{equation}
where DTW is the Dynamic Time Warping distance and $|P|$ and $|R|$ are the path lengths~\cite{robotic_navigation2025}. Success weighted by nDTW (SDTW) combines this path similarity score with task success, rewarding agents that not only reach the goal but also follow the intended route closely~\cite{aerialvln2023, sim_to_real2020}. Another path-following metric, Coverage weighted by Length Score (CLS), is particularly useful in tasks where covering the reference path is more important than ending at the exact destination, as it jointly considers path coverage and length efficiency~\cite{robotic_navigation2025, history_aware2023}. While these metrics effectively benchmark end-to-end performance, their monolithic nature obscures the specific reasons for failure, motivating a shift towards more diagnostic evaluation frameworks.

\subsubsection{Granular Assessment of Decomposed Embodied Skills}
To overcome the diagnostic limitations of holistic metrics, recent benchmarks advocate for a paradigm shift towards decomposing complex missions into a chain of fundamental, measurable embodied skills. This approach enables more precise analysis of agent failures and targeted model improvement. The principal example of this paradigm is the BEDI benchmark, which introduces a Dynamic Chain-of-Embodied-Task framework to systematically evaluate agents across a spectrum of core competencies~\cite{bedi_benchmark}.

\begin{figure*}[!htb]
    \centering
    \includegraphics[width=0.9\textwidth,height=0.5\textheight,keepaspectratio]{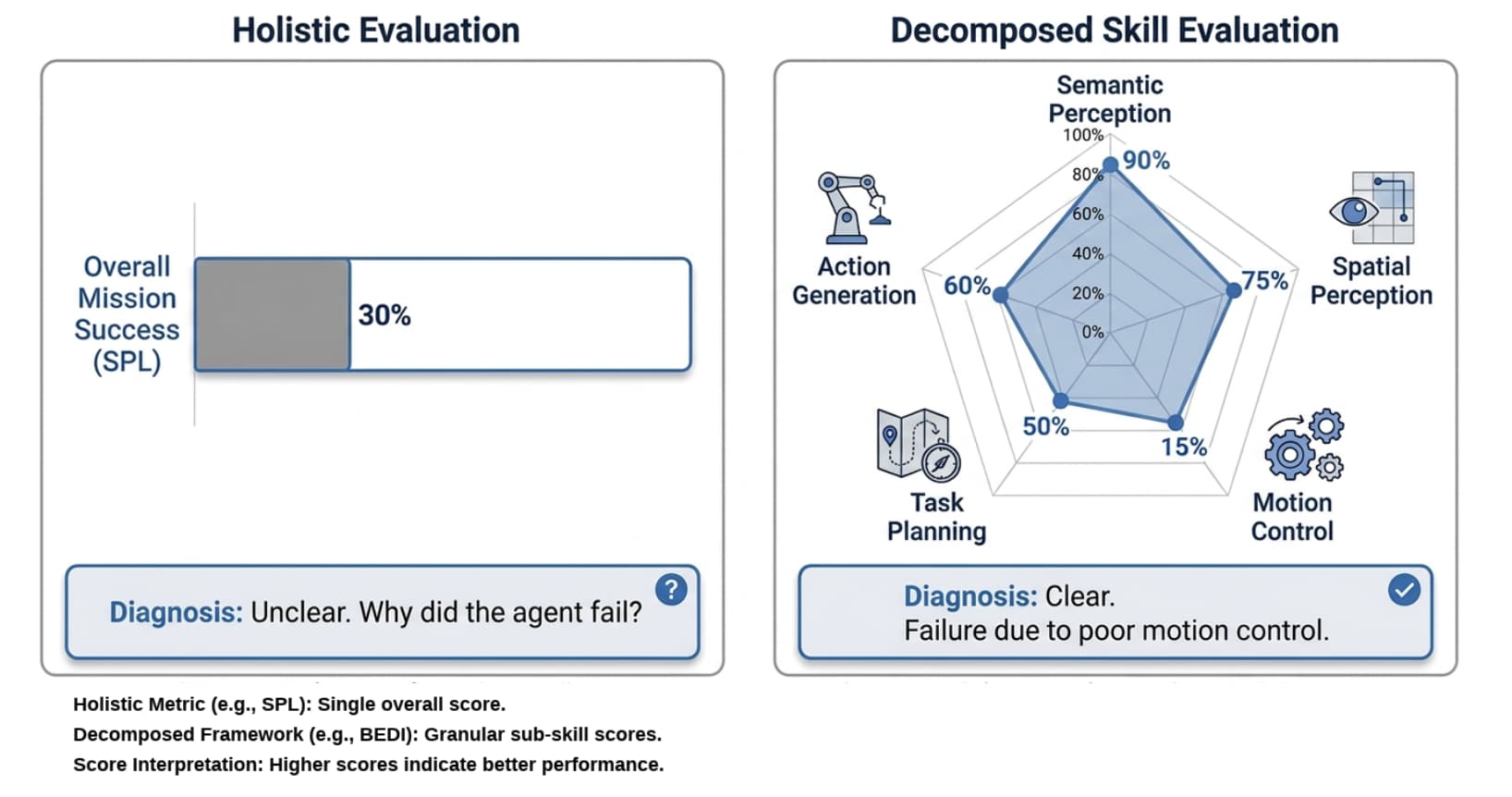}
    \caption{A comparison of evaluation paradigms. Holistic metrics like SPL provide a single score for task completion, while decomposed frameworks like BEDI offer a granular assessment of core sub-skills, enabling precise diagnosis of agent failures.}
    \label{fig:essential_resources__simulators,_datasets,_and_evaluation_auto_2}
\end{figure*}
 Instead of a single navigation score, BEDI dissects performance into six independently measurable sub-skills: semantic perception, spatial perception, motion control, tool utilization, task planning, and action generation. Performance is quantified using fine-grained metrics like Accuracy and Completeness Score, which offer deeper insights into where a model excels or fails, moving beyond the ambiguity of a low SPL score~\cite{bedi_benchmark}.

This trend toward analytical evaluation is reflected across the field, with many benchmarks introducing metrics tailored to specific sub-tasks and capabilities. To assess perception, models are evaluated on semantic segmentation using mean Intersection over Union (mIoU) or F1-score~\cite{disasterm3_2025, semantic_segmentation2024, postdisaster_evaluation2026}, on object detection using mean Average Precision (mAP), often with thresholds like AP50 and AP75 tailored for challenging small objects~\cite{uav_recognition2024, uav_human2021, uemm_air2021}, and on localization accuracy via Root Mean Squared Error (RMSE) of the estimated pose~\cite{multi_modality_ground_to_air}. For language and reasoning, performance is measured using text-similarity scores like BLEU, METEOR, CIDEr, and ROUGE-L~\cite{vrsbench2024, pixels_prose2025, rsgpt2023, sarlang-1m_benchmark}, with a rising trend of using advanced LLMs like GPT-4 for automated, reference-free assessment of open-ended answers~\cite{open3d_vqa2024, sarlang-1m_benchmark}. This decompositional philosophy has led to a proliferation of specialized metrics, including collision rates for social compliance~\cite{ha_vln2025}, sub-task success rates for long-horizon missions~\cite{long_horizon_vln2025}, accuracy scores for topological spatial reasoning~\cite{foundation_models2025}, quantitative measures of temporal understanding~\cite{dynamicvl2025}, and even metrics for agent trust under adversarial conditions~\cite{trust_based_assured2024}. This decomposition is crystallized in benchmarks like CognitiveDroneBench for high-level cognitive skills~\cite{cognitivedrone2025} and NAV-NUANCES, which tests atomic instructions~\cite{vision_language_navigation}, underscoring a collective move toward more precise and insightful evaluation.

\subsubsection{Contextualizing Evaluation: The Sim-to-Real Challenge}
The practical utility of any evaluation protocol is ultimately constrained by the sim-to-real gap, where performance on standard metrics can degrade significantly due to discrepancies in visual domains, action spaces, and unmodeled environmental dynamics~\cite{drone_simulators2019}. High scores in simulation do not guarantee real-world competence, a challenge starkly illustrated by early real-world VLN experiments. For example, one pioneering study documented a dramatic drop in Success Rate from 46.8\%, when using a pre-built map, to just 22.5\% when operating without prior mapping in a real-world setting, highlighting how sensitive even map-based agents are to real-world conditions~\cite{sim_to_real2020}. This performance gap is not limited to the simulation-to-reality transfer; it is also evident in the generalization between different simulated environments, where agents trained on the OpenFly benchmark saw their SR drop from 33.2\% in seen test environments to just 10.7\% in unseen ones~\cite{openfly2025}. This underscores that evaluation scores are highly context-dependent.

This challenge has directly motivated the development of more robust and diverse benchmarks designed to better predict real-world performance. The creation of platforms like OpenFly is a direct response to the data diversity bottleneck, which integrates multiple rendering engines such as Unreal Engine, GTA V, and Google Earth to train agents on a wide distribution of visual styles and environmental assets~\cite{openfly2025}. Similarly, the BEDI benchmark explicitly incorporates dynamic real-world video scenarios alongside high-fidelity simulation to test an agent's robustness to unmodeled, real-world dynamics~\cite{bedi_benchmark}. This field-wide recognition demonstrates a critical evolution in evaluation philosophy: robust and meaningful assessment requires not only sophisticated metrics but also testing across a broad and varied distribution of scenarios that faithfully represent the complexity of the real world.

\section{Core Challenge 1: The Sim-to-Real Gap} \label{sec:Core Challenge 1: The Sim-to-Real Gap}

This section provides a systematic analysis of the simulation-to-reality (sim-to-real) gap, a foundational challenge that limits the practical deployment of UAV-VLN agents. The discussion synthesizes key research on both the underlying causes of this performance degradation and the principal strategies developed to mitigate it. Our analysis is structured to first deconstruct the problem into three primary components identified in robotics literature~\cite{sim_to_real2020}---disparities in visual perception, physical dynamics, and environmental complexity---before surveying a corresponding set of solutions, including neural rendering, domain adaptation, and standardized deployment frameworks. As illustrated in Table~\ref{tab:sim2real_tree}, this structured breakdown of the sim-to-real gap in UAV vision-language navigation reinforces our analytical framework by mapping the core challenges to their respective mitigation strategies. The section begins by \textit{Characterizing the Reality Gap}, a foundational step that deconstructs this multifaceted challenge to establish a clear problem definition.
\begin{table*}[!htb]
\centering
\resizebox{\textwidth}{!}{%
\begin{tabular}{p{2.5cm}p{4cm}p{6cm}p{6cm}}
\toprule
\textbf{Aspect} & \textbf{Challenge / Solution} & \textbf{Description} & \textbf{Key Issues / Techniques} \\
\midrule
\multicolumn{4}{l}{\textbf{\textit{Characterizing the Reality Gap (The Problem)}}} \\
\midrule
\multirow{2}{*}{\parbox{2.5cm}{Visual Domain\\Shift}}
 & Rendering \& lighting discrepancies & Differences in rendering fidelity, textures, and lighting between simulators and real-world sensor data cause perception failures. & Real-world variations (motion blur, viewpoint, scale); partial observability from limited FoV; failures on reflective surfaces and adverse lighting. \\
\cmidrule(l){2-4}
 & Semantic understanding errors & Flawed environmental understanding from noisy depth, segmentation errors, and sensor noise profiles. & High cost of real-time semantic map maintenance; geographical and sensor resolution biases in datasets. \\
\midrule
\multirow{2}{*}{\parbox{2.5cm}{Dynamics\\Mismatch}}
 & Action space abstraction & Simplified discrete actions in simulation fail to capture continuous real-world physics (aerodynamics, actuator latency, wind). & Overestimation of safe velocities; mismatch between high-level decisions and low-level controller execution. \\
\cmidrule(l){2-4}
 & State estimation drift & Imperfect localization in GPS-denied environments using visual-inertial odometry or LiDAR; accumulated EKF and SLAM errors. & Sensor noise at high speed; processing latency; inaccurate calibration; single-policy controllers failing beyond training distribution. \\
\midrule
\multirow{2}{*}{\parbox{2.5cm}{Environmental\\Complexity Gap}}
 & Dynamic elements \& long-horizon & Simulators lack semantic richness and unpredictable real-world dynamics; cascading errors over long trajectories. & High collision rates; navigation deadlocks; 5.1\% (AI) vs.\ 80.8\% (human) SR in AerialVLN. \\
\cmidrule(l){2-4}
 & Operational constraints & Extreme environments (smoke, weather) and regulatory constraints (battery, airspace rules) invalidate simulation assumptions. & Multi-agent coordination introduces additional complexities: communication delays, security vulnerabilities. \\
\midrule
\multicolumn{4}{l}{\textbf{\textit{Bridging the Gap (The Solutions)}}} \\
\midrule
\multirow{2}{*}{\parbox{2.5cm}{Photorealistic\\Simulation}}
 & Neural rendering (3DGS) & Creates visually indistinguishable-from-reality environments from real-world captures for sample-efficient training. & GRaD-Nav, SINGER achieve direct sim-to-real policy transfer; OpenFly integrates 3DGS as one of multiple engines. \\
\cmidrule(l){2-4}
 & Advanced simulation platforms & NVIDIA Isaac Sim provides RTX rendering + PhysX 5 for parallelized RL training in photorealistic environments. & Executable 3DGS environments with object-level semantics and physics for realistic agent interaction. \\
\midrule
\multirow{2}{*}{\parbox{2.5cm}{Domain\\Adaptation}}
 & Domain randomization & Exposes agents to diverse visual and physical conditions during training to learn domain-invariant features. & Varied textures, lighting, dynamics; complements photorealistic simulation for additional robustness. \\
\cmidrule(l){2-4}
 & Action space abstraction & Subgoal models predict reachable waypoints; online self-supervised fine-tuning on robot using consistency losses. & Attention-level distillation aligns model focus with real-world cues; robust controllers enable direct transfer. \\
\midrule
\multirow{2}{*}{\parbox{2.5cm}{Deployment\\Frameworks}}
 & ROS-based integration & Model-agnostic frameworks (e.g., ROS4VSN) decouple navigation policy from hardware-specific drivers. & Standardized sensor integration, state estimation, and motion control; plug-and-play model comparison. \\
\cmidrule(l){2-4}
 & Multi-stage validation & Progressive SIL $\rightarrow$ HIL $\rightarrow$ real-world deployment pipeline for systematic verification before flight. & Safety-critical module integration; edge server offloading for real-time performance on resource-constrained drones. \\
\bottomrule
\end{tabular}}
\caption{A structured breakdown of the simulation-to-reality (sim-to-real) gap in UAV vision-language navigation. The upper section deconstructs the problem into three core components: visual domain shift, dynamics mismatch, and environmental complexity gap. The lower section outlines the corresponding solution strategies: photorealistic simulation via neural rendering, domain adaptation techniques, and systematic deployment frameworks.}
\label{tab:sim2real_tree}
\end{table*}

\subsection{Characterizing the Reality Gap} \label{subsec:Characterizing the Reality Gap}

The sim-to-real gap in UAV-VLN is a multi-faceted challenge arising from fundamental discrepancies between simulated training and real-world deployment across three primary axes: perceptual fidelity, physical dynamics, and environmental complexity.

\begin{figure*}[!htb]
    \centering
    \includegraphics[width=0.9\textwidth,height=0.5\textheight,keepaspectratio]{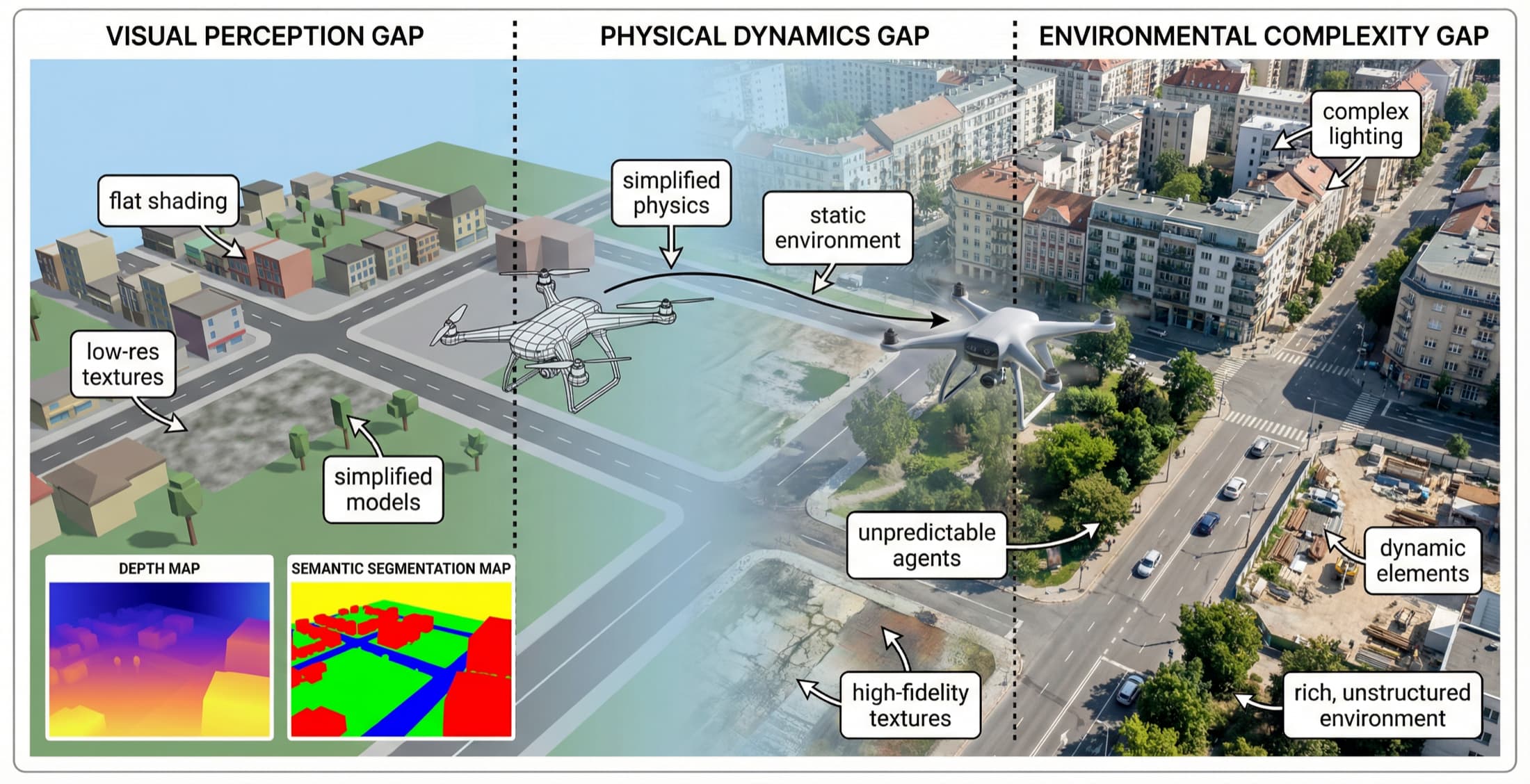} 
    \caption{A conceptual breakdown of the sim-to-real gap into its three constituent challenges: disparities in visual perception, physical dynamics, and environmental complexity.}
    \label{fig:core_challenge_1__the_sim-to-real_gap_auto_1}
\end{figure*}
 While training in simulation is the dominant paradigm driven by the high cost and impracticality of extensive real-world data collection~\cite{drone_simulators2019, agentic_uavs2025} the transfer of learned policies to physical hardware is often unsuccessful due to the inherent simplifications and abstractions in virtual worlds~\cite{robotic_navigation2025, uav_navigation2019, learning_based_navigation2026}. This discrepancy between the abstracted action space in simulation and the continuous physical world remains a critical bottleneck, limiting real-world generalization and the effective integration of vision-language features~\cite{drl_uav_navigation, grounded_vision2025, sim_to_real2020}. The scarcity of large-scale, high-quality visuomotor data from highly dynamic and unstable drones further exacerbates this challenge, making robust sim-to-real transfer a central research focus~\cite{singer2025}. This section characterizes the primary causes of this gap, providing a framework for understanding the technical challenges that subsequent solution-oriented sections address.

\subsubsection{Visual Domain Shift: The Photorealism and Perception Gap}
The most immediate and well-studied dimension of the reality gap is the visual domain shift, where discrepancies in rendering fidelity, lighting, and environmental textures between simulators and real-world sensor data cause catastrophic failures in perception models. The high-dimensional and variable nature of real-world RGB images, with significant variations in scale, viewpoint, rotation, motion blurs, and resolution changes, presents a persistent challenge for models trained on cleaner, often less diverse, simulated data~\cite{visuomotor_policies2026, enhancing_visual2025, uav_human2021, multimodal_registration2024, hierarchical_image_matching}. These visual discrepancies can cause critical misalignments between vision and language modalities~\cite{multimodal_pretrained_knowledge2023} and lead to out-of-domain inputs that cause VLM prediction inaccuracies~\cite{search_tta2025}, perception errors under varying observation angles~\cite{spatial_representation2025}, and insufficient spatiotemporal reasoning~\cite{aerial_vision2025}. Even advanced scene representations like 3D Gaussian Splatting, while photorealistic, can lack the fine-grained object-level semantics and physical executability required for complex interaction, creating a subtle but critical gap between representation and reality~\cite{executable_3d_gaussian2024, reasongrounder2025}.

This perceptual gap is exacerbated by a range of factors including partial observability due to the limited field of view of monocular cameras~\cite{sim_to_real2024, vision_based2024}, noisy depth reconstruction artifacts~\cite{lavira2025, memory_based_deep2018}, and flawed environmental understanding. Such flaws often stem from inaccurate semantic segmentation or object misclassification~\cite{goal_oriented_semantic_exploration2020, zero_shot_object_centric2025}, or from the high computational and memory costs of maintaining accurate semantic maps in real time~\cite{semantic_mapping2025, semantic_octree_mapping, morphonavi2025}. Models often fail in visually challenging real-world conditions that are underrepresented in synthetic datasets, such as highly reflective surfaces, adverse lighting~\cite{model_free_uav2024}, or featureless environments like deserts and snow-covered areas~\cite{vision_based_navigation2023, autonomous_decision2025}. Even with high-fidelity rendering, subtle differences in lighting models and sensor noise profiles can lead to poor generalization~\cite{vision_language_navigation2021, on_device_learning2024}, a problem compounded by geographical and sensor resolution biases in existing datasets~\cite{multi_modality_ground_to_air}. Modern platforms like OpenFly attempt to mitigate this by integrating multiple rendering engines to enhance data diversity~\cite{openfly2025, openfly2024}, but challenging benchmarks like AerialVLN still reveal that current simulation fidelity is insufficient for robust transfer~\cite{aerialvln2023}.

\subsubsection{Dynamics Mismatch: From Abstracted Actions to Real-World Physics}
A second critical gap arises from the abstraction of dynamics, where the simplified, often discrete, action spaces and perfect actuation assumptions in simulators fail to model the continuous physics of UAV flight, including aerodynamics, actuator latency, and localization drift~\cite{waypoint_models2021}. A foundational investigation demonstrated a significant performance drop when moving from simulated to real-world deployment, attributing it partly to the mismatch between high-level agent decisions and the low-level controllers executing them~\cite{sim_to_real2020}. This problem is magnified for UAVs due to their unstable 3D dynamics and susceptibility to unmodeled external forces like wind~\cite{explainable_deep_reinforcement2021, trajectory_planning_drone_landing, real-time_cooperative2024, drone_landing2026}, with idealized assumptions leading to overestimation of safe velocities~\cite{roofline_model2022}.

The gap is further widened by challenges in real-world state estimation, especially in GPS-denied or poorly lit environments where localization relies on imperfect sensor data from visual-inertial or LiDAR odometry~\cite{control_method_uavs, short_distance_uavs2025, gps_denied_ibvs2025, collaborative_mapping2026, hierarchical_multi_uav2024, heightmap_gradients}. Common sensor fusion algorithms like the Extended Kalman Filter (EKF) are limited in handling nonlinear systems and can accumulate significant errors over time~\cite{uav_positioning_gps_denial}. Similarly, SLAM systems are prone to error drift that degrades performance, particularly in dynamic or featureless environments~\cite{simultaneous_control_uavs2024}. At high speeds, these issues are compounded by increased sensor noise, processing latency from hardware-specific inference pipelines~\cite{lightweight_drone2024, vision_transformers2025}, inaccurate sensor calibration~\cite{real_time_multi_modal2022,zeng2025yoco}, and image distortion that affects positioning accuracy~\cite{path_planning_algorithm}. Even state-of-the-art models like FlightGPT output high-level waypoints, implicitly relying on a separate, perfectly-tuned controller to manage the dynamics gap~\cite{flightgpt2024}. This reliance remains a source of deployment failure, as single-policy controllers often fail to generalize beyond their training distribution~\cite{pareto_optimal2026, uav_control2025}.

\subsubsection{Environmental and Task Complexity Gap}
The third dimension of the reality gap is environmental and task complexity, where simulators often lack the semantic richness, unpredictable dynamic elements, and long-horizon challenges of real-world scenarios, causing failures in spatiotemporal reasoning. The real world is not a static collection of assets; it is a dynamic environment with moving objects, complex structures, and unpredictable events that are difficult to model exhaustively~\cite{asma2026, uav_vln2025, uav_vln2025, path_planning2022, prompt_informed_reinforcement2025}. Unexpected obstructions can create discrepancies between navigation instructions and the actual environmental topology, causing planning failures~\cite{navigating_beyond2024}. This leads to high collision rates and navigation failures like deadlocks in real-world tests, even for models that perform well in simulation~\cite{uav_on2025, embodied_gap2025, reprohrl2023}. Scenarios with sudden crowd convergence or navigation through tight corridors present dynamic challenges rarely captured in simulation~\cite{ha_vln2025}. Extreme environments, such as those in firefighting, fundamentally alter perception and invalidate typical simulation assumptions due to elements like smoke~\cite{air_ground_robots2025, human_centric_uav2024}. Furthermore, real-world operation must account for physical and regulatory constraints, including limited battery life, adverse weather, signal loss, and aviation rules~\cite{decision_network2026, uav_search_rescue2023, ground_aerial_transportation2025, urbanvideo_bench}. The large performance delta between humans (80.8\% SR) and AI agents (5.1\% SR) in the AerialVLN benchmark~\cite{aerialvln2023} illustrates how small, cumulative errors cascade into mission failure over long paths~\cite{aerial_vision2025G2025}. This highlights a systemic gap in robust, long-term reasoning known limitation of large models~\cite{embodied_ai2024Y2024}  just instantaneous perception or control. This challenge becomes even more pronounced when coordinating multi-agent swarms, which introduces further real-world complexities like ensuring safe choreographies, communication delays, and security vulnerabilities~\cite{uxv_swarms2025, uav_vision_language2024, uav_swarms2024, swarm_gpt2023}.

\subsection{Bridging the Gap: Photorealism and Deployment Frameworks} \label{subsec:Bridging the Gap: Photorealism and Deployment Frameworks}

Successfully transferring vision-language navigation agents from simulation to reality requires a dual approach: enhancing simulation fidelity to minimize the perceptual gap, and developing robust software frameworks to bridge the action and control gap. The field is evolving from domain randomization and action-space abstraction towards end-to-end training in photorealistic environments and systematic, framework-driven deployment. While training in simulation is essential for its safety and scalability~\cite{drone_simulators2019}, policies learned in this way often fail when deployed on physical hardware due to inherent discrepancies in visual appearance, physics, and environmental complexity~\cite{sim_to_real2020, drone_landing2026}. Closing this sim-to-real gap is therefore a central challenge for creating truly autonomous aerial agents.

\subsubsection{Photorealistic Simulation with Neural Rendering}
The adoption of high-fidelity rendering techniques, particularly 3D Gaussian Splatting (3DGS), is a primary strategy to minimize the perceptual sim-to-real gap, enabling more direct, sample-efficient, and end-to-end training of visuomotor policies. By creating virtual worlds that are visually almost indistinguishable from reality, this approach aims to make the data distribution seen during training as close as possible to that encountered during deployment. A new wave of simulators leverages neural rendering techniques like 3DGS to construct environments with unprecedented fidelity from real-world captures~\cite{grad_nav2025, embodied_gap2025}. This realism allows for the training of lightweight Vision-Language-Action (VLA) models that map raw sensory inputs directly to low-level control commands, bypassing traditional intermediate representations and achieving high success rates in both simulation and real-world tests~\cite{singer2025}. Platforms like OpenFly exemplify this trend by integrating multiple rendering engines, including 3DGS, to generate diverse and photorealistic training data~\cite{openfly2024}. This technological shift is moving beyond static scenes to create executable environments with object-level semantics and physics, allowing for realistic agent interaction and closing the gap between representation and reality~\cite{executable_3d_gaussian2024}. The use of advanced platforms like NVIDIA Isaac Sim further accelerates this by providing both photorealistic rendering and robust physics simulation, enabling parallelized training of complex policies~\cite{navrl2025}.

\subsubsection{Adaptation via Domain Randomization and Action Abstraction}
When perfect simulation fidelity is unattainable or insufficient, agents must bridge the reality gap through explicit adaptation techniques like domain randomization for perception and action space abstraction for control. This approach accepts that a gap will likely persist and builds mechanisms to make policies robust to this domain shift. Domain randomization exposes the agent to a wide variety of visual and physical conditions during training such as different textures, lighting, and dynamics forcing it to learn domain-invariant features~\cite{sim_to_real2020, task_oriented_flying2025}. This strategy is often used to complement photorealistic simulation, providing an additional layer of robustness against unmodeled variations~\cite{singer2025}. To address the dynamics gap, action space abstraction translates an agent's high-level, often discrete, decisions (e.g., 'move to the next viewpoint') into continuous, low-level commands compatible with a real robot's controller. A prime example is the use of a subgoal model that predicts a set of reachable waypoints for the high-level policy to select from, effectively bridging the abstract navigation graph of simulation with the continuous physical world~\cite{sim_to_real2020}. Other adaptation strategies include online, self-supervised fine-tuning on the robot using consistency losses derived from onboard odometry~\cite{on_device_learning2024}, using attention-level distillation to align a model's focus with relevant real-world cues~\cite{vision_language_attention_distillation}, or designing policies with advanced controllers that are inherently robust to model discrepancies, enabling direct transfer without randomization~\cite{learning_based_navigation2026}.

\subsubsection{Systematizing Deployment with Robotic Frameworks}
Practical and scalable sim-to-real transfer relies on robust, modular software frameworks, typically based on the Robot Operating System (ROS), which decouple the navigation policy from hardware-specific drivers and provide essential functionalities for systematic deployment, testing, and error correction. These engineering platforms are critical for translating algorithmic advances into real-world success. Model-agnostic frameworks such as ROS4VSN provide a plug-and-play architecture for integrating different navigation models with physical robots, handling the complexities of sensor integration, state estimation, and low-level motion control~\cite{visual_semantic_navigation2025}. By abstracting away hardware specifics such as a Pixracer flight controller, an Intel Realsense camera, and a Nvidia Jetson Orin Nano for onboard computation~\cite{grad_nav2025, racevla2025, dialogue_px42025} these frameworks enable rigorous, apples-to-apples comparisons of different AI models in the real world. This systematic approach is crucial for identifying performance bottlenecks and closing the research loop. The deployment process is often structured into a multi-stage pipeline, including Software-in-the-Loop (SIL) and Hardware-in-the-Loop (HIL) testing, to progressively validate the system before full real-world deployment~\cite{autonomous_drone_testing2025}.

\begin{figure*}[!htb]
    \centering
    \includegraphics[width=0.9\textwidth,height=0.5\textheight,keepaspectratio]{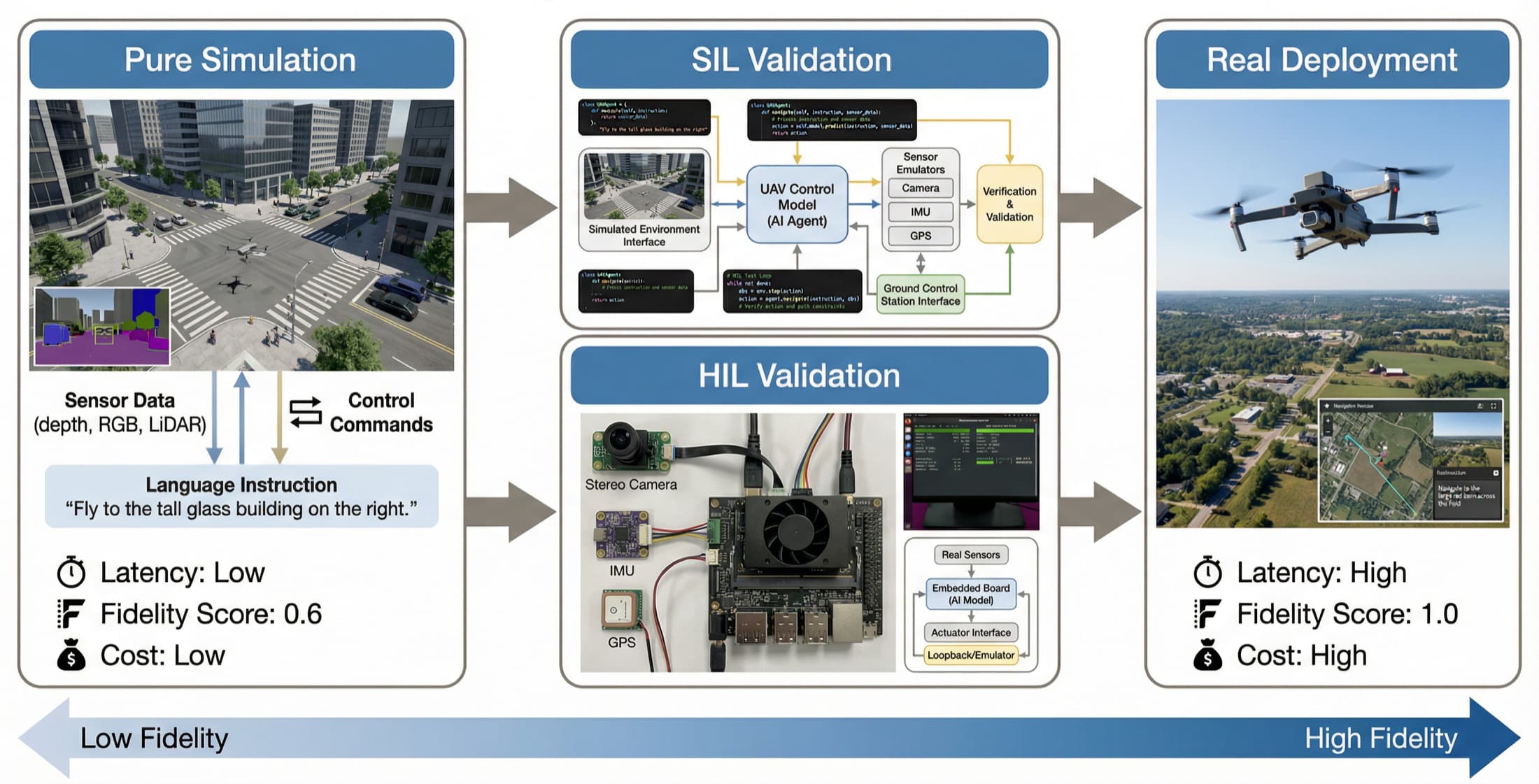}
    \caption{The typical multi-stage pipeline for sim-to-real transfer, progressing from pure simulation to Software-in-the-Loop (SIL) and Hardware-in-the-Loop (HIL) validation before final deployment on a physical UAV.}
    \label{fig:core_challenge_1__the_sim-to-real_gap_auto_2}
\end{figure*}
 These frameworks also enable the integration of safety-critical modules, such as collision-avoidance filters~\cite{swarm_gpt2023}, and facilitate architectures that offload heavy computation to edge servers to maintain real-time performance on resource-constrained drones~\cite{autonomous_navigation2025}.

\section{Core Challenge 2: Robustness, Safety, and Efficiency} \label{sec:Core Challenge 2: Robustness, Safety, and Efficiency}

Transitioning UAV-VLN from controlled simulations to dynamic, real-world environments necessitates confronting a triad of interdependent challenges: achieving robust performance under uncertainty, guaranteeing operational safety, and enabling efficient deployment on resource-constrained platforms. This section moves beyond algorithmic efficacy to focus on the practical impediments to deployment, examining the mechanisms for reliable decision-making amidst perceptual and linguistic ambiguity, the formal and learned methods for assuring safety, and the architectural strategies for deploying large models under strict hardware constraints. We begin by addressing the cognitive foundation of this triad, exploring how agents must achieve robust \textbf{Reasoning under Uncertainty and Ambiguity} to interpret and act upon incomplete or noisy information inherent in dynamic missions, such as navigating unmapped outdoor environments or operating in multi-agent systems with potentially untrustworthy data. This capability relies on techniques like leveraging large language models (LLMs) for predictive reasoning and context-aware decision-making, implementing trust-based frameworks for assured sensor fusion, and performing introspective self-verification for error recovery during long-horizon tasks \cite{reasoning_unseen2024,agentic_uavs2025,agentic_robot2025,trust_based_assured2024}

\subsection{Reasoning under Uncertainty and Ambiguity} \label{subsec:Reasoning under Uncertainty and Ambiguity}

Effectively reasoning under uncertainty requires agents to move beyond reactive control by embedding structured reasoning mechanisms. This is achieved through three primary routes: decomposing ambiguous instructions into procedural steps, grounding language in explicit world representations, and leveraging interactive clarification or the powerful implicit priors of large foundation models.

\begin{figure*}[!htb]
    \centering
    \includegraphics[width=0.9\textwidth,height=0.5\textheight,keepaspectratio]{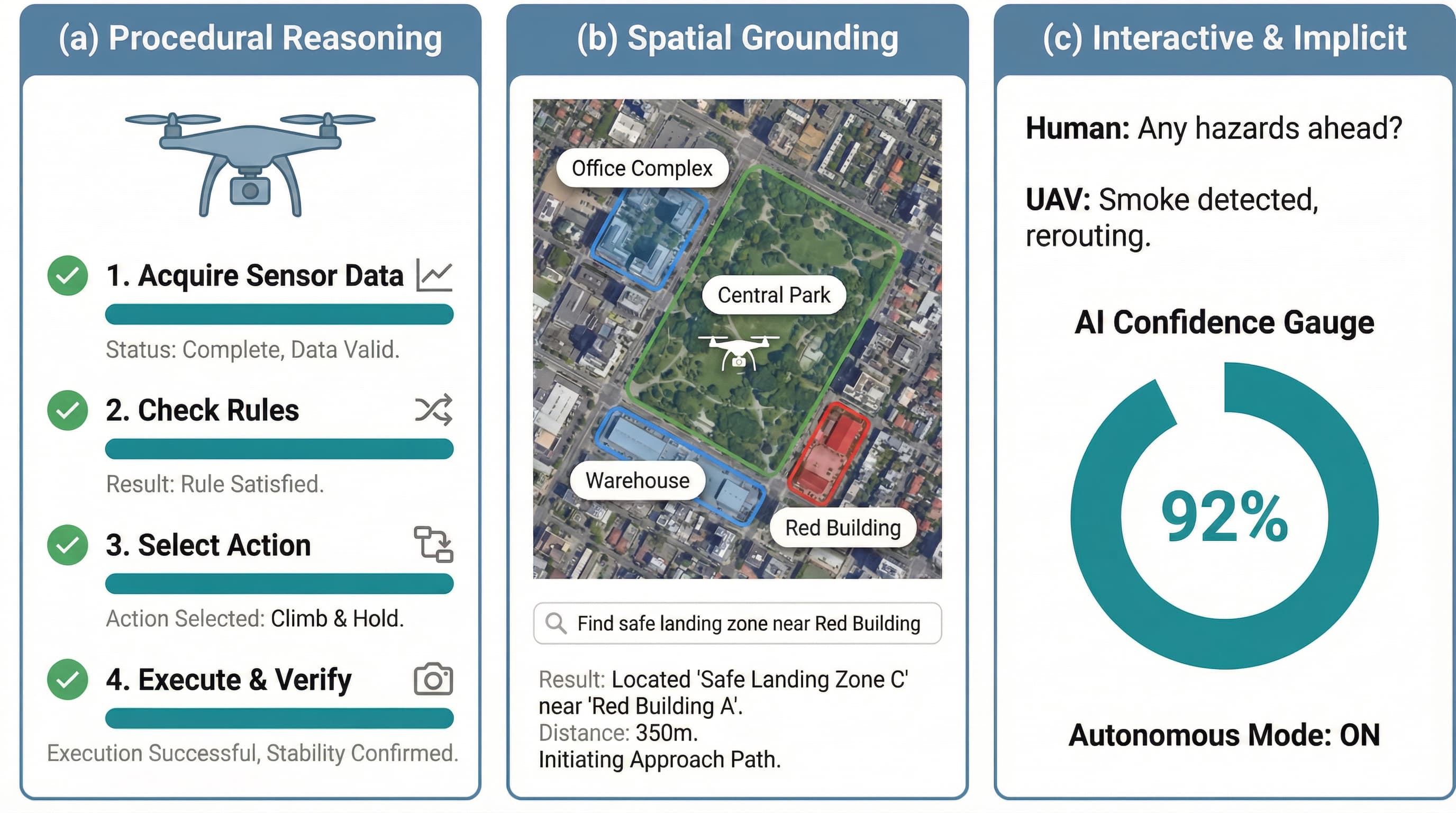}
    \caption{A comparison of three primary strategies for reasoning under uncertainty. Procedural reasoning decomposes instructions into steps, spatial grounding uses explicit world models for disambiguation, and interactive/implicit methods either query a human for clarification or rely on end-to-end foundation models.}
    \label{fig:core_challenge_2__robustness,_safety,_and_efficiency_auto_1}
\end{figure*}
 Agents operating in dynamic, unpredictable outdoor environments are constantly confronted with uncertainty stemming from ambiguous natural language instructions and partially observable, noisy perception~\cite{anti_jamming2025, uav_navigation2019, control_method_uavs, grounded_vision2025, navrl2025, sim_to_real2024, real-time_cooperative2024, vl_nav2025, spatial_assisted2024}. These perceptual challenges are exacerbated by real-world conditions like varying viewpoints and motion blurs~\cite{uav_human2021}, adverse weather~\cite{multi_modal_uav2024}, and dynamic obstacles or unpredictable targets in urban airspace~\cite{safe_uav_operation, vision_based_navigation2023, path_planning2022, multi_uav_adaptive2024, drone_landing2026}. These difficulties, which include semantic parsing, visual grounding, and robust planning~\cite{uav_vln2025, uav_vln2025, multimodal_spatial_reasoning2025, multimodal_uav2025, embodied_gap2025, low_cost2024}, are significant barriers to reliable autonomy, particularly as current VLMs still struggle with tasks like regression and handling non-RGB data~\cite{visual_language_models2025}. The severity of this gap is highlighted by benchmarks where human pilots achieve success rates of 80.8\% while the best models lag at just 5.1\%~\cite{aerialvln2023}, a performance drop attributed directly to challenges in handling natural language ambiguity and perception in unseen environments~\cite{citynav2025}. These difficulties are further amplified in multi-agent systems, where uncertainty is compounded by non-stationarity, communication latency, and partial observability~\cite{multi_uav2026, scalable_cooperative2023, uav_control2025, coordfield2025, swarms_uavs2022, task_offloading2024, dnn_task2024, uav_swarm2024}. Addressing them is critical for moving beyond the limitations of purely reactive systems and toward architectures that can deliberate, ground, and act robustly despite incomplete information~\cite{llm_robotic_autonomy2025, embodied_ai2024, long_horizon_vln2025}.

\subsubsection{Procedural Reasoning for Decomposing Ambiguous Instructions}
To combat linguistic ambiguity, modern agents employ procedural reasoning frameworks, such as Chain-of-Thought (CoT), to decompose high-level, abstract instructions into a sequence of explicit, verifiable sub-goals and actions. Early navigation models often struggled with underspecified or complex commands, particularly when instructions involve intricate causal reasoning or refer to objects not immediately visible~\cite{vision_language_navigation2021, agrivln2024, goal_oriented_semantic_exploration2020, urbanvideo_bench, indooruav2025, uav_path2023, navigating_beyond2024}. In response, models like FlightGPT~\cite{flightgpt2024}, SkyVLN~\cite{skyvln2025}, and others adopting ReAct-style workflows~\cite{agentic_uavs2025} generate a structured, textual reasoning process before predicting actions. This breaks down a command like "fly to the red building behind the park" into discrete, logical steps, a trend seen across numerous frameworks that parse free-form text into actionable plans for geospatial tasks~\cite{lm_nav2022, llm_enhanced_path2025, hierarchical_language_models2025, streetviewllm2024, aerial_agents2025, nextgen_llm_uav2025}. This approach can also be used to interpret descriptions of dynamic environmental changes to update navigation plans~\cite{contextualized_drone2025} or to incorporate visual prompts that proactively reduce instructional ambiguity~\cite{vision_language2024}. The same principle extends to multi-UAV swarms, where a central LLM interprets a high-level command and decomposes it into a synchronized plan for multiple agents~\cite{tacos2025, llm_mars2023}. This principle is also realized through hierarchical planning, where agents like CityNavAgent~\cite{citynavagent2025} break down missions into landmark, object, and motion-level sub-tasks, leveraging multi-step reasoning~\cite{open_vocabulary_indoor2024} to infer semantic relevance~\cite{l3mvn2023}. However, this deliberative approach introduces a trade-off between robustness and efficiency, as the generative reasoning step can increase latency and is susceptible to hallucination or logical failures~\cite{large_language_models2025P2025, convoi_context_aware, leviosa_trajectory_generation}. For instance, even advanced LLM-enhanced models exhibit weaknesses with complex or long-horizon instructions~\cite{uav_vla2025, mapping_instructions2024} and struggle with numerical comprehension or fine-grained temporal understanding~\cite{vision_language_navigation, embodied_navigation2024, dynamicvl2025}.

\subsubsection{Spatial Grounding via Structured World Representations}
To resolve spatial and semantic ambiguity in cluttered environments, agents construct explicit world representations such as language-grounded maps and hierarchical scene graphs that serve as a persistent memory for disambiguating landmarks and reasoning about geospatial relationships. Robust reasoning is contingent on a robust world representation that achieves deep visual-language alignment for interpreting complex scenes and tracking navigation state over long trajectories~\cite{aquila_visual_language_model, imaia_assistant, vlca2026, target_grounded2023, fine_grained_alignment2024, tina2024, ok_robot2024}. This memory is essential for handling perceptual challenges like occlusions, implicit instructions, and visually similar distractors~\cite{trackvla2025J2025, reasongrounder2025}, and can be improved by explicitly modeling spatial and semantic relations to resolve ambiguous queries~\cite{open_vocabulary_object_grounding}. Foundational techniques like VLMaps fuse visual-language model embeddings with 3D reconstructions to create queryable maps~\cite{visual_language_maps2023}. More sophisticated architectures, exemplified by GeoNav~\cite{geonav2025} and NavAgent~\cite{navagent_urban2024}, employ multi-layered spatial memory, such as a dynamic topological map or an episodic world model~\cite{trivla2025}, to enable a human-like coarse-to-fine search strategy. These methods integrate VLM reasoning with geometric priors to ensure commands are both semantically and spatially valid~\cite{soranav2025}, tackling challenges like distinguishing visually similar objects~\cite{Liu_2025_ICCV}, identifying complex topological relations~\cite{foundation_models2025}, and reasoning over complex spatial layouts~\cite{spatial_representation2024, geotext1652_2024, aeroreformer2025, open3d_vqa2024}. Other systems formalize this process by maintaining an explicit probabilistic model of the environment. For instance, a target probability map can be updated using a Bayesian framework during a cooperative search task~\cite{multi_uav_cooperative_search}, a cognitive map can be updated to handle sensor inaccuracies~\cite{multi_uav_cooperative2024}, or latent environmental dynamics can be inferred with an Adaptive Extended Kalman Filter~\cite{multi_uav_planning2022}.
\begin{equation}
p_i^{t+1}(k_x, k_y) = \frac{P_{i,D}^z p_i^t(k_x, k_y)}{P_{i,D}^z p_i^t(k_x, k_y) + P_{i,F}^z (1 - p_i^t(k_x, k_y))}
\end{equation}
where $p_i^t(k_x, k_y)$ is the prior probability of a target in cell $(k_x, k_y)$, and $P_{i,D}^z$ and $P_{i,F}^z$ are the detection and false alarm probabilities, respectively. The primary trade-off involves high computational costs and a high sensitivity to odometry or reconstruction errors, which can corrupt the map~\cite{visual_spatial_geometric2023, collaborative_mapping2026, semantic_mapping2025}. This problem can be mitigated by fusing visual, IMU, and GNSS data through factor graph optimization~\cite{aerial_ground2022} or by leveraging high-resolution sensor data to reduce ambiguity in challenging environments~\cite{multi_modality_ground_to_air}.

\subsubsection{Interactive and Implicit Methods for Uncertainty Resolution}
A diverging trend addresses uncertainty either by offloading ambiguity resolution to a human user through interactive dialogue or by leveraging the potent, implicit priors of end-to-end Vision-Language-Action (VLA) models that map perception directly to control. The first approach forms a human-in-the-loop system where the agent can actively seek clarification, as exemplified by the Aerial Vision-and-Dialog Navigation (AVDN) benchmark~\cite{aerial_vision_dialog2024}. This strategy acknowledges the trade-offs between full autonomy and operator mental load, providing a collaborative framework for complex decision-making~\cite{teleoperation_system2023, human_drone2025}. In contrast, the second approach moves toward greater autonomy by using powerful foundation models to handle ambiguity implicitly. Lightweight VLAs like Grad-Nav~\cite{grad_nav2025}, RaceVLA~\cite{racevla2025}, and TrackVLA~\cite{trackvla2025} are trained end-to-end to map pixels and text directly to low-level motor commands, tightly coupling recognition and planning to reduce error accumulation. Similarly, diffusion policies can implicitly handle ambiguous instructions by learning a distribution over multiple equivalent paths~\cite{dagger_diffusion2021}. These models implicitly learn to handle ambiguous situations by drawing on the powerful commonsense knowledge and generalization capabilities embedded within pretrained VLMs~\cite{exploration_soft_constraints2023, reasongrounder2025, rt2_2023}, and can even learn to conform to social conventions in crowded spaces~\cite{ha_vln2025}. Uncertainty can also be handled within the learning paradigm, for instance by using privileged learning where a critic with access to the true state guides an actor that operates on noisy observations~\cite{vision_based2024}, using LLM-generated semantic feedback within an RL loop~\cite{prompt_informed_reinforcement2025}, or dynamically adjusting viewpoints to reduce occlusions~\cite{virtual_view2025}. This technical route trades the interpretability of structured reasoning for the efficiency and reactivity of direct policy learning. However, benchmarks like BEDI reveal that even state-of-the-art foundation models perform poorly on many UAV-specific perception tasks, partly due to their "black-box" nature and reliance on training data~\cite{bedi_benchmark, uav_navigation_xai2024}, motivating both interactive and more powerful end-to-end approaches to circumvent these persistent weaknesses.

\subsection{Safety Assurance for Autonomous Flight} \label{subsec:Safety Assurance for Autonomous Flight}

Ensuring safety for autonomous flight in vision-language navigation requires a shift from static, reactive obstacle avoidance to dynamic, context-aware safety frameworks. As agents are deployed in complex, unpredictable real-world environments, the ability to manage uncertainty and guarantee safe operation becomes paramount~\cite{control_method_uavs, anti_jamming2025}. Emerging techniques address this critical need by integrating formal guarantees from control theory, learned predictive models that anticipate hazards, and verifiable decision-making architectures that foster trust and enable certification.

\begin{figure*}[!htb]
    \centering
    \includegraphics[width=0.9\textwidth,height=0.5\textheight,keepaspectratio]{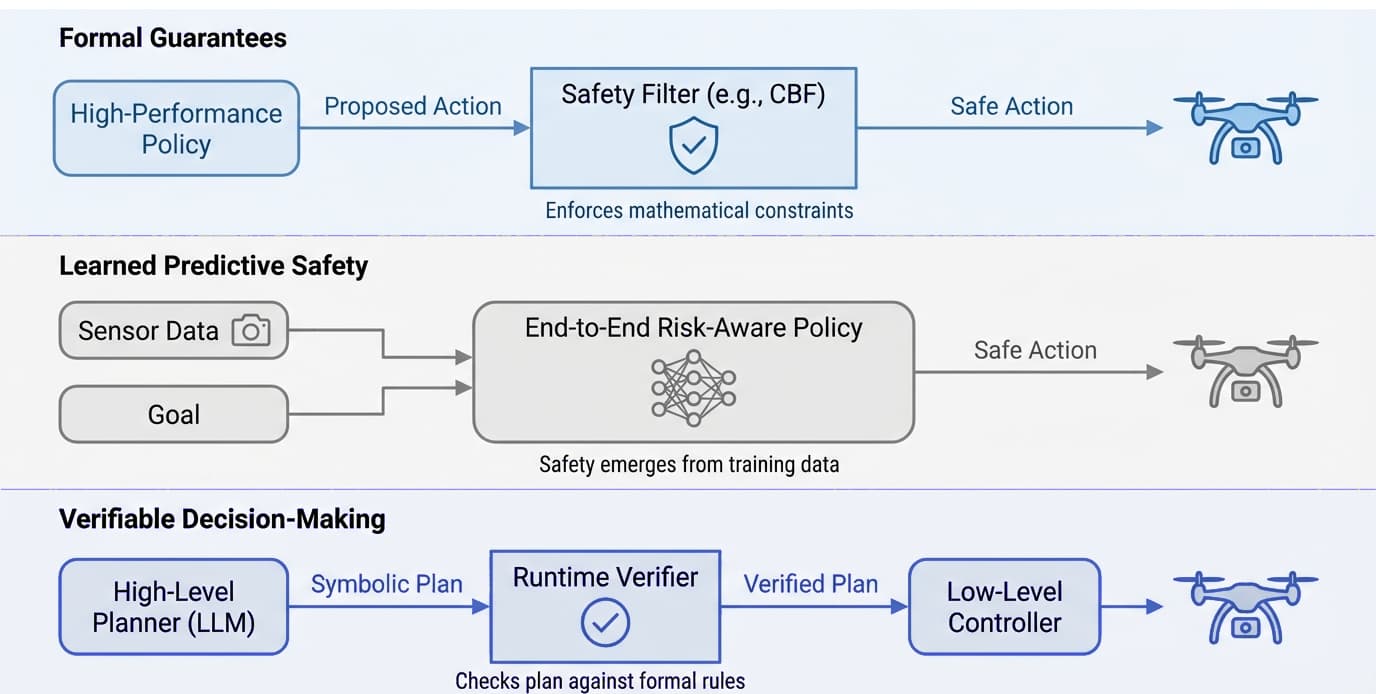}
    \caption{Three architectural paradigms for safety assurance in autonomous flight. Formal methods use explicit safety filters, learned approaches embed safety into an end-to-end policy, and verifiable systems use logic-based oversight to validate high-level plans before execution.}
    \label{fig:core_challenge_2__robustness,_safety,_and_efficiency_auto_2}
\end{figure*}
 These advanced methods aim to manage the inherent risks of operating in dynamic airspace alongside humans and other agents, moving beyond simple collision avoidance to a more holistic concept of operational safety.

\subsubsection{Formal Safety Guarantees via Control-Theoretic Methods}
A primary technical route for safety assurance involves imposing explicit mathematical constraints on the UAV's control actions using formal methods like Control Barrier Functions (CBFs), which provide provable safety certificates by maintaining a dynamically adjusted safety margin from obstacles. This approach defines a safe set of states, $\mathcal{C} = \{x \in \mathbb{R}^n | h(x) \ge 0\}$, where $h(x)$ is the CBF, and synthesizes a controller to ensure the system remains within this set. For a system with dynamics $\dot{x} = f(x) + g(x)u$, any valid control input $u$ must satisfy the constraint:
\begin{equation}
L_f h(x) + L_g h(x)u + \alpha(h(x)) \ge 0
\end{equation}
where $L_f h(x)$ and $L_g h(x)$ are the Lie derivatives of $h$ along the system dynamics, and $\alpha$ is a class $\mathcal{K}$ function. A leading example is the Adaptive Safety Margin Algorithm (ASMA), which integrates scene-aware CBFs within a Model Predictive Control (MPC) framework to dynamically adjust safety margins based on environmental context, significantly improving success rates in cluttered scenes~\cite{asma2026, asma2026}. This paradigm of enforcing mathematically-grounded safety constraints is also realized through related formalisms, such as the velocity obstacle concept used in safety shields~\cite{navrl2025}, model-based safety filters for multi-agent systems~\cite{swarm_gpt2023}, and combined CBF-Control Lyapunov Function (CLF) strategies for balancing safety and performance objectives~\cite{distributed_localization2024}. This control-theoretic philosophy provides a strong foundation for building certifiably safe autonomous systems~\cite{spatial_assisted2024}.

\subsubsection{Implicit and Learned Predictive Safety}
An alternative approach embeds safety implicitly within end-to-end learning frameworks, training agents to anticipate and avoid hazards by using learned collision predictors or by shaping behavior through risk-aware reward functions. This is particularly effective in scenarios where complex sensory data makes explicit modeling of all potential hazards intractable. Instead of relying on rigid, pre-defined constraints, these methods learn emergent safe behavior from data. For instance, some models employ a context estimator to predict collision probabilities from perceptual inputs, directly informing the control policy to execute safe navigation actions~\cite{grad_nav2025}. A complementary strategy involves shaping agent behavior within a reinforcement learning paradigm. By designing reward functions that heavily penalize unsafe actions or proximity to hazards, agents can learn policies that are inherently cautious and compliant, achieving high success rates with zero collisions or regulatory violations in complex simulations~\cite{safe_uav2025, uav_cluster_reinforcement_learning, autonomous_emergency_landing2026}. This concept of risk awareness can be extended to interpret linguistic cues; for example, by generating semantic cost maps from natural language instructions, where phrases like "be careful around the construction site" translate into high-cost regions to be avoided, effectively creating a map of perceived risk~\cite{language_as_cost2025}. Other methods focus on ensuring safety at the planning level by using learning-based techniques to filter out unsafe or hallucinated high-level commands generated by LLMs before they are executed~\cite{aermani_vlm2025}.

\subsubsection{Verifiable and Interpretable Safety Decision-Making}
To bridge the gap between the strong empirical performance of learning-based models and the need for trust in real-world deployment, a critical research thrust focuses on making safety mechanisms transparent and verifiable. This is achieved through interpretable models, runtime verification, and hybrid architectures that ensure safety-critical decisions are auditable. One approach builds interpretability directly into the model's architecture. For example, a Decision Network can provide clear, probabilistic explanations for its actions by modeling the environment with chance nodes, decisions, and utilities, allowing a human supervisor to understand the rationale behind a safety-critical choice~\cite{decision_network2026}. This drive for explainability is shared by frameworks that use techniques like Monte Carlo Tree Search within an XAI framework to generate transparent and optimizable trajectories~\cite{trajectory_optimization2024, safe_uav_operation}. A second approach uses external, formal verification at runtime. Rule-Based Runtime Verification (RBRV) systems employ ontologies to represent safety constraints and logical rules, continuously monitoring the UAV's behavior to ensure it complies with formalized safety protocols~\cite{rule_based_verification}. A growing trend is toward hybrid systems that combine the perceptual power of neural networks with the logical rigor of symbolic reasoning. Neuro-symbolic frameworks provide verifiable safety assessments by translating neural outputs into a symbolic domain where logical inference can be applied~\cite{neuro_symbolic2026}. Similarly, hierarchical architectures like TACOS separate high-level LLM-based reasoning from a low-level, safety-constrained supervisor, ensuring that ambitious plans do not violate fundamental safety rules~\cite{tacos2025}. This synthesis of learning and logic is crucial for building certifiably safe and trustworthy autonomous systems.

\subsection{Efficient Onboard Model Deployment} \label{subsec:Efficient Onboard Model Deployment}

Efficiently deploying large foundation models on resource-constrained UAVs is a primary bottleneck for real-world autonomy, necessitating a shift from brute-force computation to a multi-pronged strategy combining parameter-efficient adaptation, innovative model architectures, and hardware-aware system integration. The significant computational and memory demands of modern vision-language models often clash with the stringent limitations on power and payload inherent to aerial platforms~\cite{vision_language_action_models, vision_language_action2026, autonomous_uav_navigation2022, foundation_model2025, foundation_models2023, large_language_models2024, pure_vla2025, urban_monitoring2025}. High computational resource requirements for multi-modal fusion, processing high-resolution imagery~\cite{aquila_visual_language_model}, and overcoming model latency for real-time inference remain critical deployment challenges~\cite{securing_skies2025, airvista_ii2025, citynavagent2025, uav_trajectory2025, multimodal_registration2024, fm_planner2025, vision_language_models2025}, often forcing a reliance on ground computers for heavy processing and introducing communication latency~\cite{semantic_scene2024}. This is especially true in multi-agent systems where overhead for communication and computation scales with the number of agents, posing a significant challenge to the scalability of MARL-based methods~\cite{swarm_intelligence2024, task_offloading2024, uav_swarm2024, collaborative_decision_making2026}. This pushes the field towards optimization techniques that can bridge the gap between powerful, large-scale models and their practical application in autonomous flight~\cite{vision_language_action2025Y2025, vision_language_navigation, llm_robotic_autonomy2025, large_model_embodied_ai2025, robotic_navigation2025}.

\subsubsection{Parameter-Efficient Adaptation for Foundation Models}
Parameter-Efficient Fine-Tuning (PEFT) techniques, especially Low-Rank Adaptation (LoRA), have become the standard for customizing large pre-trained models in UAV-VLN, sidestepping the prohibitive costs of full fine-tuning while enabling effective task-specific adaptation. Fully fine-tuning multi-billion parameter models is computationally infeasible given the hardware constraints of UAVs, often requiring thousands of GPU hours for training, as seen with NavFoM's 4,032 GPU-hour training cycle~\cite{embodied_navigation_foundation_model2024}. PEFT methods address this by freezing the majority of the pre-trained model's weights and only training a small number of additional parameters, dramatically reducing memory and computational overhead~\cite{transformers_remote_sensing2022}. LoRA, which injects trainable low-rank matrices into the transformer layers, has emerged as a dominant approach, seeing widespread adoption for efficiently adapting foundation models for robotics~\cite{vision_language_action2024}, remote sensing~\cite{geochat2023, sarlang-1m_benchmark, skysensegpt2024, omnigeo2025, rs_gpt4v2024, remote_sensing_lvlm2025}, multi-agent systems~\cite{llm_mars2023}, and sim-to-real transfer~\cite{bench_2advlm2025, uav_codeagents2025, grounded_vision2025, urbanvideo_bench}. This strategy is exemplified by open-source Vision-Language-Action (VLA) models like OpenVLA, which are designed for efficient adaptation via LoRA and achieve state-of-the-art performance with fewer parameters~\cite{openvla2024}. An alternative philosophy avoids fine-tuning altogether, leveraging the powerful zero-shot capabilities of pre-trained VLMs for tasks like reward shaping in reinforcement learning loops, further reducing deployment overhead~\cite{vision_language_models2023, reasongrounder2025, prompt_informed_reinforcement2025}.

\subsubsection{Architectural Innovations for Real-Time Inference}
Achieving high-frequency control loops for agile navigation requires architectural innovations that go beyond post-hoc compression, such as specialized network components and asynchronous processing, which strategically manage computational resources by decoupling high-level reasoning from low-level control. Raw inference speed, not just model size, is a critical bottleneck for dynamic flight control, as highlighted by the latency challenges of large VLMs like RT-2, which operate at only 1-5 Hz~\cite{rt2_2023}, and the high computational costs of integrating VLMs into Model Predictive Control loops~\cite{vlmpc_robotic_manipulation, asma2026}. A systematic approach involves co-designing the hardware and software stack to identify performance bounds~\cite{roofline_model2022}. Such models can determine the maximum safe velocity $V_{\text{safe}}$ based on the system's total action latency $T_{\text{action}}$ and maximum acceleration $a_{\text{max}}$:
\begin{equation}
V_{\text{safe}} = a_{\text{max}} \left( \sqrt{T_{\text{action}}^2 + \frac{2d}{a_{\text{max}}}} - T_{\text{action}} \right)
\end{equation}
where $d$ is the obstacle detection distance. A key system-level strategy is the adoption of distributed or split-computing architectures, which range from offloading computationally intensive VLM queries to a remote workstation~\cite{lm_nav2022} to hierarchical pipelines across onboard, edge, and cloud servers~\cite{aero_llm2025, multimodal_uav2025, contextualized_drone2025, agentic_uavs2025}. This principle is refined in advanced architectures that feature Mixture-of-Experts (MoE) heads for adaptive computation~\cite{virtual_view2025} and asynchronous update loops for the vision-language model, enabling high-frequency policy execution, such as the 25 Hz rate achieved by GRaD-Nav++~\cite{grad_nav2025, vision_language_action2025}.

Beyond system-level design, researchers are developing intrinsic model efficiencies by creating lightweight architectures from the ground up. This involves designing networks tailored for resource-constrained embedded platforms, such as lightweight segmentation models coupled with optimized octree mapping~\cite{semantic_octree_mapping}, end-to-end models that avoid explicit mapping to reduce latency~\cite{model_free_uav2024}, or object detectors with efficient backbones like FasterNet~\cite{low_cost2024}. Similarly, reinforcement learning algorithms are selected and optimized for efficiency; for example, Actor-Critic (A2C) methods have shown potential for faster learning with fewer parameters compared to alternatives~\cite{reinforcement_learning_uav2023}, and DRQN-based architectures have achieved inference rates as high as 60 Hz on embedded hardware~\cite{memory_based_deep2018}. Other approaches focus on making the interaction with the model more efficient. This includes developing domain-specific models~\cite{dynamicvl2025}, leveraging structured prompting with predefined skill libraries~\cite{aermani_vlm2025} or protocols~\cite{agentic_robot2025} to constrain the model's generative search space, or using edge-based in-context learning to adapt to new tasks with lightweight prompts, completely avoiding the need for retraining~\cite{prompts_protection2025}.

\subsubsection{Deployment Frameworks and the Efficiency-Robustness Trade-off}
Successful real-world deployment hinges on standardized software frameworks like ROS to manage hardware abstraction and sensor noise, yet the final push for efficiency via model compression introduces a critical dilemma: trading numerical precision for the risk of degrading the generalization capabilities of foundation models. Modular frameworks built on the Robot Operating System (ROS) provide a vital bridge from simulation to reality by offering a model-agnostic architecture for integrating navigation policies with physical hardware~\cite{visual_semantic_navigation2025, sim_to_real2020, agent_controller2023, tacos2025, simultaneous_control_uavs2024}. These frameworks manage the complexities of interfacing with sensors and flight controllers on a diverse range of onboard computers, including custom drones with NVIDIA GPUs~\cite{racevla2025, trackvla2025J2025}, Qualcomm Snapdragon boards~\cite{spatial_assisted2024}, and compact platforms like the Jetson AGX Orin~\cite{uavs_avionics2024}, Xavier NX~\cite{human_drone2025}, and Orin Nano/NX~\cite{dialogue_px42025, grad_nav2025, soranav2025, vl_nav2025}. The feasibility of this approach is shown by systems like TrackVLA achieving 10 FPS inference speed~\cite{trackvla2025}, other VLA systems processing images 6.5 times faster than human operators~\cite{uav_vla2025}, and portable algorithms designed for resource-constrained hardware from the outset~\cite{path_planning_algorithm}. To meet tight constraints, common final steps include knowledge distillation and model quantization~\cite{uav_vlpa2025, geollava_efficient, imaia_assistant, nuplanqa2025, scalable_uav2021, drone_landing2026}, with concrete examples including INT8-quantized OpenVLA models on OrangePi hardware~\cite{vlh2025} and NPU-accelerated inference for object detection~\cite{lightweight_drone2024}.

However, this optimization introduces a significant risk. The BEDI benchmark highlights the poor performance of even full-precision foundation models on many UAV-specific perception tasks~\cite{bedi_benchmark}. Aggressive compression can exacerbate these weaknesses; for instance, compressing high-resolution satellite imagery can cause small but critical objects, sometimes only a few pixels in size, to shrink to a single pixel, resulting in a catastrophic loss of information~\cite{xlrs_bench2025}. Similarly, while photorealistic scene representations like 3D Gaussian Splatting render faster, they can be more difficult to train to convergence than traditional meshes, creating another trade-off between different forms of efficiency~\cite{executable_3d_gaussian2024}. This dilemma motivates a clear need for future work on lightweight foundation models, more efficient pre-training strategies~\cite{foundation_models_survey}, and standardized deployment frameworks tailored for UAVs~\cite{uavs_llms2025, large_language_models2025, distributed_machine_learning2026}, with emerging trends pointing towards edge computing and hardware-accelerated solutions~\cite{uav_navigation_xai2024, real_time_multi_modal2022}.

\section{Future Frontiers and Research Roadmap} \label{sec:Future Frontiers and Research Roadmap}

Building upon the established foundations of single-agent navigation, this section charts a research roadmap for UAV-VLN by addressing the next frontier of challenges centered on collaborative, multi-agent, and multi-domain systems. Our analysis is scoped to a pivotal research trajectory gaining significant traction: the coordination of autonomous Unmanned Aerial Vehicle (UAV) swarms, which leverage Artificial Intelligence (AI) as a key technological enabler to operate as a cohesive unit through advancements in collaborative path planning, decentralized task allocation, and formation control \cite{uav_swarms2024,uxv_swarms2025,coordfield2025,intelligent_swarm2024,uav_swarm2024,collaborative_trajectory2024,uav_agentic2025}. These domains are selected not only for their inherent complexity but because the maturation of foundation models, previously discussed, now provides a viable technological substrate for moving beyond isolated agent operation toward complex, real-world applications. We begin this exploration by examining the extension of intelligence from an individual to a collective, focusing on the architectural, planning, and safety challenges that define the domain of multi-agent UAV swarm coordination.

\subsection{Towards Multi-Agent Systems: UAV Swarm Coordination} \label{subsec:Towards Multi-Agent Systems: UAV Swarm Coordination}

The extension of UAV-VLN to multi-agent swarms is being driven by Multimodal Large Language Models (MLLMs) acting as centralized planners. This paradigm shift enables complex, collaborative task execution by decomposing high-level instructions into synchronized actions, but introduces fundamental architectural trade-offs between centralized reasoning and decentralized execution, and elevates safety from a single-agent problem to a systemic challenge. This transition marks a significant step towards realizing complex, real-world applications such as coordinated disaster response~\cite{disasterm3_2025, autonomous_decision2025, multi_uav_planning2022}, large-scale automated inspection~\cite{automated_inspection2024, multi_uav_trajectory2024}, and logistics~\cite{uav_control2025}. While much academic research has focused on single-UAV reasoning, the field is now increasingly targeting the multi-agent coordination challenges that are critical for industrial and societal applications, reflecting a broad consensus on this research trajectory~\cite{large_language_models2025, uav_swarm2024, satellite_image2025, citynav2025, uav_path2023}.

\subsubsection{MLLMs as Centralized Planners for Task Decomposition}
The dominant trend in UAV swarm coordination is the conceptualization of an MLLM as a centralized 'swarm brain' that interprets high-level commands, decomposes them into spatially-grounded sub-tasks, and allocates them to individual agents~\cite{remote_sensing_foundation_models2024}. This vision, articulated in conceptual frameworks like that of~\cite{multimodal_uav2025}, leverages the advanced reasoning capabilities of MLLMs to manage dynamic missions like firefighting or search-and-rescue~\cite{human_centric_uav2024, safe_uav2025}, with demonstrations coordinating formations of up to a thousand drones~\cite{leviosa_trajectory_generation}. For this centralized planning to be effective, it requires structured, hierarchical reasoning capabilities analogous to those developed for single-agent systems, where high-level cognitive frameworks, such as bidirectional architectures or dedicated integration layers, can serve as a foundation for managing multi-agent coordination~\cite{aerial_agents2025, agentic_uavs2025}. The coarse-to-fine, multi-stage navigation strategies seen in single-agent models serve as a logical prerequisite for a central planner's ability to decompose a complex mission into a coherent set of sub-tasks for a swarm~\cite{geonav2025, citynavagent2025}. This centralized, one-to-many natural language control has been demonstrated in multi-robot systems where a single command is parsed into parallel actions for multiple ground agents~\cite{tacos2025, llm_mars2023}. The effectiveness of this approach hinges on the MLLM's underlying capacity for robust 3D spatial reasoning~\cite{open3d_vqa2024, omnigeo2025}, motivating tighter integration between LLM deliberation and formal path-planning algorithms~\cite{vlm_rrt2025}. While MLLMs are a focal point, this centralized philosophy also encompasses other paradigms, such as hierarchical frameworks using evolutionary game theory for task allocation~\cite{evolutionary_task2022} or density maps for cooperative inspection~\cite{hierarchical_multi_uav2024}. The widespread interest in extending single-agent models to multi-UAV systems is a recurring theme, signaling a clear research trajectory~\cite{drl_uav_navigation, efficient_coverage2024, spar2025, uav_codeagents2025, search_tta2025, airvista_ii2025, trivla2025, path_planning_algorithm}.

\subsubsection{Architectural Trade-offs: Centralized Reasoning versus Decentralized Execution}
The adoption of MLLM-based planners forces a critical trade-off between centralized architectures, which excel at optimal global task allocation but introduce single points of failure, and decentralized or hierarchical architectures that enhance robustness and scalability but complicate coherent group behavior~\cite{distributed_collaborative2026}.

\begin{figure*}[!htb]
    \centering
    \includegraphics[width=0.9\textwidth,height=0.5\textheight,keepaspectratio]{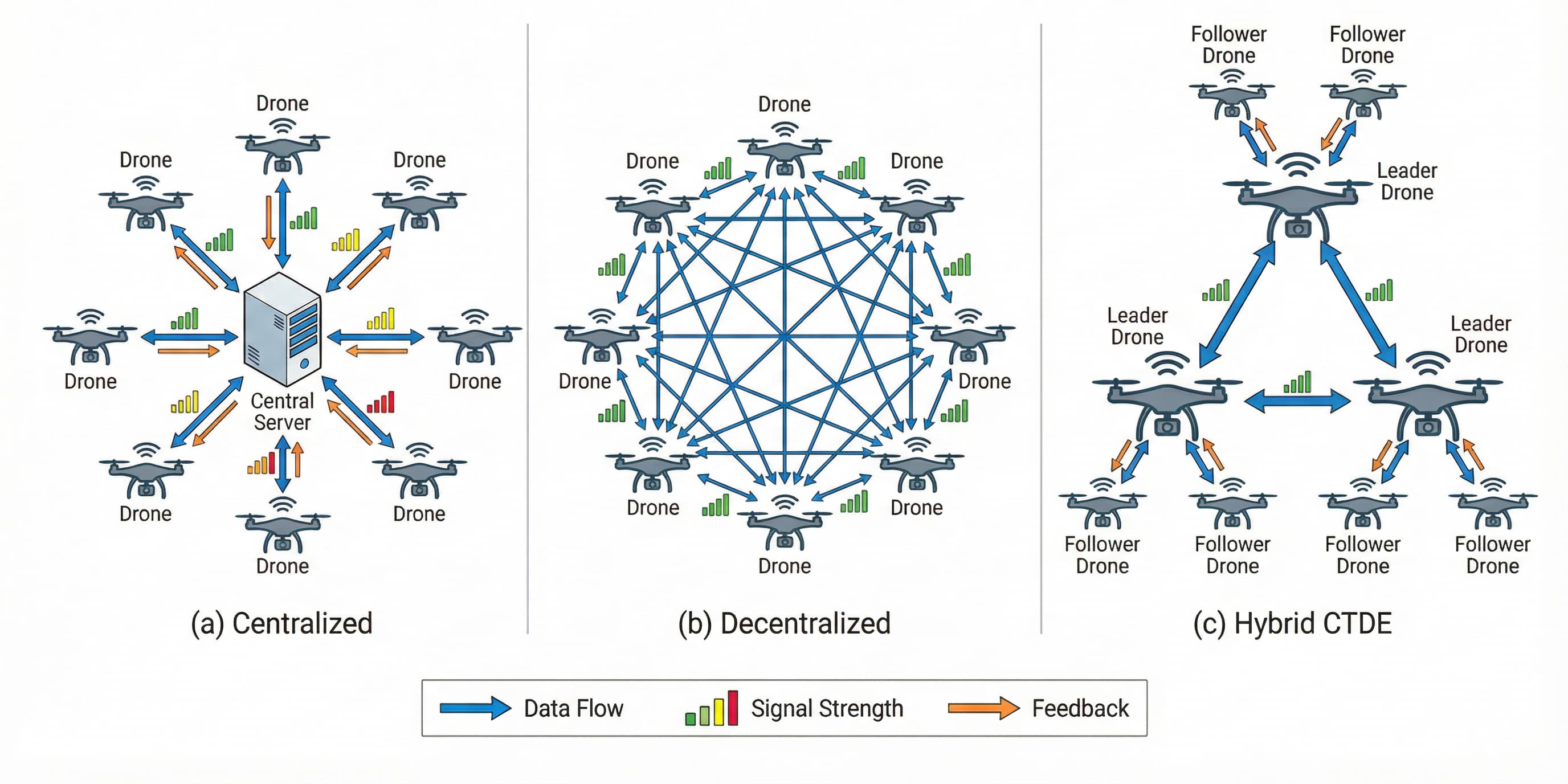}
    \caption{A comparison of coordination architectures for UAV swarms. (a) Centralized models rely on a single planner, offering global optimality at the cost of a single point of failure. (b) Decentralized models enhance robustness through peer-to-peer communication. (c) Hybrid models, such as Centralized Training with Decentralized Execution (CTDE), balance these trade-offs.}
    \label{fig:future_frontiers_and_research_roadmap_auto_1}
\end{figure*}

\begin{figure*}[!htb]
    \centering
    \includegraphics[width=0.9\textwidth,height=0.5\textheight,keepaspectratio]{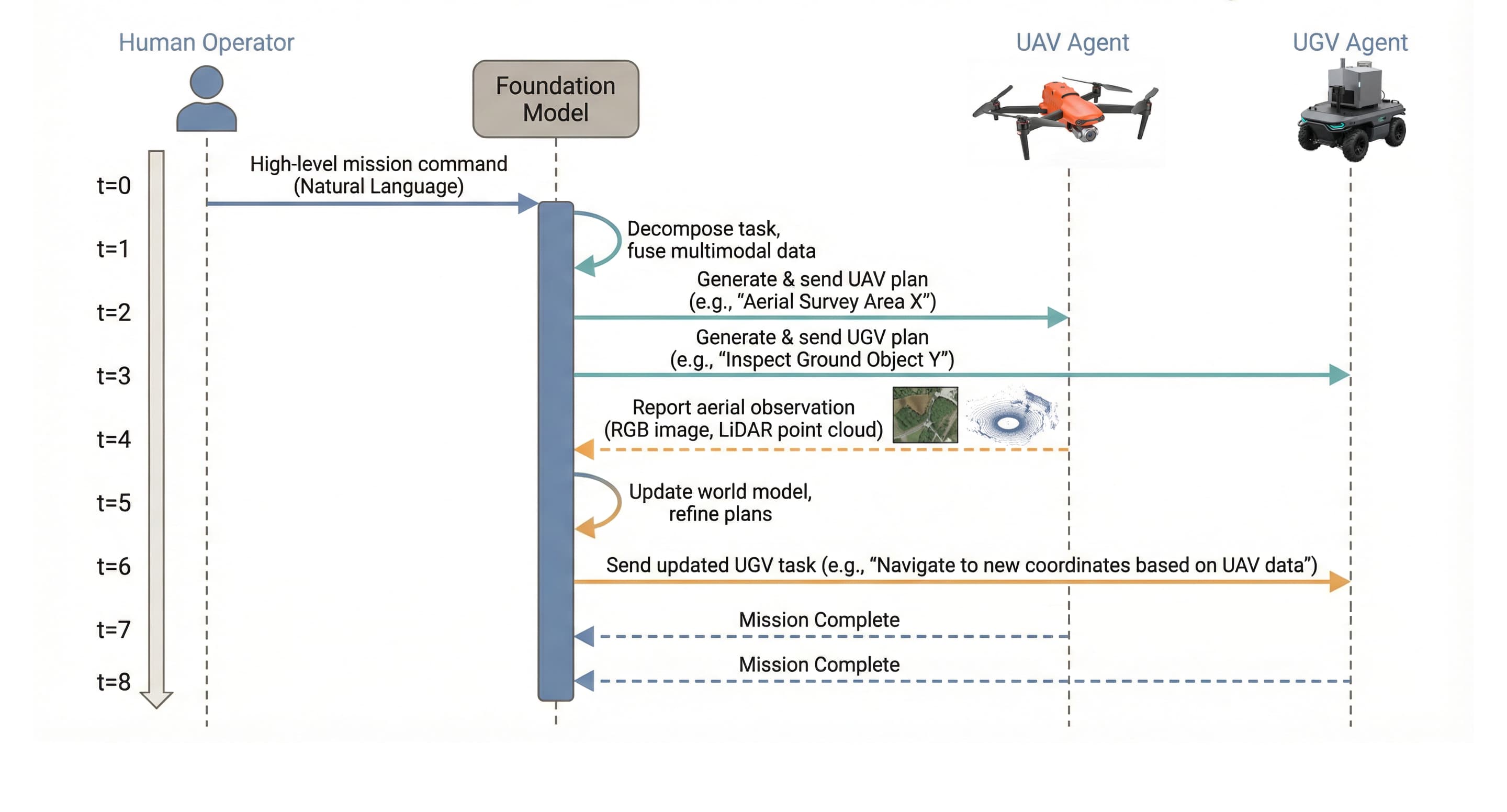}
    \caption{Illustration of a foundation model acting as a collaborative intelligence layer for an air-ground team. The model fuses multimodal sensory data from both the UAV and UGV with high-level human commands to generate synchronized, executable plans for the entire system.}
    \label{fig:future_frontiers_and_research_roadmap_auto_2}
\end{figure*}
 While single drones are simpler to deploy, multi-drone systems offer superior coverage and resilience at the cost of increased complexity and communication overhead~\cite{swarms_uavs2022, large_language_models2025P2025}. The centralized model, implicitly favored by reasoning-heavy approaches~\cite{multimodal_uav2025}, can optimize resource allocation and path planning for the entire swarm~\cite{uav_swarms2026, multi_uav_formation2024}, but its reliance on a single planner makes it vulnerable. In contrast, fully decentralized approaches enhance robustness by allowing agents to operate with limited communication, enabling tasks like distributed onboard mapping with resource-constrained nano-UAVs~\cite{onboard_slam2023}, or using emergent coordination through systems like CoordField to manage heterogeneous swarms~\cite{coordfield2025, distributed_vision2024, cooperative_area_coverage2020}. Hybrid models offer a compromise, ranging from classic Leader-Follower paradigms with dynamic leader election~\cite{intelligent_swarm2024, path_planning_formation_control} to centralized-training-with-decentralized-execution (CTDE) frameworks. Common in multi-agent reinforcement learning (MARL), CTDE leverages a centralized critic to evaluate actions taken by decentralized actors, improving collaborative decision-making in complex environments~\cite{collaborative_decision_making2026}. This paradigm has been applied to optimize cooperative search~\cite{multi_uav_cooperative_search}, collaborative task offloading~\cite{task_offloading2024, dnn_task2024}, and competitive games~\cite{rally2025, multi_agent_drl_uav}. The success of any distributed architecture hinges on robust communication models~\cite{distributed_machine_learning2026} and a shared world representation. This can take the form of a dynamically updated cognitive map for cooperative search~\cite{multi_uav_cooperative2024} or language-grounded 3D maps, which are explicitly designed to be shared among multiple agents to ground commands and deconflict actions~\cite{visual_language_maps2023}. The development of advanced single-agent representations is often pursued with this multi-agent future in mind, aiming to support dynamic interactions within a shared world~\cite{executable_3d_gaussian2024}.

\subsubsection{From Single-Agent Safety to Swarm-Level Guarantees}
Formal safety methods developed for single-agent VLN provide a necessary but insufficient basis for swarms, where the critical open challenge is scaling safety guarantees to manage complex, high-dimensional inter-agent dynamics and prevent cascading failures. Control-theoretic methods like Control Barrier Functions (CBFs) have proven effective for guaranteeing safety for a single agent by maintaining a provable safety margin from obstacles, with extensions to multi-agent systems being a recognized future direction for this line of work~\cite{asma2026}. However, swarm safety is not merely an N-fold replication of single-agent safety; it is a problem of compounded complexity involving inter-agent collision avoidance, formation control, and coordinated hazard response in high-density airspace~\cite{safe_uav_operation, multi_uav_adaptive2024}. Beyond mere collision avoidance, this extends to ensuring predictable and socially compliant behaviors, especially in human-centric environments~\cite{companion_uavs2020}. This introduces a "curse of dimensionality" that complicates the formal verification of swarm behavior, as the state space grows exponentially with the number of agents. Recent work attempts to mitigate this using techniques like hierarchical attention mechanisms that reduce the effective state and action spaces while modeling inter-agent interactions~\cite{uav_swarm_confrontation}. Beyond formal methods, ensuring coherent and safe collaboration also involves higher-level concepts like trust, requiring frameworks that can assure performance in dynamic, multi-agent contexts~\cite{trust_based_assured2024}. The research frontier is therefore focused on developing scalable and formally verifiable safety protocols that can be integrated into multi-agent systems~\cite{safe_uav2025, multi_uav_systems}. Achieving this requires balancing stringent safety guarantees with the flexibility needed to accomplish mission objectives efficiently, a core trade-off in practical swarm deployment that represents a key direction for future work~\cite{skyvln2025}.

\subsection{Air-Ground Collaborative Robotics} \label{subsec:Air-Ground Collaborative Robotics}

The integration of air-ground robotic teams is evolving from rigid, perception-to-action pipelines towards dynamic, intelligent systems where foundation models orchestrate complex collaborative missions by fusing aerial reconnaissance with ground-level interaction~\cite{air_ground_robots2025}. This synergistic approach combines the broad situational awareness of Unmanned Aerial Vehicles (UAVs) with the precise manipulation and interaction capabilities of Unmanned Ground Vehicles (UGVs), unlocking applications in domains such as disaster response, precision agriculture, and infrastructure inspection~\cite{collaborative_trajectory2024, agrivln2024, waypoint_planning2026, real-time_cooperative2024, multi_uav_cooperative2024, multi_uav_planning2022, citynav2025}. The core challenge lies in moving beyond simple data hand-offs to achieve a deeper level of collaborative intelligence, addressing fundamental problems in shared perception, coordination, and high-level reasoning~\cite{transformers_remote_sensing2022, uav_swarms2024}.

\subsubsection{Foundational Architectures: UAV-Led Mapping for Ground Navigation}
Effective air-ground collaboration is traditionally architected as a sequential pipeline where aerial mapping provides the global context required for informed ground-level navigation, but the performance of this model is fundamentally limited by the static nature and computational cost of the maps in dynamic environments. In this dominant 'scout-actor' paradigm, a UAV acts as an 'eye-in-the-sky' to construct a map of the operational area, which is then consumed by a UGV to plan and execute its tasks~\cite{air_ground_robots2025, uav_navigation2025}. The technical maturity of this approach hinges on robustly building a shared world model by fusing multi-modal and multi-perspective data, a challenge addressed by specialized datasets~\cite{multi_modality_ground_to_air} and algorithms that integrate aerial and ground perspectives into unified representations like semantic octrees~\cite{semantic_octree_mapping}. Specific frameworks tackle this challenge by developing cooperative localization algorithms that correct for drift by fusing visual, IMU, and GNSS data through factor graph optimization~\cite{aerial_ground2022} or by integrating visual and LiDAR data from both UAVs and UGVs to create unified 3D maps of urban environments~\cite{collaborative_mapping2026}. More advanced architectures move towards language-grounded 3D scene graphs that can answer complex spatial queries~\cite{open_vocabulary_object_grounding}. The tight integration of these capabilities is exemplified by novel platforms like morphing aerial-ground vehicles that leverage onboard semantic mapping to autonomously navigate and localize objects in unstructured terrain~\cite{morphonavi2025}. However, this pipeline's primary bottleneck is its reliance on a pre-mapped or slowly updated world, which renders it ineffective in rapidly changing scenarios~\cite{multimodal_pretrained_knowledge2023} or GPS-denied environments that demand simultaneous control and mapping~\cite{simultaneous_control_uavs2024}.

\subsubsection{Coordination Strategies and Task Allocation}
Beyond shared perception, robust team performance hinges on the coordination mechanism that dictates how tasks are allocated and roles are adapted, with a clear trade-off between the global optimality of centralized planners and the scalability and robustness of decentralized frameworks. Centralized architectures, where a single entity such as a cloud server or leader robot makes decisions for the entire team, can compute globally optimal plans but introduce communication bottlenecks and single points of failure~\cite{large_models2024}, necessitating integrated control and communication designs to manage team-wide data flows~\cite{scaleable_communication2024}. In contrast, decentralized methods inspired by multi-agent reinforcement learning (MARL) offer greater resilience and scalability by enabling agents to negotiate locally and adapt their roles dynamically~\cite{rally2025, uav_control2025}. Such frameworks allow for emergent, cooperative behaviors capable of dynamic coordination adjustment in response to changing network conditions or task requirements~\cite{coordfield2025, distributed_collaborative2023}, which is essential for tasks like disaster response~\cite{scalable_uav2021}, distributed localization~\cite{distributed_localization2024}, and surveillance~\cite{multi_uav2026}. The choice of architecture directly impacts how complex, collaborative missions are executed, from precision agriculture~\cite{ground_aerial_transportation2025} to large-scale surveillance~\cite{cooperative_multi_agent} or data collection from ground-based sensor networks~\cite{data_retrieving_heterogeneous2002, resilient_uav2025}. While these explicit coordination strategies are powerful, they often struggle to interpret ambiguous, high-level human intent, creating an opening for human-in-the-loop paradigms~\cite{human_drone2025, spatial_assisted2024} and more intuitive, language-driven control.

\subsubsection{The Intelligence Layer: Foundation Models as Collaborative Brains}
Multimodal foundation models are emerging as a unifying intelligence layer for air-ground teams, replacing the explicit programming of mapping and coordination with implicit reasoning capabilities that translate high-level commands and multimodal sensory data into synchronized, multi-agent actions. This paradigm shift positions a Vision-Language Model (VLM) as a collaborative 'brain' that can ingest heterogeneous data streams and generate a coherent, executable plan for the entire team~\cite{large_language_models2025P2025, multimodal_uav2025}.

 A concrete realization of this concept is seen in dynamic warehouse environments, where a VLM orchestrates a heterogeneous swarm of UAVs and AGVs by adjusting control parameters in real-time to maintain formation and avoid obstacles~\cite{swarmvlm2025}. Similarly, hierarchical frameworks use LLMs to interpret high-level instructions and decompose them into guided actions for ground robots based on aerial perception~\cite{hierarchical_language_models2025, llm_mars2023}. This extends to complex manipulation, where VLM-based controllers can identify dynamic sub-goals to enable ground robots to perform intricate pick-and-place tasks within unstructured, real-world environments~\cite{vlmpc_robotic_manipulation, ok_robot2024}. Realizing this potential requires grounding these models in diverse, multi-resolution geospatial data~\cite{omnigeo2025, earthdial2025, streetviewllm2024} and learning generalizable visual representations that bridge low-level spatial features with high-level semantic understanding~\cite{language_driven_representation2023}. By directly grounding language into the physical world~\cite{llm_grounder2023}, this trend moves the field away from engineered, rule-based systems and towards learned, emergent collaboration that integrates perception, navigation, and manipulation within a single intelligence layer~\cite{trackvla2025J2025, aeroduo2025, vision_language_models2023, vision_language_models2025}.

\section{Conclusion} \label{sec:Conclusion}

This survey has charted the intellectual trajectory of UAV-based Vision-and-Language Navigation, from its formalization as a Partially Observable Markov Decision Process to the contemporary era of agentic systems driven by large foundation models. The methodological evolution reveals a clear progression from decomposed robotics pipelines and early end-to-end learning policies toward more integrated and cognitively sophisticated architectures. This shift is characterized by the adoption of transformer-based models for long-horizon temporal reasoning and the construction of explicit, semantically-rich world representations like visual language maps. The current paradigm, dominated by large-scale pre-trained models, leverages Vision-Language Models (VLMs) as high-level cognitive cores for planning and Vision-Language-Action (VLA) models as direct sensorimotor policies \cite{vision_language_navigation2024, large_model_embodied_ai2025, vision_language_action_models}. Most recently, the field has witnessed a pivotal convergence of generative world models with VLA policies, as exemplified by models such as $\pi_0$~\cite{pi0_2024}, GR00T N1~\cite{groot_n1_2025}, and Cosmos-Reason1~\cite{cosmos_reason1_2025}, which equip agents with physical common sense and predictive foresight for more robust, long-horizon navigation. This progress has been built upon an essential ecosystem of high-fidelity simulators, benchmark datasets, and standardized evaluation protocols that enable reproducible research and rigorous comparison.

The advancements chronicled herein signify a fundamental reconceptualization of UAVs, recasting them from remotely operated tools into autonomous agents capable of interpreting and acting upon high-level human intent in complex, unstructured 3D environments \cite{agentic_uavs2025}. This convergence of perception, language, and action unlocks profound applications in domains such as emergency response \cite{uav_search_rescue2023} and smart city management, heralding a future of more intuitive and effective human-robot collaboration. However, the path to robust real-world deployment is obstructed by several persistent and critical challenges. The foremost of these is bridging the sim-to-real gap, ensuring that policies trained in simulation perform reliably on physical hardware amidst unpredictable environmental dynamics \cite{model_free_uav2024, embodied_ai2024Y2024}. Concurrently, achieving robust perception and reasoning under the uncertainty of dynamic outdoor settings and ambiguous linguistic commands remains a significant hurdle \cite{multimodal_spatial_reasoning2025}. Finally, the practical challenge of efficient model deployment ptimizing large, resource-intensive foundation models for the stringent computational and power constraints of onboard UAV hardware presents a major barrier to widespread adoption \cite{foundation_models2023, large_language_models2024, lightweight_drone2024}.

Addressing these challenges requires a forward-looking research agenda that pushes beyond single-agent navigation toward more complex and collaborative frontiers. A primary direction is the development of scalable frameworks for multi-agent and swarm coordination, where multimodal large language models can function as central planners to orchestrate missions that exceed the capabilities of any single agent \cite{uav_swarm2024, collaborative_trajectory2024, uav_control2025}. Another promising frontier is the creation of heterogeneous air-ground collaborative robotic teams, in which UAVs provide aerial intelligence and strategic oversight for ground robots performing physical interaction tasks, thereby enhancing mission safety and efficiency \cite{air_ground_robots2025}. Foundational to these pursuits is the need to establish comprehensive safety-aware learning and evaluation protocols that extend beyond task-success metrics to explicitly measure and guarantee operational robustness in real-world conditions \cite{real-time_cooperative2024, vision_language_action2026}. The integration of world models with VLA architectures~\cite{embodied_llm_wm2025, wm_vla_survey2026} represents a particularly promising direction, enabling agents to simulate future states and reason about physical consequences before acting, which is essential for safe and reliable deployment in complex, dynamic environments. Accomplishing these goals will depend on continued advances in parameter-efficient adaptation and lifelong learning techniques, enabling agents to continuously acquire new knowledge and skills from novel environments and instructions without the need for extensive retraining.\newpage

\bibliography{references_corrected}

 \end{document}